\documentclass{article}
\usepackage[preprint]{ewrl_2026}

\usepackage[utf8]{inputenc}
\usepackage[T1]{fontenc}
\usepackage{hyperref}
\usepackage{url}
\usepackage{booktabs}
\usepackage{amsfonts}
\usepackage{amsmath}
\usepackage{amssymb}
\usepackage{nicefrac}
\usepackage{microtype}
\usepackage{xcolor}
\usepackage{graphicx}
\usepackage{algorithm}
\usepackage{algpseudocode}
\usepackage{subcaption}

\usepackage{xcolor}
\usepackage{soul}
\usepackage{paralist}

\usepackage{multirow}
\usepackage{booktabs}

\usepackage{placeins}

\title{Dynamic Context Scheduling: \\ Learning Beyond the Static Universe}

\author{%
  Martin Mráz \\
  University of Freiburg \\
  \texttt{mrazm@cs.uni-freiburg.de} \\
  \And
  André Biedenkapp \\
  University of Freiburg \\
  \texttt{biedenka@cs.uni-freiburg.de} \\
}

\begin{document}

\maketitle

\begin{abstract}
We study \emph{dynamic context scheduling} as a training instrument for contextual reinforcement learning. Rather than treating intra-episode context variation as a deployment reality, we treat it as a controlled shaping mechanism. Thereby, context evolves within each training episode according to a predetermined schedule, exposing the policy to a richer and more temporally structured region of the environment parameter space.
We introduce \textsc{DynamicCARLEnv}, a framework that wraps contextual environments with pluggable schedule families, such as sinusoidal offsets or cosine annealing. Across CartPole, BipedalWalker and VehicleRacing with \textsc{Carl} contextualization, we show that dynamic schedules match or outperform static context baselines in the out-of-distribution (OOD) regimes. Interestingly, for the more complex BipedalWalker and VehicleRacing environments we also achieve higher in-distribution (ID) evaluation performance.
Preliminary findings indicate that automatic search for multi-stage curricula can successfully discover schedules that improve generalization, performing comparably to extensive grid search over single-stage schedulers.

\end{abstract}

\section{Introduction}
\label{sec:intro}

Reinforcement learning (RL) achieves strong performance in simulation, yet policies trained under narrow,
stationary conditions often fail when environment parameters shift at deployment.
Small changes in friction, payload, actuator gain, or gravity can cause large drops in
return~\citep[see, e.g., ][]{tobin2017domainrandomization, zhou2019environmentprobinginteractionpolicies,9294027,zhang-iclr21a,cobbe-icml20a,wang-neurips21a,benjamins-tmlr23a,kirk-jair23a,gumbsch2024learning,prasanna-rlc24a,suau-rlc24a,iannotta2025contextbridgerealitygap}.
This brittleness reflects a clear mismatch between the single-dynamics regime encountered during
training and the potentially different dynamics faced at test-time.

Several established paradigms address this issue. Predominantly, they aim to expose agents to a broader variety of experiences during training, such that transfer to novel experiences does not pose such a drastic, and potentially catastrophic shift at test-time.
\emph{Domain randomization} (DR) improves robustness by exposing the agent to many environment
instances~\citep{tobin2017domainrandomization, packer2019assessinggeneralizationdeepreinforcement} in an unstructured manner. DR typically samples related environments from a distribution without explicitly informing learning agents about the changes in environments. 
\citet{cobbe-icml20a} proposed a suite of environments that leverage \emph{procedural content generation} (PCG) to vary level structure and visuals of video games. This broadly enables learning of behaviors that are robust to changes in an environment but is highly dependent on the choice of distribution. A too broad distribution might even cause agents to unlearn desirable behavior as sampled environments might require diametrically opposed solutions. Thus, DR and PCG are often coupled with \emph{curriculum learning} techniques to guide learning to more and more complex scenarios \citep{openai-arxiv19a,klink-neurips20a}.
\emph{Robust RL} optimizes worst-case objectives~\citep{pinto2017robustadversarialreinforcementlearning}.
For example, in real-world systems noisy sensor readings are to be expected \citep{robust_rl_adversary_neurips20}. To increase robustness of RL policies the robustness objective is typically modeled as a max-min problem. In this setting the goal is to learn a policy that maximizes the reward under the worst possible adversarial setting \citep{robust_rl_offline_neurips22}. This style of learning can mitigate worst-case outcomes but largely sacrifices performance in average or best case scenarios as the learned policies act conservatively.
\emph{Meta-RL} enables online adaptation to new
environments~\citep[see, e.g., ][]{duan2016rl2fastreinforcementlearning,finn-icml17a,rakelly2019pearl,pmlr-v162-melo22a,grigsby2024amago,beck-tmlr25a,shala-iclr25a}. Such approaches, however, often require complex architectures and expensive bi-level optimization.
Furthermore, Meta-RL often builds on system identification approaches \citep{yu-rss17a,zhou2019environmentprobinginteractionpolicies,evans-icra22a}. Thereby agents attempt to estimate or recognize environment dynamics from a history of observations. While enabling online adaptation, such approaches require further environment interactions at deployment to continue to learn \citep{beck-tmlr25a,sparc-aaai26a}.

Counter to the prior examples, \emph{Contextual RL} (cRL) aims to explicitly provide agents with the knowledge of how environments are related to each other. To this end, cRL assumes that transitions and rewards depend on explicit context variables $c$, and the agent learns a policy conditioned on the current context~\citep{hallak-arxiv15a,modi-alt18a,benjamins-tmlr23a,zhou-aaai26a}.
While na\"ively treating context as another observable can aide in learning more general behavior, more dedicated architectures have been explored in which context is injected into the latent-representations \citep{beukman-neurips23a,prasanna-rlc24a,benad2025shared,engwegen2025modular}.

A key assumption shared by most approaches listed above is that context is treated as a monolithic, static element \citep{biedenkapp2026contextual}. The corresponding context $c$
is sampled once at the beginning of an episode and held fixed until the episode ends.
This design choice is convenient for training stability and credit assignment, but it is also highly restrictive.
Real-world physics do not reset between timesteps; payloads shift, actuators fatigue, and terrain variations
unfold continuously.
More importantly for training, per-episode static sampling discards a rich source of structured signal, i.e., the temporal variation of context \emph{within} an episode.

We propose to exploit intra-episode context variation not as a model of deployment conditions but as a
\emph{training instrument}.
Concretely, we define a family of \emph{dynamic context schedules} that govern how the context $c$ evolves within
each training episode, including sinusoidal sweeps, piecewise constant regimes, linear drift, random walks, cosine
annealing, and hybrid compositions, and study their effect on policy robustness and generalization.
However, our evaluation protocol always
considers policies under static context.
Thus, the temporal variation serves the training-time purposes
\begin{inparaenum}[(i)]
\item 
\emph{regularization};
\item and \emph{exploration}.
\end{inparaenum}
The former prevents policies from overfitting while the latter ensures that agents experience a broader, temporally structured set of state-context pairs.
This training strategy is related to the \emph{dynamic contextual MDP} (dcMDP) formalism of
\citet{tennenholtz-icml23a}, which allows context to evolve exogenously via a process
$\Omega(c_{t+1} \mid c_t)$.
Here, we instantiate $\Omega$ with deterministic, parametric schedules rather than a learned or stochastic model,
keeping the framework simple and interpretable while retaining full cMDP semantics. Thereby, context remains outside
the agent's control, transitions and rewards depend on $c_t$, and the agent may or may not observe $c_t$.

Our work provides the following contributions:

\begin{compactenum}
    \item We provide a novel training paradigm for (contextual) reinforcement learning to facilitate better generalizability of learned policies;
    \item We empirically evaluate a broad suite of scheduling families to study how and which dynamic context changes facilitate better generalization;
    \item We present an open-source extension of the CARL~\citep{benjamins-tmlr23a} benchmark to facilitate easy use of dynamic context schedules;
    \item We conduct a state-space coverage analysis, revealing the counter-intuitive result that dynamic schedules improve generalization without expanding the agent's state-space footprint.

\end{compactenum}

\section{Related Work}
\label{sec:background}

A contextual Markov Decision Process (cMDP)~\citep{hallak-arxiv15a} augments a standard MDP \citep{bellman-jmm57} with the notion of context.
An MDP $M = \left(\mathcal{S}, \mathcal{A}, \mathcal{T}, \mathcal{R}, \rho\right)$ entails a state space $\mathcal{S}$, an action space $\mathcal{A}$, transition dynamics $\mathcal{T}\colon\mathcal{S}\times\mathcal{A}\times\mathcal{S}\to\left[0,1\right]$, a reward function $\mathcal{R}\colon\mathcal{S}\times\mathcal{A}\to\mathbb{R}$ and an initial state distribution $\rho$.
Contextual MDPs introduce the notion of context $c \in \mathcal{C}$ that parametrizes transition function $\mathcal{T}_c$, the reward function $\mathcal{R}_c$ as well as the initial state distribution $\rho_c$ while leaving the state and action spaces unchanged.
Context spaces can be either discrete or described by a distribution $p_{\mathcal{C}}$ \citep{benjamins-tmlr23a}.
Consequently, a cMDP $\mathcal{M}$ represents a family of related MDPs $\mathcal{M}=\left\{M_c\right\}_{c\sim\mathcal{C}}$ and can be seen as a sub-class of partially observable MDPs \citep{kirk-jair23a}.
\citet{tennenholtz-icml23a} presents a special case of cMDPs where contexts are history-dependent and are allowed to evolve over time, which they dubbed dynamic cMDPs (dcMDPs). 
Recently \citet{biedenkapp2026contextual} proposes a novel taxonomy of context that distinguishes between allogenic  (environment-imposed) and autogenic (agent-driven) contexts. They further discuss how context might evolve over time. The notion of autogenic context thus relates to the dcMDP setting, as autogenic contexts may be directly influenced by an agent's behaviour (e.g. battery power), whereas allogenic context is independent of an agent's decisions.

\citet{kirk-jair23a} highlight the utility of the cMDP setting for assessing the zero-shot generalizability of learned policies and propose a novel evaluation protocol. \citet{benjamins-tmlr23a} implement this protocol in their study on the effect of context on training various deep RL agents on their novel CARL benchmark. CARL extends common RL benchmarks and environments \citep[e.g., ][]{brockman-arxiv16a,tassa-arxiv18a,freeman-neuripsdbt21a} with physical contexts, such as gravity, friction or masses of robots.
Their study shows that agents that are simply trained on a distribution of contexts without having explicit access to the true context value tend to learn robust behavior but do not necessarily solve every environment optimally. On the other hand, context aware agents that na\"ively concatenate the context to the state-observation, might be able to perfectly adapt to the changes in environments but may require changes to the RL pipeline (such as choice of hyperparameters \citep{eimer-ecorl21}) to be able to do so.
Beyond na\"ive concatenation, multiple works explore how to employ hypernetworks to facilitate better adaptability of learned policy by learning adapter modules or the weights of a policy directly \citep{beukman-neurips23a,benad2025shared,engwegen2025modular}. 
Opposite to these lines of work, \citet{prasanna-rlc24a,gumbsch2024learning} try to exploit contextual information by injecting contextual information into latent representations of a world model.

Most commonly in contextual RL research, context is treated as a monolithic, static quantity and assumed to be mostly static throughout an episode \citep{biedenkapp2026contextual}. 
To the best of our knowledge, few works explore dynamic changes of context. \citet{gumbsch2024learning} for example, aims to learn when shifts in context, such as the opening or closing of a door, occur. Whenever works aim to learn to estimate context on-the-fly, however, policies might operate under the assumption that context changes between states \citep{RMA-Kumar-2021,ren-iclr23a,ndir-ewrl24a}.
Relatedly, \citet{chandak2020_nonstationarMDP} aims to learn policies that are not only working well with the context, but simultaneously aim to estimate how the MDP changes between episodes, such that the policy is setup well for solving this future task as well.
Counter to these approaches, our work proposes to leverage the fact that most of the training occurs in simulation and that it is possible to explicitly adapt context throughout an episode with the goal to push the generalization capabilities of RL agents.

\section{Dynamic Context Scheduling}
\label{sec:method}

We introduce a training framework that replaces the standard static per-episode context with a parametric intra-episode schedule. The context $c_t$ evolves according to a chosen schedule family throughout each training episode, while evaluation always uses a predefined set of static contexts so that the policy is judged on its generalization to fixed dynamics, not on its ability to track change.

\paragraph{Dynamic Contextual Environments}
\label{subsec:wrapper}

We open-source our \href{https://github.com/mrazmartin/dynamicCARL}{\textsc{DynamicCARLEnv}} as a lightweight wrapper that sits on top of any \textsc{Carl} environment \citep{benjamins-tmlr23a}.
We intercept each environment step to
\begin{inparaenum}[(i)]
    \item advance a parametric context schedule, 
    \item pass the updated $c_t$ to the underlying simulator, and
    \item deliver the chosen context signal to the policy observation.
\end{inparaenum}
The transition $s_t \xrightarrow{a_t,\,c_t} s_{t+1}$ is therefore governed by the live scheduled
context rather than the static episode draw.
Context is read from and written to the simulator via user-supplied getter/setter callables, which
isolates the scheduling logic from environment internals and makes the wrapper applicable to any
\textsc{Carl}-compatible physics backend.

\begin{figure}[t]
  \centering
  \includegraphics[width=0.75\linewidth]{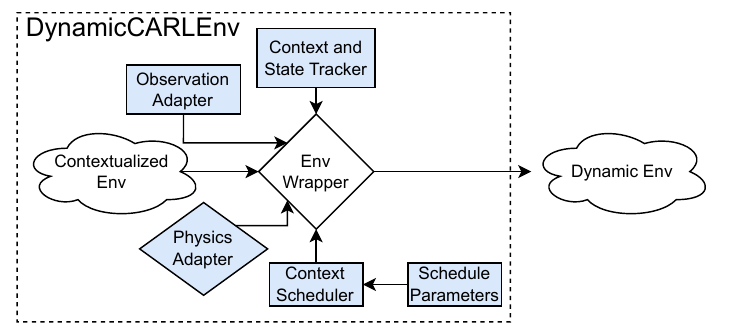}
  \caption[Dynamic context wrapper schematics]{Dynamic context wrapper around CARLEnv. Blue components are user-controlled. The \emph{Context Scheduler} produces the per-step context \(c_t\) given the \emph{Schedule Parameters} (e.g., amplitude, period, drift rate, dwell times, change points). The wrapper applies \(c_t\) through a \emph{Physics Adapter}, controls what the policy observes through an \emph{Observation Adapter} (none, \(c_0\), or \(c_t\)), and logs \((s_t,a_t,r_{t+1},c_t,c_{\mathrm{vis}})\) via the \emph{Context and State Tracker}. The base contextualized environment still receives \(c_0\) at reset.}
  \label{fig:met-dyn_wrapper}
\end{figure}

We study two context observability modes (Table~\ref{tab:obs_modes}).
\emph{Live} mode allows the context to evolve within an episode and can be explicitly observed by an agent, similar to a dcMDP transition model.
\emph{None} mode recovers domain randomization with dynamic changes within an episode. However \emph{None} does not make the context observable and lets us isolate the effect of temporal structure from explicit context conditioning.\footnote{Note, in settings where only static contexts have been considered, \emph{None} is commonly referred to as ``hidden" and live as ``concatenation'' or ``na\"ive''.}

\begin{table}[t]
  \caption{Context observability modes.}
  \label{tab:obs_modes}
  \centering\small
  \begin{tabular}{lll}
    \toprule
    \textbf{Mode} & \textbf{Observation} & \textbf{Policy input} \\
    \midrule
    None & $s_t$ & State only; dynamics change silently. \\
    Live & $[s_t,\; c_t^{*}]$ & State concatenated with current context. ${}^{*}$Context normalized; see text. \\
    \bottomrule
  \end{tabular}
  \vspace{2pt}  
\end{table}

When context is observable by an agent, we normalize $c_t$ to lie in $[-1,+1]$ before appending it to the state.
We motivate this choice as some context might be obtainable with special sensors.
Such sensors would need to be calibrated and provide  sensor-limit bounds.
Crucially, these bounds are set to a physically plausible range that extends \emph{beyond} the evaluation contexts, rather than being fit to the training or evaluation splits.
This ensures that OOD contexts remain well within the normalized range at test time, so the agent is never exposed to saturated inputs.
Further, this ensures that the evaluation grid is not inadvertently encoded into the policy's input representation.

\paragraph{Schedule Families}
\label{subsec:schedules}

We implement six canonical schedule families, each inducing qualitatively different temporal structure on the context trajectory within an episode. Three additional composite families (Sinusoidal Jump, Ornstein--Uhlenbeck, Phased OU) are described in Appendix~\ref{app:schedulers}.

\begin{compactitem}
  \item \textbf{Sinusoidal.} $c_t = c_0 + A\sin(\omega t)$, where $c_0$ is the episode-initial context
    drawn from the training pool, $A$ is the amplitude, and $\omega = 2\pi/T$ the angular frequency.
    The direction is randomized per episode; values are reflected at context bounds.
    Repeatedly sweeps the policy across a wide context range within a single episode.
  \item \textbf{Cosine Annealing.}
    $c_t = c_\text{end}+\tfrac{1}{2}(c_\text{start}-c_\text{end})(1+\cos(\pi t/T_0))$,
    with optional periodic restarts.
    Produces a smooth monotone drift per cycle; with restarts the policy must repeatedly re-adapt from diverse starting points, inspired by \citep{loshchilov2017sgdrstochasticgradientdescent-cosine_anneal}.
  \item \textbf{Continuous Incrementer.}
    $c_{t+1} = c_t \pm \delta$, reflected at bounds.
    The drift direction is fixed per episode (randomized at reset).
    Requires generalization across the full context range as a single monotone sweep per episode.
  \item \textbf{Piecewise Constant.}
    Context is held fixed within segments of random duration and jumps abruptly to a new value
    sampled from a discrete set at each change point.
    Simulates discrete regime shifts (e.g., sudden payload drops or actuator failure).

  \item \textbf{Random Walk.}
    $c_{t+1} = \operatorname{clip}(c_t + \epsilon_t,\; c_\text{min},\; c_\text{max})$,
    $\epsilon_t \sim \mathcal{N}(0,\sigma^2)$.
    Bounded stochastic drift; tests robustness under slow, persistent perturbations.
    
  \item \textbf{Stochastic Jumps.}
    Large discrete displacements of magnitude $\Delta \sim \operatorname{Uniform}(m_\text{lo}, m_\text{hi})$
    occur at random timesteps, with direction sampled independently.
    Unlike piecewise constant, the magnitude and timing are fully random, producing heavy-tailed
    context trajectories.
\end{compactitem}

Composite families can be formed by combining some of the above.
For example, a \emph{sinusoidal jump} layers a sinusoidal
oscillation on top of an intermittently jumping baseline; a \emph{Levy walk} combines stochastic jumps
with additive Gaussian noise to obtain heavy-tailed drift \citep{Zaburdaev_2015_levy-walks}.
Amplitude and step-size parameters are expressed as fractions of the full context range so that
a single hyperparameter setting transfers across context variables with different physical scales.

Appendix~\ref{app:schedulers} shows trajectory examples for all schedule families included in our experiments.

\paragraph{Context Pool Design}
\label{subsec:pool}

Each training episode draws its initial context $c_0$ from a finite \emph{context pool}.
We construct pools hierarchically by recursive midpoint insertion: \emph{Pool-1} contains the single midpoint of the context range; \emph{Pool-3} adds the two range extremes; \emph{Pool-5} inserts midpoints between each adjacent pair; and so on.
This construction keeps the global range constant across pool sizes, isolating the effect of context density from boundary effects.
The episode-initial context is resampled from the pool at the start of each episode, providing inter-episode diversity independently of intra-episode schedule variation.

\paragraph{Multi-Stage Curriculum Scheduling}
\label{subsec:curriculum}

A single schedule family applied for the full training run may not be optimal.
In the early phases the agent might benefit from aggressive exploration to seed the policy with a broad behavioral repertoire, while late training calls for a more focused refinement pass.
We therefore support \emph{multi-stage} schedules that divide training into $K$ fixed-length stages, each governed by an independently configured scheduler from any of the families above.
At each stage boundary the scheduler is re-initialized without interrupting policy training, allowing exploration pressure to be varied throughout the learning lifecycle.

\section{Experiments}
\label{sec:experiments}

\begin{figure}[t]
    \centering
    \includegraphics[width=\linewidth]{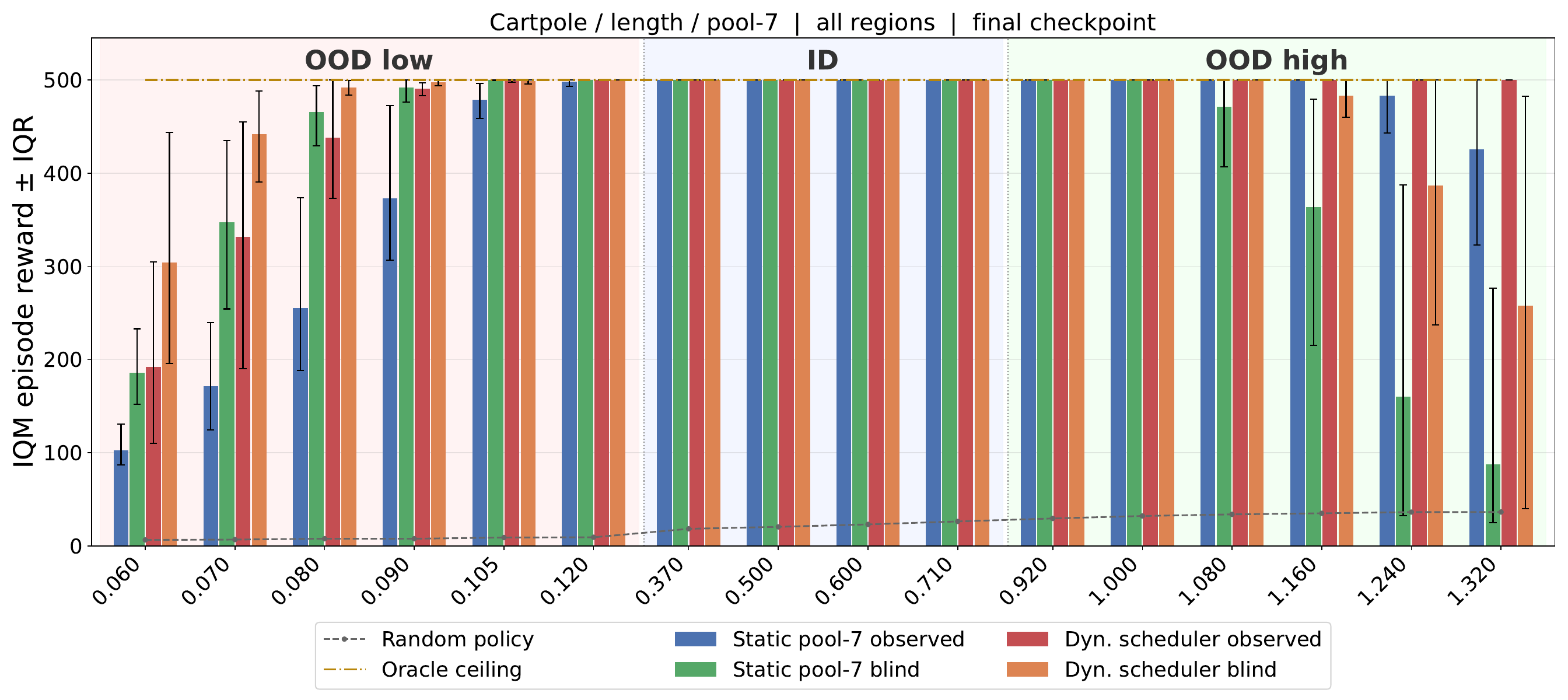}
    \caption{IQM episode reward (bars: Q1--Q3) at the final checkpoint across all 16 evaluation contexts for CartPole. Dashed lines show the random and oracle policy baselines. For dynamic schedulers we show the two best representatives by combined IQM: one observing only states, one also observing the context value.}
    \label{fig:cartpole-bars}
\end{figure}

We empirically investigate whether dynamic context schedules improve policy robustness and generalization compared to static context training.
Our core ablation on CartPole systematically compares schedule families, pool sizes, and observability modes. We then validate the findings on the more demanding BipedalWalker and VehicleRacing environments.
We additionally analyze the state-space coverage induced by different schedules or other multi-context configurations and explore automated multi-stage curriculum search via Optuna \citep{akiba2019optuna}.

\subsection{Experimental Setup}
\label{subsec:setup}

\paragraph{Environments.}
We empirically evaluate the impact of dynamic context variation during training on three contextualized environments from \textsc{Carl}~\citep{benjamins-tmlr23a}.
We use our dynamic context scheduler as introduced in Section~\ref{sec:method}. 
We thoroughly focus our experiment on CartPole, varying \emph{pole length} as the context. This axis is challenging at both OOD extremes: short poles and long poles each present distinct failure modes (see Figure~\ref{fig:cartpole-bars}), unlike contexts such as gravity or force magnitude where only one extreme is typically difficult.
Its fast simulation enables a thorough ablation across scheduler families and configurations.
We further validate the methodology on \textsc{Carl} BipedalWalker, where we vary a single \emph{payload x-axis offset} by attaching a rigid payload to the torso (see Figure~\ref{fig:app-walker-payload}).
Finally, we extend the evaluation to CarRacing, a vision-based environment with pixel observations and more complex dynamics, testing whether the methodology scales to image-based settings.
CarRacing is derived from \textsc{Carl} VehicleRacing: we fix the vehicle type to a single car and replace the original discrete vehicle-type context with a continuous payload offset to support our scheduling framework, hence the rename.
We control the offset along the car's longitudinal and lateral axes (see Figure~\ref{fig:app-racer-payload}).\footnote{We provide an interactive notebook in which users can drive the car themselves under both static and dynamic payload contexts; see \url{https://github.com/mrazmartin/dynamicCARL}}.

\paragraph{Evaluation protocol.}
Following \citet{kirk-jair23a,benjamins-tmlr23a}, we split evaluation contexts into three regimes:
\emph{in-distribution} (ID, within the training context pool boundaries),
\emph{OOD-low} (below the minimum training context value), and
\emph{OOD-high} (above the maximum training context value).
Full context ranges and eval grids are listed in Appendix~\ref{app:env_details}.
All agents are trained with PPO~\citep{schulman-arxiv17a} as implemented in Stable-Baselines3~\citep{sb3} and evaluated deterministically over $30$ episodes per evaluation context at the final training checkpoint.
Following \citet{agarwal-deepRLeval}, we report the interquartile mean (IQM) of episode rewards aggregated across evaluation contexts, which provides a robust estimate of central tendency in the presence of outlier seeds.
Due to computational restrictions we report experiments using $10$ seeds for CartPole, $8$ for BipedalWalker and $5$ for the CarRacing experiments. Full hyperparameter settings are provided in Appendix~\ref{app:hyperparameters}. Finally, when context is observable by the agent, we normalize the context input to stabilize PPO training across both static and dynamic conditions (see Appendix~\ref{app:normalization}).

\subsection{Scheduler Search Results}
\label{subsec:experiments-scheduler-search}

Table~\ref{tab:iqm_last} reports the combined IQM at the last training checkpoint for all three environments. We opt to report final scores throughout to preserve evaluation fairness: selecting the best-achieved checkpoint would require knowledge of generalisation performance during training, which is information we do not assume access to in practice. Best-anytime results are reported in Appendix~\ref{app:additional} for reference.

\begin{table}[b]
    \centering
    \caption{Combined IQM score at the \textbf{last checkpoint}, computed from
      the per-seed \texttt{avg\_combined} metric (average over all ID and OOD
      contexts). Values: IQM\,[Q1,\,Q3]. Bold marks the best scheduler per
      environment.}   
    \label{tab:iqm_last}
    \setlength{\tabcolsep}{5pt}
    \begin{tabular}{lccc}
    \toprule
    \textbf{Condition}
      & \textbf{CartPole}
      & \textbf{Walker}
      & \textbf{CarRacing (lateral)} \\
    \midrule
    Static observed
      & $429.0\;[401.5,\;451.4]$
      & $31.0\;[-34.9,\;63.8]$
      & $-85.9\;[-93.1,\;-76.2]$ \\
    Static blind
      & $410.0\;[383.1,\;436.5]$
      & $91.9\;[60.6,\;113.0]$
      & $103.6\;[-59.0,\;378.9]$ \\
    \midrule
    Best sched.\ observed
      & $\mathbf{468.2}\;[454.7,\;484.3]$
      & $\mathbf{151.4}\;[122.2,\;185.3]$
      & $-35.2\;[-67.3,\;-7.4]$ \\
    Best sched.\ blind
      & $462.9\;[430.9,\;494.1]$
      & $131.1\;[94.9,\;167.6]$
      & $\mathbf{592.3}\;[549.1,\;625.7]$ \\
    \bottomrule
  \end{tabular}
\end{table}

Across all environments and observability modes, the best dynamic scheduler outperforms its static counterpart (Table~\ref{tab:iqm_last}). The gains are modest but consistent for CartPole ($+39$ IQM observed, $+53$ blind), where the ceiling effect of the task limits headroom.
Walker shows the most dramatic improvement under the observed condition ($+120$ IQM). The static observed baseline collapses to an IQM of only $31.0$, likely due to overfitting to the three fixed training contexts when the context value is directly visible, while dynamic schedules force broader generalization.
CarRacing presents the sharpest contrast between observability modes as the best blind dynamic scheduler reaches $592.3$ ($+489$ IQM over static blind), while both observed conditions result in negative IQM, representing failed policies.
We report the lateral axis (COM\_Y) as the primary CarRacing result, as it was the main focus of our scheduler search; longitudinal (COM\_X) results are provided in Table~\ref{tab:ext_carracing_comx}.
The best-anytime results in Appendix~\ref{app:best-any-time} show that observed policies do learn during training but ultimately degrade, suggesting late-stage instability rather than a fundamental inability to train on this environment.
This reversal, where blind outperforms observed by a wide margin in CarRacing but not in simpler environments, is consistent with the context-observability analysis in Appendix~\ref{app:cross-env}.

Figure~\ref{fig:cartpole-bars} shows per-context IQM episode return for CartPole across all 16 evaluation contexts, separated out into OOD-low, ID, and OOD-high regions. The regional breakdown reveals that aggregate IQM improvements can mask meaningfully different dynamics across regimes.
In the ID and OOD-low regions, dynamic and static schedulers perform comparably, with dynamic schedulers providing modest but consistent gains.
The OOD-high region tells a different story. Static baselines tend to exhibit a characteristic generalization drop around the midpoint of training (see Figure~\ref{fig:line-cartpole-ood-high} in the Appendix), where OOD-high performance peaks early before degrading and struggles to recover as the policy specializes to the training distribution. Dynamic schedulers substantially limit this effect. Continuously varying the context during training prevents a policy from overfitting to a fixed regime, so several families either never exhibit the drop or recover more quickly.
Looking at only the best-anytime scores would obscure the degradation that static training accumulates over time. Analogous regional figures for Walker and CarRacing appear in Appendix~\ref{app:additional}; these show the same behaviour to be much less pronounced.
Results per scheduler families are detailed in Appendix~\ref{app:per-family-results}.

\subsection{Multi-Stage Schedulers}
\label{subsec:multistage}

We use Optuna~\citep{akiba2019optuna} with a TPE sampler to automatically search for effective multi-stage schedule curricula on CartPole and BipedalWalker.
Each trial independently selects a scheduler mode from $\{\text{constant},\,\text{sinusoidal},\,\text{cosine annealing}\}$ and its associated parameters for each stage.
The trial objective is the mean across seeds of each seed's best combined evaluation score observed at any stage boundary during training.
For CartPole we search over $K=4$ stages (30\,k\,/\,30\,k\,/\,45\,k\,/\,45\,k steps, 150\,k total) on the pool-7 distribution across 240 trials with 4 seeds each; the stage lengths were chosen to align with the OOD-high generalization loss pattern observed in the single-stage experiments, where performance peaks early and degrades thereafter.
For BipedalWalker we use $K=3$ equal stages (500\,k each, 1.5\,M total) on the pool-3 distribution across 120 trials with 3 seeds each; the 1.5\,M budget is half the full single-stage run to keep search tractable, and $K=3$ equal stages was a pragmatic choice to keep the search space manageable.

The top-20 CartPole configurations are retrained with 10 seeds to obtain reliable IQM estimates.
For BipedalWalker, 16 top configurations that clearly outperformed the static baselines during the 3-seed search are retrained with 8 seeds each.
All comparisons are at the last training checkpoint; single-scheduler baselines are evaluated at the same training budget to ensure a fair comparison.

\begin{table}[t]
  \centering
  \caption{Multi-stage Optuna retrain results at the \textbf{last checkpoint}.
    Walker values are evaluated at the matched 1.5\,M-step budget (half the full
    single-scheduler training run); single-scheduler baselines are re-evaluated
    at the same budget for a fair comparison.
    Values: IQM\,[Q1,\,Q3]. Top-3 retrained trials shown per environment.}
  \label{tab:optuna_retrain}
  \setlength{\tabcolsep}{5pt}
  \begin{tabular}{lcc}
    \toprule
    \textbf{Condition} & \textbf{CartPole} & \textbf{Walker (1.5\,M)} \\
    \midrule
    Static observed        & $429.0\;[401.5,\;451.4]$ & $\phantom{0}77.6\;[\phantom{0}54.2,\;107.8]$ \\
    Static blind           & $410.0\;[383.1,\;436.5]$ & $108.0\;[\phantom{0}72.3,\;127.5]$ \\
    \midrule
    Best sched.\ observed  & $\textbf{468.2}\;[454.7,\;484.3]$ & $119.3\;[\phantom{0}95.7,\;136.6]$ \\
    Best sched.\ blind     & $462.9\;[430.9,\;494.1]$ & $114.8\;[\phantom{0}88.6,\;131.9]$ \\
    \midrule
    Optuna \#1 (T168 / T118) & $459.3\;[445.9,\;480.9]$ & $\textbf{125.0}\;[\phantom{0}72.5,\;161.7]$ \\
    Optuna \#2 (T218 / T117) & $459.3\;[441.1,\;470.0]$ & $115.6\;[\phantom{0}96.0,\;157.4]$ \\
    Optuna \#3 (T145 / T066) & $455.5\;[427.1,\;483.5]$ & $104.1\;[\phantom{0}55.9,\;124.7]$ \\
    \bottomrule
  \end{tabular}
\end{table}

Table~\ref{tab:optuna_retrain} summarises the last-checkpoint IQM for the top-3 retrained trials alongside the static and best single-stage baselines; best-anytime results are in Appendix~\ref{app:optuna-best-anytime}.
On both environments the top retrained configurations largely outperform the static baseline.
However, the best multi-stage trials do not consistently exceed the best single-stage dynamic schedulers found by the grid search over the full set of schedule families.

This comparison comes with important caveats: the Optuna search is restricted to two active scheduler families (sinusoidal and cosine annealing) and the context-observed mode only, whereas the single-stage grid search covers the full diversity of families and both modes.
The 3--4 seeds used to score each trial during search are insufficient to reliably rank configurations: after retraining with more seeds, rankings shift substantially and some configurations that appeared promising fall below the static baseline (Figures~\ref{fig:cartpole-optuna-rank_shift} and~\ref{fig:walker-optuna-rank_shift}).
Given these constraints, we cannot conclude whether the multi-stage structure itself is beneficial beyond what a single well-chosen stage already provides.

\begin{figure}[t]
    \centering
    \begin{subfigure}[c]{0.42\linewidth}
        \centering
        \includegraphics[width=\linewidth]{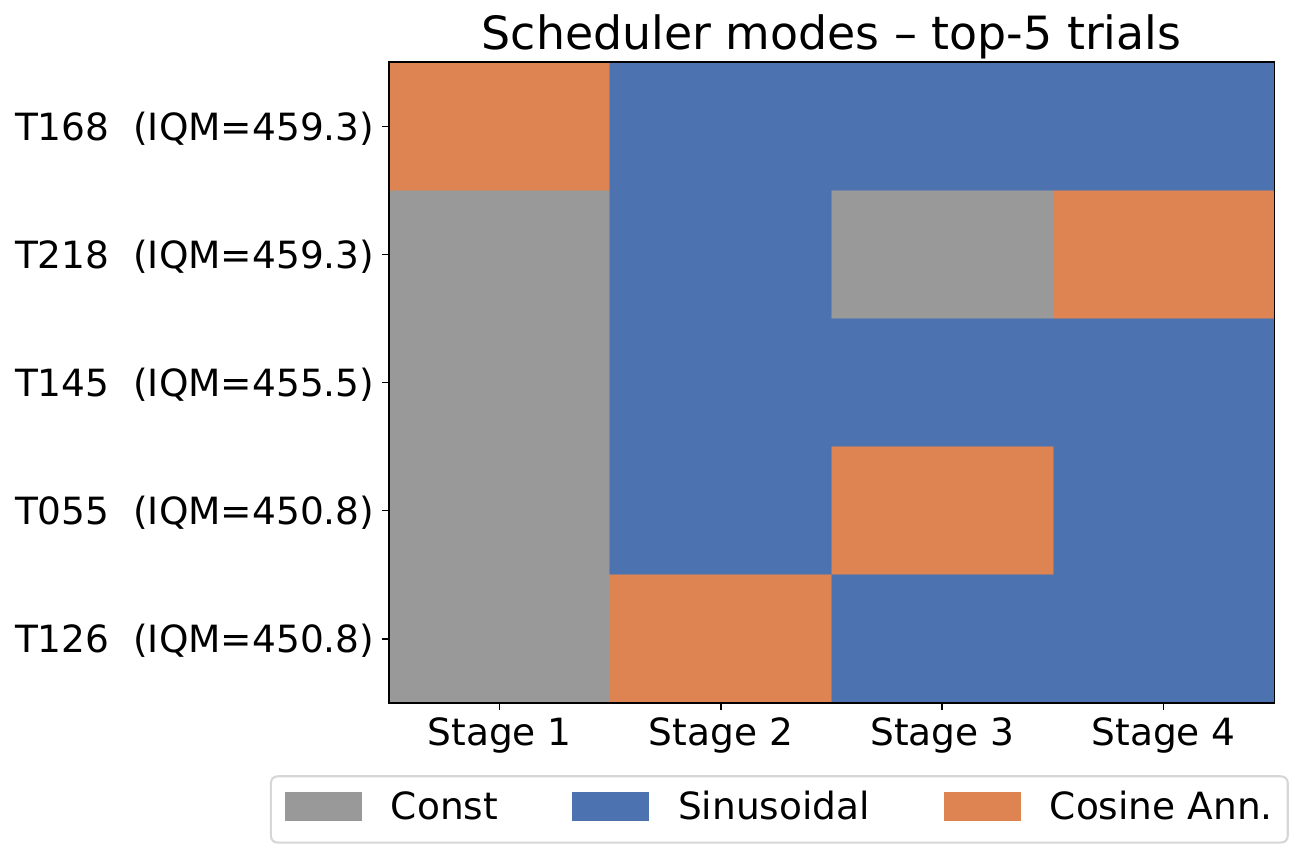}
        \caption{Scheduler mode per stage for the top-5 retrained CartPole trials, sorted by retrain IQM. Grey: constant; blue: sinusoidal; orange: cosine annealing.}
        \label{fig:cartpole-optuna-heatmap}
    \end{subfigure}
    \hfill
    \begin{subfigure}[c]{0.48\linewidth}
        \centering
        \includegraphics[width=\linewidth]{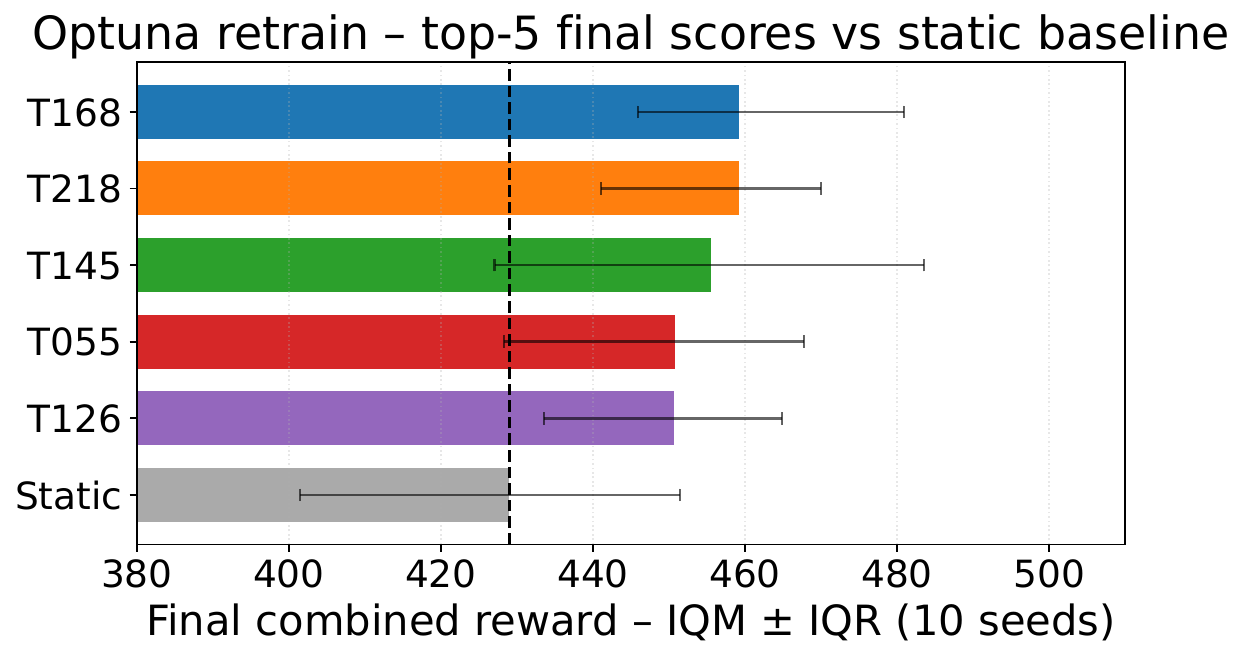}
        \caption{Final-checkpoint IQM\,$\pm$\,IQR for the top-5 retrained CartPole trials (10 seeds). The dashed line marks the static pool-7 baseline IQM.}
        \label{fig:cartpole-optuna-scores}
    \end{subfigure}
    \caption{CartPole Optuna multi-stage retrain results.}
\end{figure}

Nevertheless, inspecting the top CartPole configurations (Figure~\ref{fig:cartpole-optuna-heatmap}) reveals a consistent structural tendency: the first stage is often idle while an active stage follows later, suggesting a \emph{warm-up then diversify} pattern.
Sinusoidal schedules dominate the active stages, and the number of active stages is essentially uncorrelated with final performance, indicating that a single well-placed active stage captures most of the benefit.
Equivalent plots for BipedalWalker are provided in Appendix~\ref{app:walker-optuna} (Figures~\ref{fig:walker-optuna-heatmap} and~\ref{fig:walker-optuna-scores}).

The total number of Optuna trials is comparable to the size of the single-stage grid search over schedule families and their hyperparameters, so both searches operate on a roughly equal compute budget.
For BipedalWalker, the per-trial training cost further restricted the search to a subset of the context range and a training run of 1.5\,M steps rather than the full 3\,M used in the single-stage ablation, which additionally limits what can be concluded from those results specifically.

\subsection{State Space Coverage}
\label{subsec:coverage}

A natural hypothesis is that dynamic schedulers improve OOD robustness by steering the agent through a broader region of the combined state--context space during training.
We test this with a discretised coverage diagnostic on both CartPole and BipedalWalker.

\begin{figure}[t]
    \centering
    \includegraphics[width=0.9\linewidth]{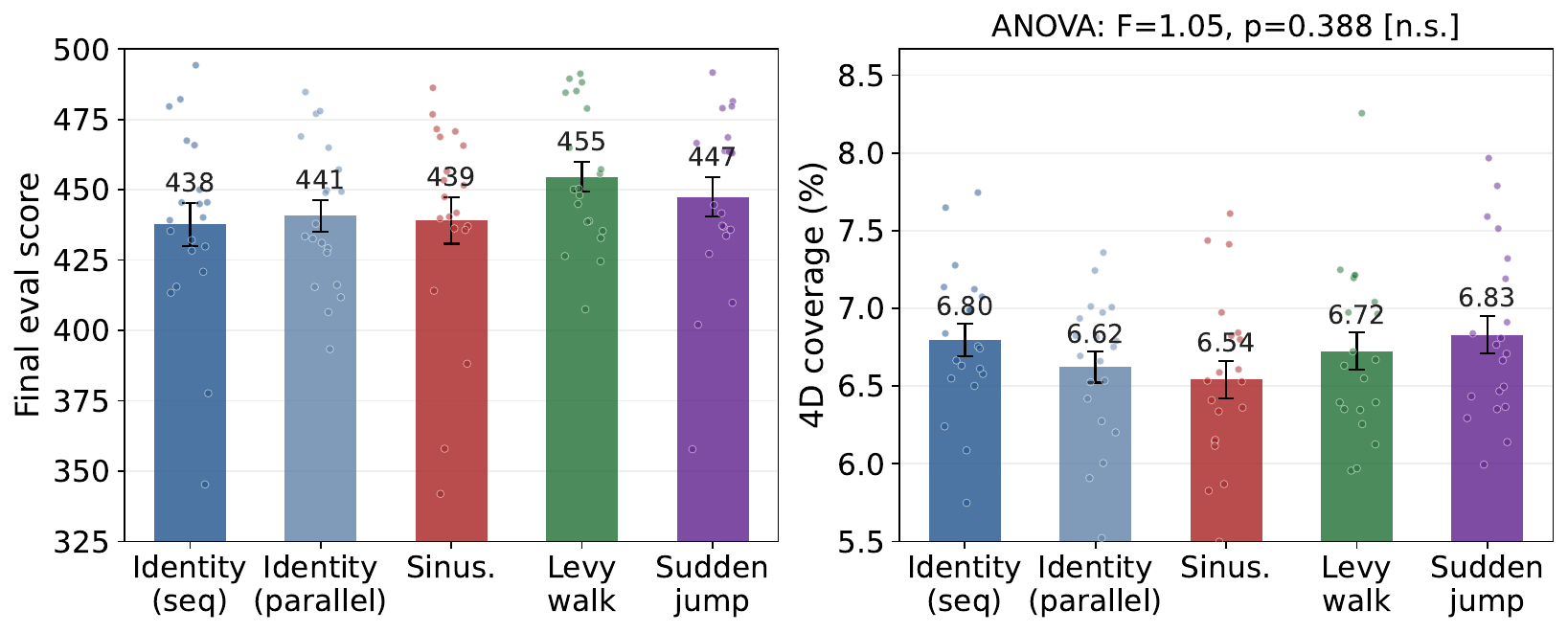}
    \caption{IQM score (left) and 4D state-space coverage (right) per scheduler for CartPole pole-length at pool sizes 3 and 7 (20 seeds each). Points show individual seeds; bars show mean $\pm$ SEM. ANOVA on coverage: $p > 0.05$.}
    \label{fig:cartpole-coverage}
\end{figure}

For CartPole, we run a dedicated coverage diagnostic, pooling pool-3 and pool-7 conditions across 20 seeds each, and discretise the four observation dimensions into $12$ bins each ($12^4 = 20{,}736$ hypercells), measuring the fraction of cells visited over $150$\,k training steps.
Figure~\ref{fig:cartpole-coverage} shows final IQM score and 4D coverage per scheduler.
While evaluation scores vary visibly across conditions, with L\'{e}vy walk and sudden jump outperforming the static baseline, coverage does not follow the same pattern: all conditions cluster between $6.54\%$ and $6.83\%$.
A one-way Analysis of Variance (ANOVA) calculated via SciPy \citep{Virtanen_2020} confirmed that these coverage differences are not statistically significant ($p > 0.05$).
This holds even for the parallel condition that doubles the effective sampling rate, ruling out that raw state throughput is the missing ingredient (see Appendix~\ref{app:par-vs-seq} for the full parallel vs.\ sequential analysis).
We repeat the analysis on BipedalWalker using four independent projections of the 24-dimensional observation space (joint angles, posture, and per-leg phase portraits) and find the same null result across all projections (ANOVA $p > 0.05$ in every case; Appendix~\ref{app:walker-coverage}).

Taken together, these results suggest that dynamic scheduling improves robustness by \emph{restructuring} the training signal within the visited state space, i.e., exposing the policy to richer temporal sequences of $(s_t, c_t)$ pairs within each episode rather than by expanding the set of states visited.
The pool-size scaling study in Appendix~\ref{app:poolsize} further shows that coverage is flat regardless of how many training contexts are used, and that virtually all cells are discovered within the first 50\,k training steps for any scheduler, after which the policy converges to a narrow behavioral manifold.

\subsection{Experiments Summary}
Across CartPole, BipedalWalker, and CarRacing, dynamic context schedules consistently match or outperform static context baselines, with the largest gains on BipedalWalker (observed) and CarRacing (blind).
Multi-stage curriculum search via Optuna reliably improves over the static baseline but does not consistently exceed the best single-stage scheduler found by grid search.
State-space coverage analysis on CartPole and BipedalWalker shows that dynamic and static schedulers visit indistinguishable fractions of the observation space; the benefit of dynamic scheduling therefore derives from the temporal structure of the training signal rather than from broader state exploration.

\section{Discussion and Conclusion}
\label{sec:conclusion}

We proposed dynamic context scheduling as a training instrument to improve zero-shot generalization and out-of-distribution (OOD) robustness in reinforcement learning. By continuously evolving physical parameters within training episodes, agents learn behaviors resilient to temporal shifts, consistently matching or outperforming static baselines across CartPole, BipedalWalker, and CarRacing. Crucially, our coverage analysis revealed that these gains do not stem from broader state-space exploration. Instead, dynamic scheduling restructures the temporal sequence of the training signal, acting as a powerful regularizer that prevents convergence to narrow, over-specialized behavioral manifolds. We also found that context observability requires careful consideration: while explicit observation benefits simpler tasks, "blind" dynamic scheduling proves superior in complex environments like CarRacing.

Our approach is currently limited by the introduction of schedule hyperparameters and the need for manually defined normalization bounds in observable modes. While automated multi-stage search via Optuna mitigates manual tuning, it remains computationally expensive.
Future work will explore integrating dynamic schedules with automated curriculum learning \citep{portelas2019teacheralgorithmscurriculumlearning} to adapt temporal structures to an agent's real-time progress.
Furthermore, rather than na\"ively appending observable context to the state, we plan to investigate more advanced integration methods, such as injecting contextual information directly into the latent representations of world models \citep[e.g.,][]{prasanna-rlc24a, gumbsch2024learning}.

\begin{ack}
The authors are funded by the Deutsche Forschungsgemeinschaft (DFG, German
Research Foundation) – 572775489.
The authors acknowledge support by the state of Baden-Württemberg through bwHPC and the German Research Foundation (DFG) through grant INST 35/1597-1 FUGG.
\end{ack}

\bibliographystyle{plainnat}
\bibliography{bib/strings,bib/local,bib/lib,bib/proc}

\clearpage
\appendix

\section{Schedulers Details}
\label{app:schedulers}
\begin{figure}[ht!]
    \centering
    \includegraphics[width=0.9\linewidth]{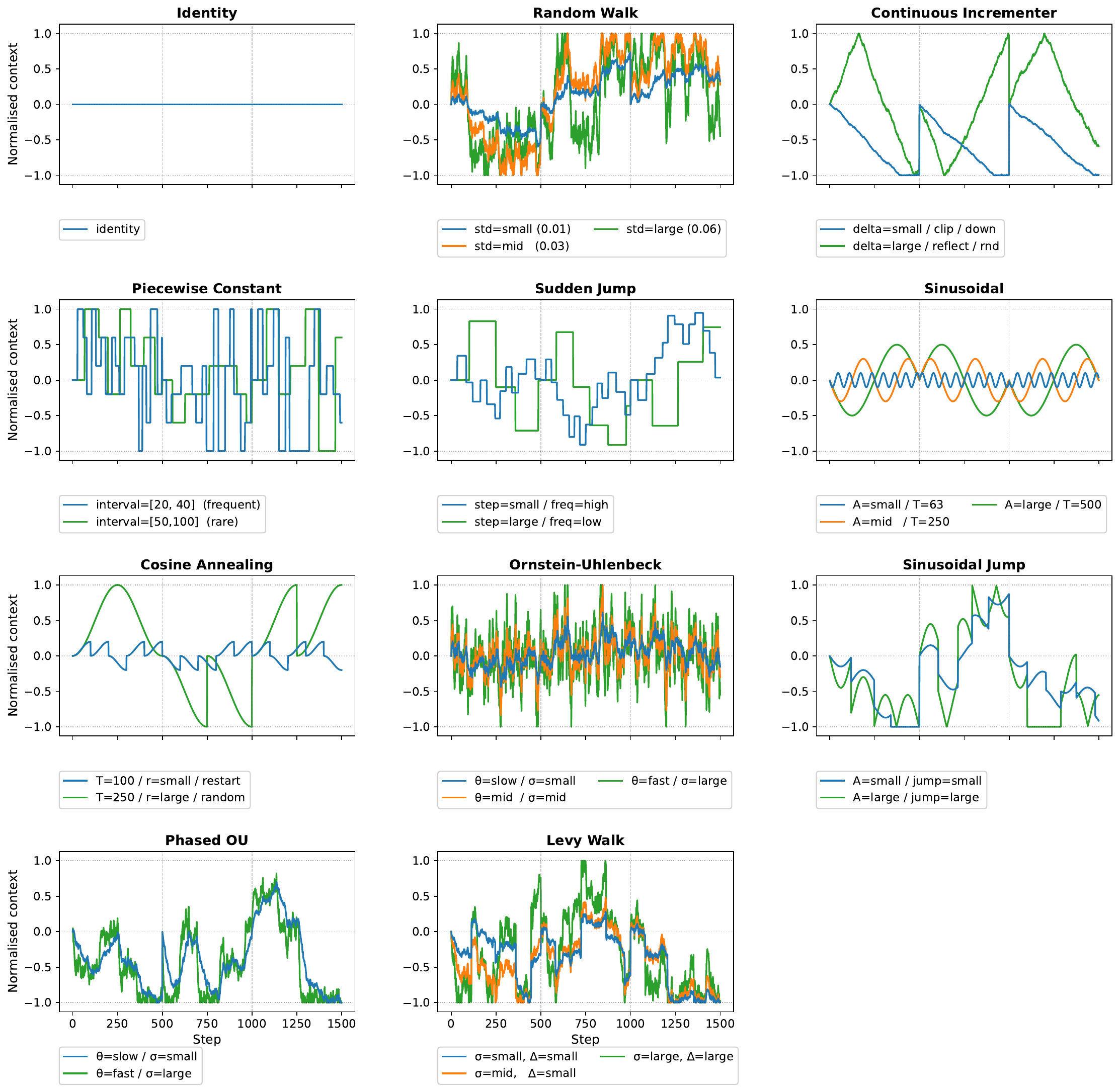}
    \caption{
        Trajectory examples for all eleven schedule families, simulated over three episodes
        of 500 steps each (episode boundaries marked by dashed vertical lines).
        All values are normalised to $[-1,+1]$ relative to the training context range so that
        qualitative structure can be compared across context variables with different physical scales.
        Dotted horizontal lines mark the range boundaries ($\pm 1$) and midpoint ($0$).
        Multiple curves per panel show how the qualitative behaviour changes with the key
        hyperparameters of each family (amplitude, step size, drift rate, etc.).
    }
    \label{fig:scheduler_overview}
\end{figure}

Figure~\ref{fig:scheduler_overview} illustrates trajectory examples for all eleven schedule families over three episodes.

In addition to the six canonical families described in Section~\ref{subsec:schedules}, three
further families appear in the figure above.

\textbf{Sinusoidal Jump.}
Superimposes the sinusoidal oscillation (Section~\ref{subsec:schedules}) onto an intermittently
jumping baseline: at random timesteps drawn from a uniform interval range, the current context
receives an additional signed displacement of magnitude
$\Delta \sim \operatorname{Uniform}(m_\text{lo}, m_\text{hi})$.
The result is a smooth oscillation punctuated by abrupt level shifts, combining the dense
in-episode sweep of the sinusoidal family with the distributional tail coverage of stochastic jumps.

\textbf{Ornstein-Uhlenbeck (OU).}
$dc_t = \theta(\mu - c_t)\,dt + \sigma\,dW_t$, discretised as
$c_{t+1} = c_t + \theta(\mu - c_t) + \sigma\,\epsilon_t$, $\epsilon_t \sim \mathcal{N}(0,1)$. The reversion target $\mu$ is the episode-initial context; $\theta$ controls mean-reversion speed and $\sigma$ the noise level. OU produces temporally correlated drift that stays statistically anchored to a reference value, unlike the unbounded random walk.

\textbf{Phased OU.}
Extends OU by periodically resampling the reversion target $\mu$ from a discrete set of anchor values (constructed identically to the context pool, Section~\ref{subsec:pool}). The retarget interval is drawn uniformly from a specified range, producing a piece-wise-drifting trajectory where each phase smoothly relaxes toward a new anchor.

\textbf{Identity (baseline).}
Context is held constant at its episode-initial value throughout the episode. This is the degenerate special case of a dynamic schedule and serves as the within-experiment reference against which all schedule-induced variation is measured.

\section{Hyperparameters and Training Details}
\label{app:hyperparameters}

All agents are trained with PPO~\citep{schulman-arxiv17a} via Stable-Baselines3~\citep{sb3}.
Table~\ref{tab:app-ppo} lists all hyperparameters; all values are the SB3 defaults and were not tuned — PPO is regarded as robust to hyperparameter variation and fixing the optimizer ensures that observed differences are attributable to the scheduling strategy rather than incidental tuning.
The policy is an MlpPolicy with two fully-connected hidden layers of 64 units each for CartPole and BipedalWalker.
CarRacing uses a MultiInputPolicy with a shared CNN feature extractor (default SB3 NatureCNN) followed by a fully-connected head; context is fed through a separate MLP branch and fused with the CNN output.
The main experiments use a single parallel training environment  (\texttt{N\_TRAIN\_ENVS}\,=\,1); a small auxiliary set of runs explored the effect of increasing the number of parallel
environments within PPO but are not part of the main comparison. Evaluation runs in four parallel workers for all conditions.

\begin{table}[ht!]
  \caption{PPO hyperparameters across environments.}
  \label{tab:app-ppo}
  \centering\small
  \begin{tabular}{lccc}
    \toprule
    \textbf{Hyperparameter} & \textbf{CartPole} & \textbf{BipedalWalker} & \textbf{CarRacing} \\
    \midrule
    Total timesteps            & 150\,000  & 3\,000\,000 & 1\,500\,000 \\
    Rollout steps ($n$)        & 2\,048    & 2\,048      & 2\,048      \\
    Minibatch size             & 64        & 64          & 64          \\
    Learning rate              & $3\times10^{-4}$ & $3\times10^{-4}$ & $3\times10^{-4}$ \\
    Discount $\gamma$          & 0.99      & 0.99        & 0.99        \\
    GAE $\lambda$              & 0.95      & 0.95        & 0.95        \\
    Clip range $\varepsilon$   & 0.2       & 0.2         & 0.2         \\
    Entropy coefficient        & 0.0       & 0.0         & 0.0         \\
    Value function coefficient & 0.5       & 0.5         & 0.5         \\
    \midrule
    Evaluation frequency       & 10\,000   & 50\,000     & 150\,000    \\
    Episodes per eval context  & 30        & 30          & 30          \\
    Seeds                      & 10        & 8           & 5           \\
    Policy network             & MLP [64,\,64] & MLP [64,\,64] & CNN + MLP [64,\,64] \\
    \bottomrule
  \end{tabular}
\end{table}

\FloatBarrier
\section{Environment and Context Details}
\label{app:env_details}

\subsection*{CartPole}

Table~\ref{tab:app-ctx-cartpole} summarises the training and evaluation context grids.
The normalization bounds used for live context observation (Section~\ref{subsec:wrapper}) are manually specified conservative intervals chosen to cover a physically plausible operating range for each variable, in this case $(0.05, 2.0)$\,m for the pole length.
This normalization is motivated by PPO training stability; its effect is documented in Appendix~\ref{app:normalization}.

\begin{table}[ht!]
  \caption{CartPole context ranges and evaluation splits.}
  \label{tab:app-ctx-cartpole}
  \centering\small
  \begin{tabular}{lcccc}
    \toprule
    \textbf{Context} & \textbf{Training range} & \textbf{Eval ID} & \textbf{Eval OOD-low} & \textbf{Eval OOD-high} \\
    \midrule
    Pole length (m)    & $[0.35,\,0.75]$ & $\{0.37,0.50,0.60,0.71\}$   & $\{0.06 \ldots 0.12\}$ & $\{0.92 \ldots 1.32\}$ \\
    \bottomrule
  \end{tabular}
\end{table}

Table~\ref{tab:observation_bounds} lists the CartPole observation space bounds used to set the bin boundaries in the 4D coverage diagnostic.

\begin{table}[htbp]
    \centering
    \caption{Observation Space Bounds of CartPole}
    \label{tab:observation_bounds}
    \begin{tabular}{lll}
    \toprule
    \textbf{Observation} & \textbf{Min} & \textbf{Max} \\
    \midrule
    Cart Position         & -4.8 & 4.8 \\
    Cart Velocity         & $-\infty$ & $\infty$ \\
    Pole Angle            & $\sim$ -0.418 rad (-24$^{\circ}$) & $\sim$ 0.418 rad (24$^{\circ}$) \\
    Pole Angular Velocity & $-\infty$ & $\infty$ \\
    \bottomrule
    \end{tabular}
\end{table}

The pool-7 used in the main CartPole experiments contains the 7 midpoint-inserted values
$\{0.35,\,0.41,\,0.48,\,0.55,\,0.62,\,0.68,\,0.75\}$ for pole length and
$\{5.0,\,7.5,\,10.0,\,12.5,\,15.0\}$ for force magnitude and gravity (pool-5).

\FloatBarrier
\subsection*{BipedalWalker}

A rigid payload of mass $2.0$\,kg and radius $0.5$\,m is attached to the torso. The context COM\_X shifts this payload along the body's longitudinal axis. Training range is $[-0.6,\,+0.6]$\,m; pool-3 contains $\{-0.6,\,0.0,\,+0.6\}$. The normalization bounds are $(-2.0,\,+2.0)$\,m, a conservatively chosen physical operating range for the payload offset.

\begin{table}[ht!]
  \caption{BipedalWalker evaluation splits (COM\_X, m).}
  \label{tab:app-ctx-walker}
  \centering\small
  \begin{tabular}{lccc}
    \toprule
    & \textbf{Eval ID} & \textbf{Eval OOD-low} & \textbf{Eval OOD-high} \\
    \midrule
    COM\_X (m) & $\{-0.4,-0.2,0.0,+0.2,+0.4\}$ & $\{-1.05,\ldots,-0.65\}$ & $\{+0.9,\ldots,+1.7\}$ \\
    \bottomrule
  \end{tabular}
\end{table}

Table~\ref{tab:bipedal_walker_bounds} lists the BipedalWalker observation space bounds used to define the coverage projections.

\begin{table}[htbp]
\centering
\caption{BipedalWalker Observation Space Bounds}
\label{tab:bipedal_walker_bounds}
\begin{tabular}{llcc}
\toprule
\textbf{Index} & \textbf{Description} & \textbf{Min} & \textbf{Max} \\
\midrule
0   & Hull Angle             & $-\pi$     & $\pi$     \\
1   & Hull Angular Velocity  & -5.0       & 5.0       \\
2   & Horizontal Speed       & -5.0       & 5.0       \\
3   & Vertical Speed         & -5.0       & 5.0       \\
4   & Hip 1 (Joint Angle)    & $-\pi$     & $\pi$     \\
5   & Hip 1 Speed            & -5.0       & 5.0       \\
6   & Knee 1 (Joint Angle)   & $-\pi$     & $\pi$     \\
7   & Knee 1 Speed           & -5.0       & 5.0       \\
8   & Leg 1 Contact          & 0.0        & 5.0       \\
9   & Hip 2 (Joint Angle)    & $-\pi$     & $\pi$     \\
10  & Hip 2 Speed            & -5.0       & 5.0       \\
11  & Knee 2 (Joint Angle)   & $-\pi$     & $\pi$     \\
12  & Knee 2 Speed           & -5.0       & 5.0       \\
13  & Leg 2 Contact          & 0.0        & 5.0       \\
14--23 & LiDAR Measurements (x10) & -1.0  & 1.0       \\
\bottomrule
\end{tabular}
\end{table}

\FloatBarrier
\subsection*{CarRacing}

Two context axes: COM\_X and COM\_Y (vehicle center-of-mass offsets), each in $[-0.8,\,+0.8]$\,m, with pool-3 containing $\{-0.8,\,0.0,\,+0.8\}$ per axis. The normalization bounds are $(-2.0,\,+2.0)$\,m per axis, conservatively covering the physical operating range. The payload has mass $1.0$\,kg and radius $0.5$\,m. Table~\ref{tab:app-ctx-carracing} summarises the evaluation splits; both axes use identical ranges.

\begin{table}[ht!]
  \caption{CarRacing evaluation splits (COM\_X and COM\_Y, m). Both axes share the same ranges.}
  \label{tab:app-ctx-carracing}
  \centering\small
  \begin{tabular}{lccc}
    \toprule
    & \textbf{Eval ID} & \textbf{Eval OOD-low} & \textbf{Eval OOD-high} \\
    \midrule
    COM\_X / COM\_Y (m)
      & $\{-0.5, \ldots ,+0.5\}$
      & $\{-1.5,\,-1.3,\,-1.1,\,-0.9\}$
      & $\{+0.9,\,+1.1,\,+1.3,\,+1.5\}$ \\
    \bottomrule
  \end{tabular}
\end{table}

\begin{figure}[ht!]
  \centering
  \begin{subfigure}[c]{0.45\linewidth}
    \centering
    \includegraphics[width=\linewidth]{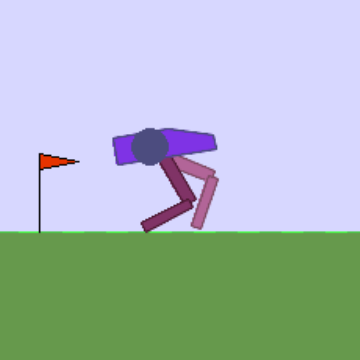}
    \caption{BipedalWalker with attached payload (dark mass on torso).}
    \label{fig:app-walker-payload}
  \end{subfigure}
  \hfill
  \begin{subfigure}[c]{0.45\linewidth}
    \centering
    \includegraphics[width=\linewidth]{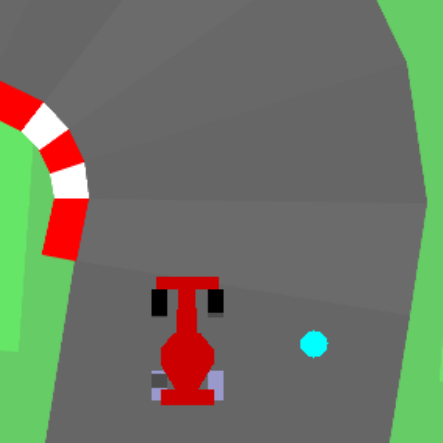}
    \caption{CarRacing with attached longitudinally offset payload (cyan marker).}
    \label{fig:app-racer-payload}
  \end{subfigure}
  \caption{Visualisation of the payload-extended environments. Both payloads are novel
  additions to the \textsc{Carl} benchmark introduced in this work. Shifting the payload
  position changes the center-of-mass of the agent, altering its dynamics in a
  physically grounded and continuously parameterisable way.}
  \label{fig:app-env-payloads}
\end{figure}

\FloatBarrier
\section{Additional Results}
  \label{app:additional}

\subsection{Cross-Environment Scheduler Analysis}
\label{app:cross-env}

We present three complementary views of the scheduler search results across all environments, aggregating over the full set of scheduler families evaluated.

\begin{figure}[h]
\centering 
\includegraphics[width=\linewidth]{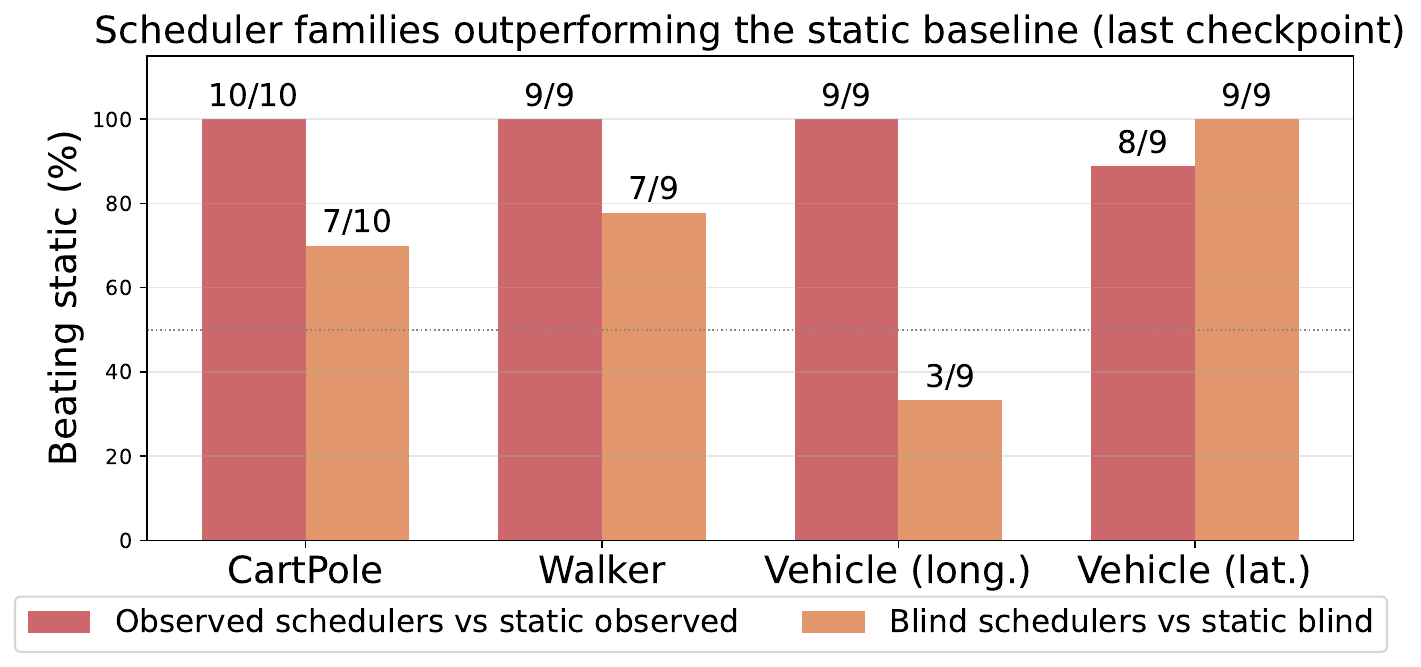}
\caption{Fraction of scheduler families (last checkpoint IQM) that outperform
  the static baseline per environment and mode. Numbers above bars indicate
  the exact count over the total number of families evaluated.}
\label{fig:beat-static}
\end{figure}

\paragraph{Fraction of schedulers outperforming the static baseline.}
Figure~\ref{fig:beat-static} shows, for each environment and context-observability mode, the fraction of scheduler families whose last-checkpoint IQM exceeds the corresponding static baseline. Context-observed dynamic schedulers beat the static observed baseline in nearly all cases (89--100\% across environments), confirming that any dynamic curriculum helps when the agent can observe the context. The blind mode is more nuanced: in CartPole and Walker roughly 70--78\% of blind schedulers improve over the static blind baseline, whereas in CarRacing the picture splits sharply --- the longitudinal (COM\_X) static blind baseline is exceptionally strong, with only 3 out of 9 schedulers surpassing it, while the lateral (COM\_Y) static blind baseline is weak and all 9 schedulers beat it.

\begin{figure}[h]
\centering 
\includegraphics[width=\linewidth]{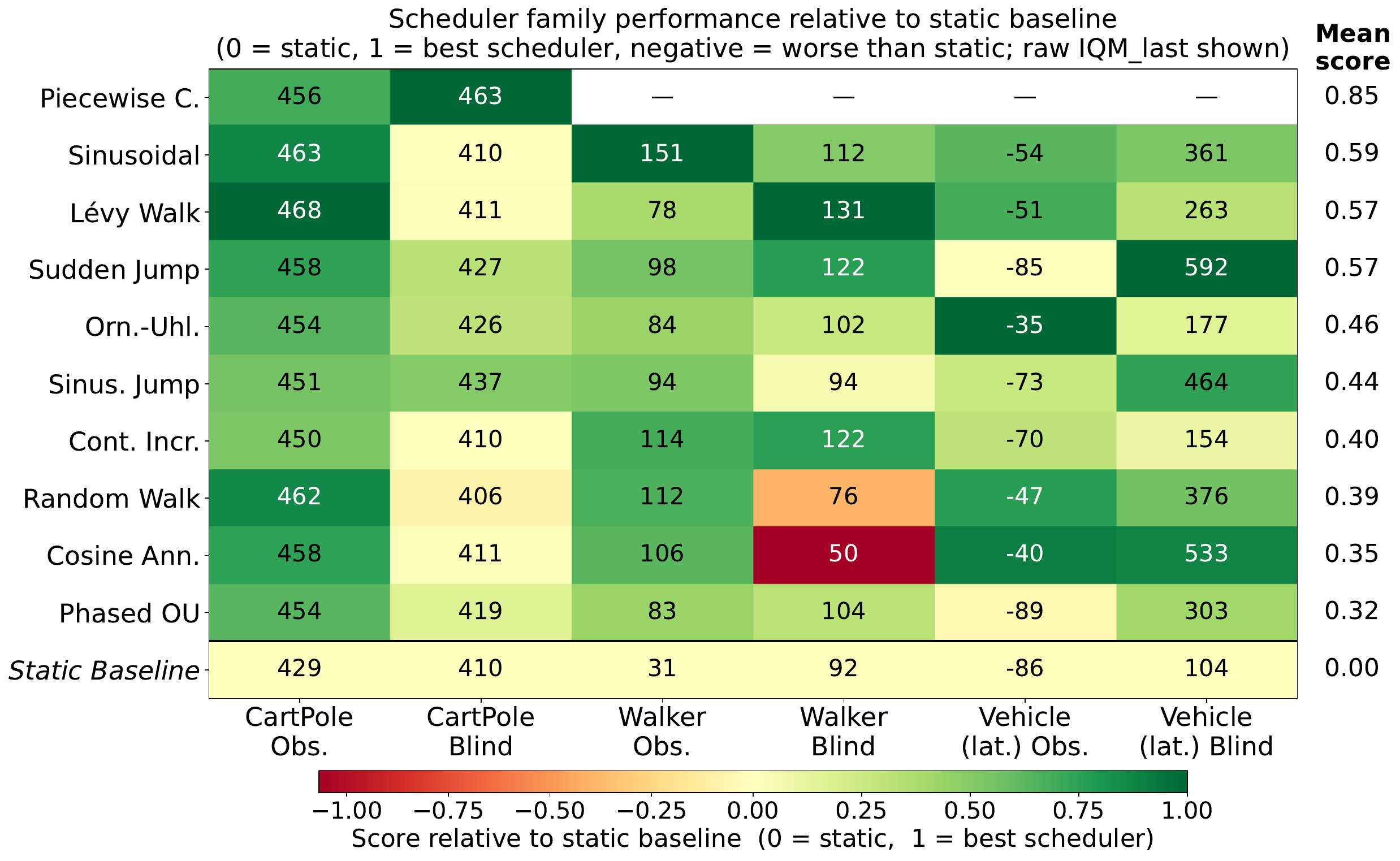}
\caption{Heatmap of scheduler family performance relative to the static
  baseline. Colour encodes the static-anchored score: yellow ($= 0$) matches
  the static baseline, green ($= 1$) is the best scheduler in that column,
  red indicates worse than static. Raw IQM values (last checkpoint) are
  shown in each cell. Families are sorted by mean score across all columns
  (right margin).}
\label{fig:cross-env-ranking}
\end{figure}

\paragraph{Cross-environment performance ranking.}
Figure~\ref{fig:cross-env-ranking} shows a heatmap of each scheduler family's performance relative to the static baseline (score $= 0$) and the best scheduler in that column (score $= 1$), with families sorted by their mean score across all six columns. No single family dominates across all environments and modes: Sinusoidal, L\'{e}vy Walk, and Sudden Jump rank most consistently above the static baseline on average (Piecewise Constant ranks highest overall but was only evaluated on CartPole, leaving its Walker and CarRacing cells empty), while Phased OU and Cosine Annealing are the weakest overall. Dynamic schedulers reliably outperform the static baseline in the context-observed columns (zero red cells for CartPole and Walker observed; only one marginal exception for CarRacing observed), and the CarRacing (lat.)\ blind column is entirely green. The most notable failures occur in the Walker blind column, where Cosine Annealing and Random Walk fall clearly below the static blind baseline, suggesting these families are particularly sensitive to the absence of context information in that environment.

\begin{figure}[h]
    \centering
    \includegraphics[width=\linewidth]{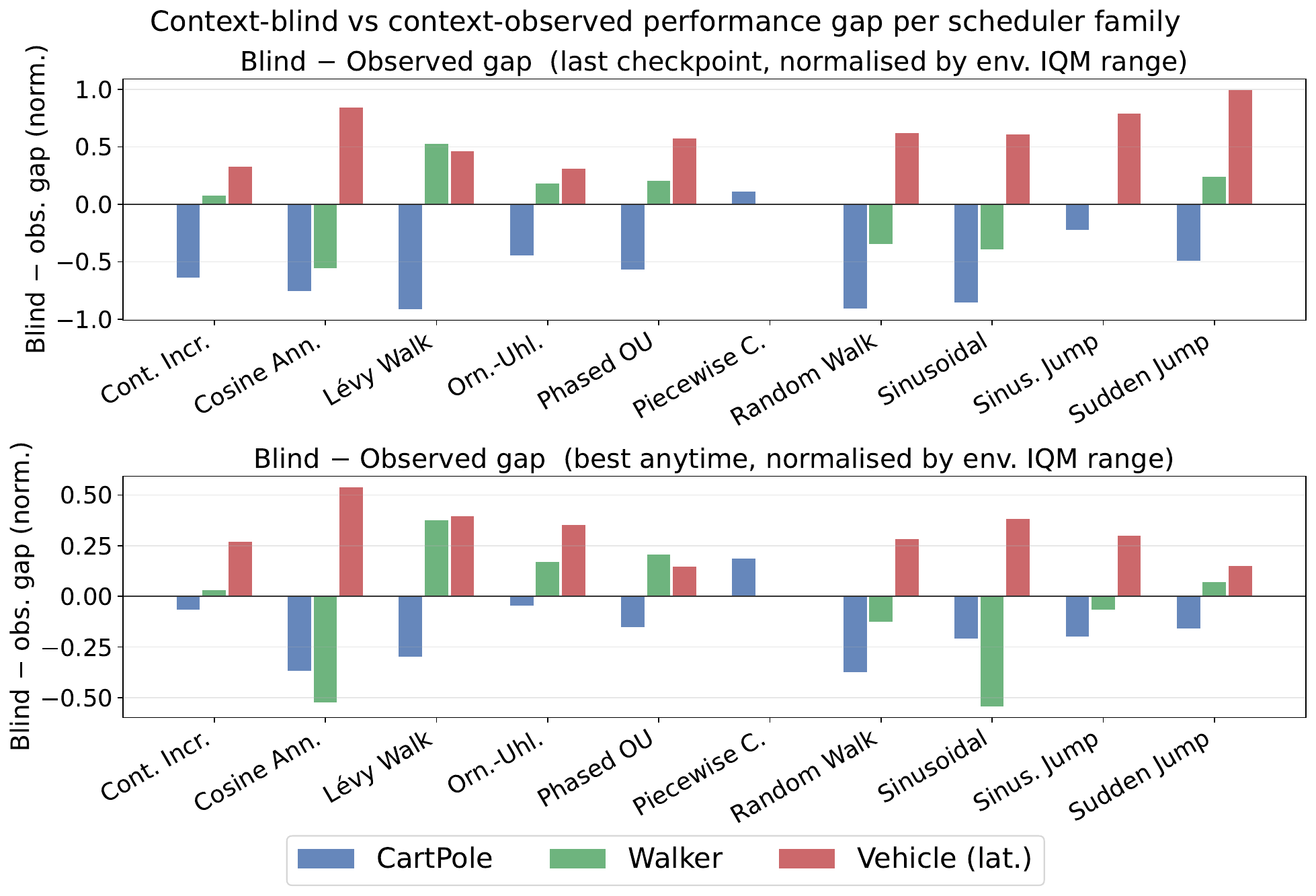}
    \caption{Blind$-$observed IQM gap per scheduler family and environment, normalised by the environment's IQM range (max$-$min across all schedulers). Positive values indicate blind schedulers outperform observed ones; negative values indicate the reverse. Top panel: last checkpoint; bottom panel: best anytime.}
    \label{fig:obs-blind-gap}
\end{figure}

\paragraph{Context-observed vs.\ context-blind gap.}
Figure~\ref{fig:obs-blind-gap} quantifies the blind$-$observed IQM gap per scheduler family, normalised by each environment's own IQM range so that results are comparable across environments. The pattern is consistent across nearly all scheduler families: CartPole benefits from observing the context (negative gap, observed wins), CarRacing benefits strongly from \emph{not} observing it (large positive gap, blind wins), and Walker sits in between with mixed results depending on the family. This environment-level reversal holds independently of which scheduler is used, suggesting the effect is driven by the environment rather than the scheduler choice. Notably, Cosine Annealing, Random Walk, and Sinusoidal are the only families that consistently prefer the observed mode in both CartPole and Walker, while Sudden Jump and Cosine Annealing show the largest blind advantage in CarRacing.

\FloatBarrier
\subsection{Best-Anytime Checkpoint Results for Scheduler Search}
\label{app:best-any-time}
Table~\ref{tab:iqm_best} reports the best-anytime IQM --- the highest \texttt{avg\_combined} score achieved at any evaluation checkpoint during training --- as a complement to the last-checkpoint results in Table~\ref{tab:iqm_last}. Dynamic schedulers improve over the static baseline across all environments and modes at peak performance, and the margins are generally larger than at the final checkpoint, indicating that the best policies are learned earlier in training and partially lost to instability or overfitting by the end.

The gains are most pronounced for Walker, where the best observed scheduler reaches $193.0$ vs.\ static observed $134.2$ ($+59$ IQM), and for CarRacing, where the best blind scheduler reaches $778.9$ vs.\ static blind $617.7$ ($+161$ IQM). CartPole improvements are modest in comparison, consistent with the ceiling effect noted in the main text. Notably, the best CarRacing observed scheduler ($634.1$) now substantially exceeds the static observed baseline ($358.5$) at peak, though it still falls short of the static blind baseline ($617.7$), confirming that context observability remains a net negative in that environment even at best-achieved performance.

\begin{table}[t]
  \centering
    \caption{Combined IQM score of the \textbf{best anytime checkpoint}, computed from the per-seed \texttt{avg\_combined}
  metric (average over all ID and OOD contexts). Values: IQM\,[Q1,\,Q3]. Bold marks the best scheduler per environment.}
    \label{tab:iqm_best}
    \setlength{\tabcolsep}{5pt}
    \begin{tabular}{lccc}
    \toprule
    \textbf{Condition}
      & \textbf{CartPole}
      & \textbf{Walker}
      & \textbf{CarRacing (lateral)} \\
    \midrule
    Static observed
      & $449.7\;[440.3,\;462.8]$
      & $134.2\;[\phantom{0}89.5,\;167.9]$
      & $358.5\;[183.6,\;450.8]$ \\
    Static blind
      & $463.4\;[453.4,\;474.0]$
      & $146.9\;[118.3,\;181.4]$
      & $617.7\;[272.5,\;803.1]$ \\
    \midrule
    Best sched.\ observed
      & $\mathbf{479.6}\;[469.3,\;491.0]$
      & $\mathbf{193.0}\;[156.7,\;233.4]$
      & $634.1\;[520.7,\;659.4]$ \\
    Best sched.\ blind
      & $479.3\;[465.5,\;495.9]$
      & $175.5\;[151.9,\;200.5]$
      & $\mathbf{778.9}\;[718.5,\;813.5]$ \\
    \bottomrule
    \end{tabular}
\end{table}

\FloatBarrier
\subsection{CartPole}
\subsubsection{Scheduler Search}
\label{app:cartpole-scheduler-search}

\begin{figure}[h]
    \centering
    \includegraphics[width=0.9\linewidth]{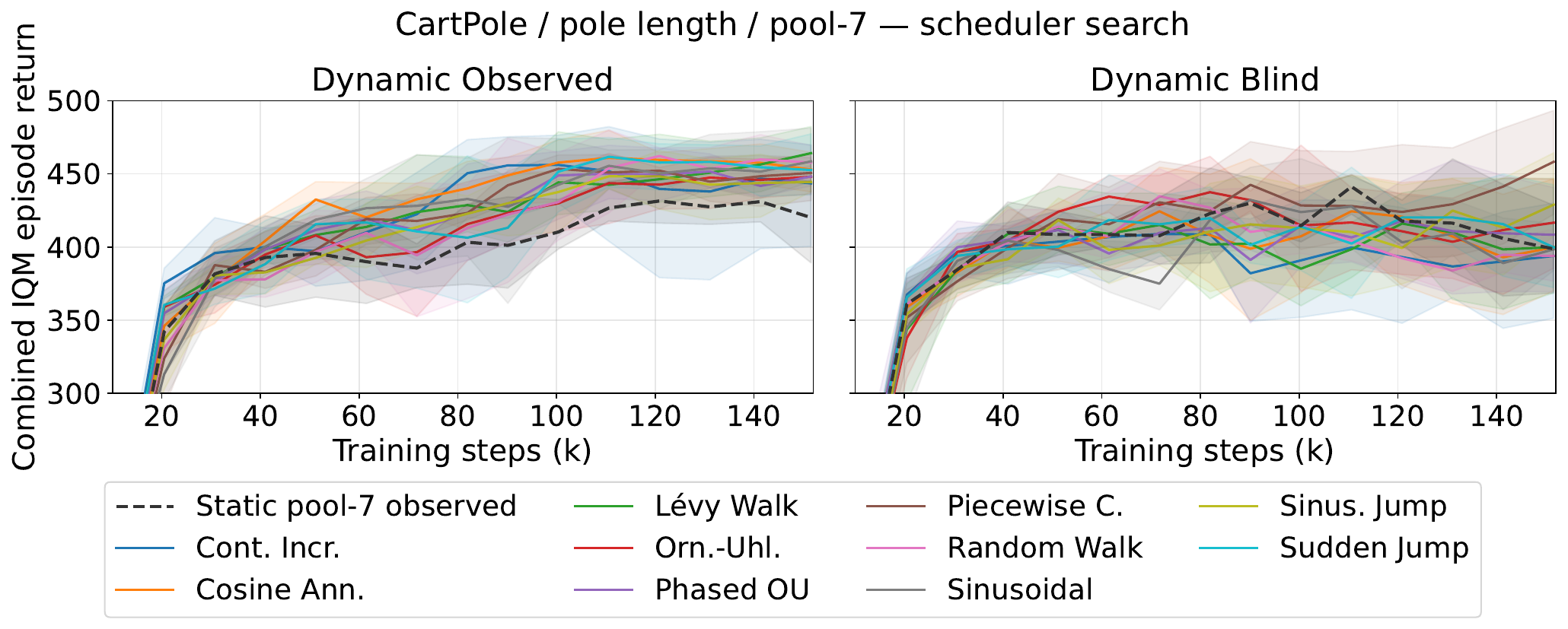}
    \caption{CartPole / pole length / pool-7 — combined IQM training progression per scheduler family (10 seeds). Each line shows the best representative configuration for that family, selected by final combined IQM. Shaded bands indicate Q1--Q3 across seeds. The dashed line marks the static pool-7 baseline.}
    \label{fig:cartpole-scheduler-progression}
\end{figure}

Figure~\ref{fig:cartpole-scheduler-progression} shows the full combined IQM training progression for all scheduler families on CartPole pole length (pool-7), with each family represented by its best-performing configuration selected by final checkpoint combined IQM.

\FloatBarrier
\subsubsection{Dynamic vs.\ Static: Coverage and Score}
\label{app:coverage-scatter}

To directly compare dynamic and static schedulers on coverage independently of pool size effects, we use all sequential tracking runs that include representative dynamic families: identity (static baseline), sinusoidal, Lévy walk, and sudden jump, across all available pool sizes (10 seeds per condition).
This selection deliberately spans scheduler families with qualitatively different temporal structure — smooth oscillation, heavy-tailed jumps, and abrupt discrete displacements — while avoiding any configuration tied to the scheduler search or the pool size study.

The 4D coverage metric discretises CartPole's four observation dimensions into $12$ bins each ($12^4 = 20{,}736$ cells), with boundaries set to physically plausible ranges: cart position $[-2.5, 2.5]$\,m, cart velocity $[-3.5, 3.5]$\,m/s, pole angle $[-0.30, 0.30]$\,rad, and pole angular velocity $[-4.0, 4.0]$\,rad/s.

\begin{figure}[h]
    \centering
    \includegraphics[width=0.6\linewidth]{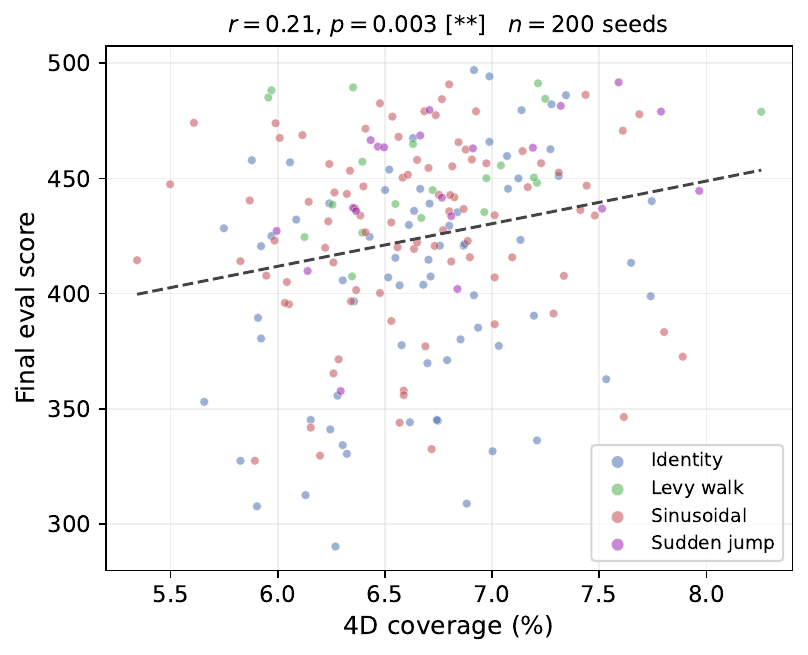}
    \caption{4D state-space coverage vs.\ final eval score across all sequential seeds (pool sizes 3 and 7 for dynamic schedulers, pool sizes 1--61 for identity), coloured by scheduler. Dynamic and static seeds are interleaved throughout the coverage axis. The dashed line shows the least-squares fit ($r = 0.21$).}
    \label{fig:cartpole-coverage-scatter}
\end{figure}

Figure~\ref{fig:cartpole-coverage-scatter} plots coverage against final eval score for each seed.
No scheduler family occupies a systematically higher or lower region of the coverage axis: dynamic and static seeds are interleaved throughout.
The Pearson correlation is $r = 0.21$ ($p = 0.003$), indicating a negligible linear relationship between coverage and score — knowing how much of the state space a scheduler visits tells you almost nothing about how well it generalises.

\subsubsection{Pool Size Scaling}
\label{app:poolsize}

\begin{figure}[t]
  \centering
  \includegraphics[width=0.85\linewidth]{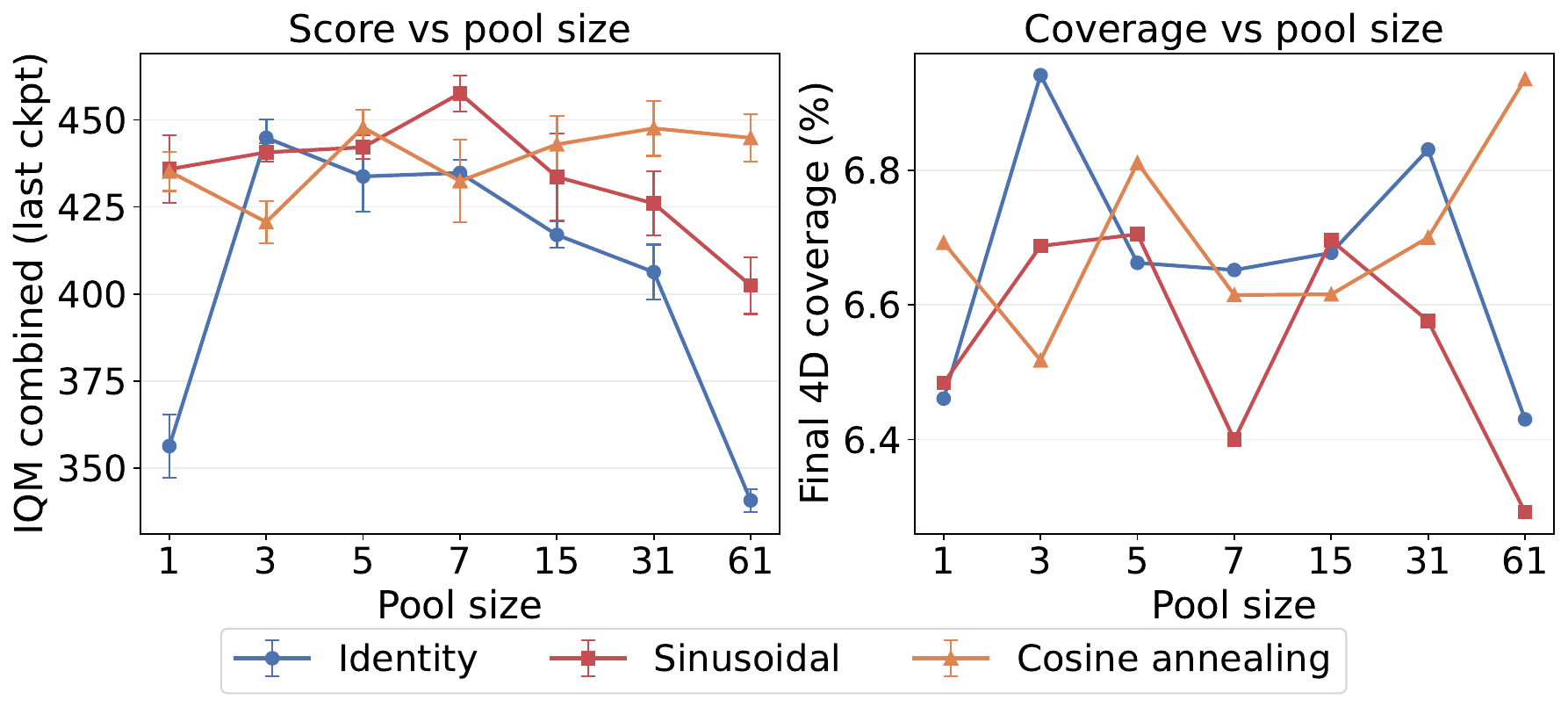}
  \caption{Last-checkpoint IQM score (A) and final 4D coverage (B) vs.\ pool size for CartPole pole-length (10 seeds). Error bars show SEM.}
  \label{fig:poolscaling_score-coverage}
\end{figure}

\begin{figure}[t]
  \centering
  \includegraphics[width=0.85\linewidth]{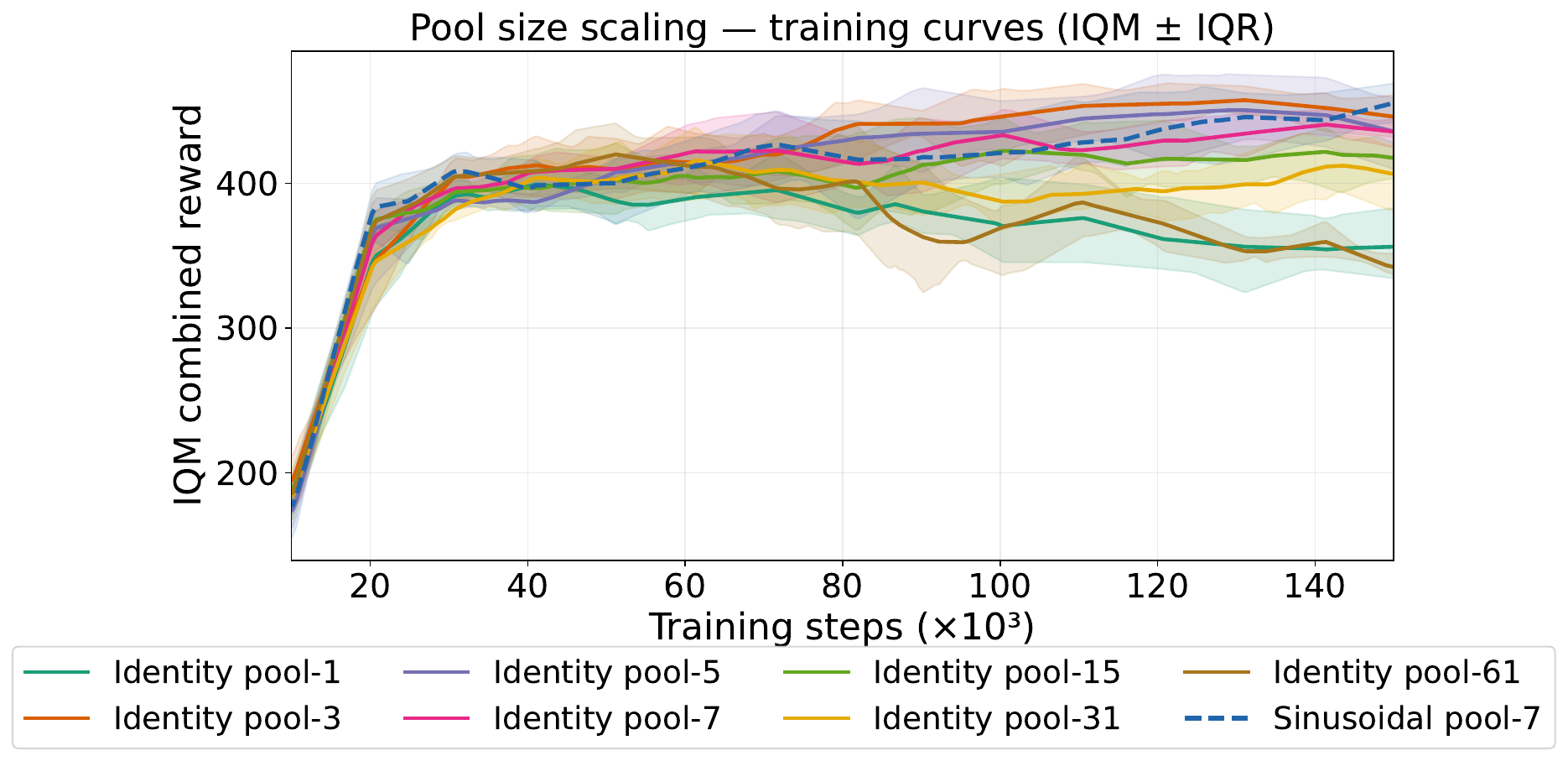}
  \caption{IQM training curves (shading: Q1--Q3) for identity pools~1--61; sinusoidal pool-7 shown dashed for reference.}
  \label{fig:poolscaling_score-curves}
\end{figure}

We vary the number of training contexts from 1 to 61 for the static identity baseline and two dynamic schedulers (sinusoidal, cosine annealing) on CartPole pole-length, using 10 seeds per condition.
The scheduler hyperparameters used here are fixed reference configurations chosen for the tracking runs and are not the result of the scheduler search described in Section~\ref{subsec:experiments-scheduler-search}; absolute scores for the dynamic conditions should therefore not be compared directly to the best-found schedulers reported elsewhere.
Figure~\ref{fig:poolscaling_score-coverage} shows the last-checkpoint IQM score and final 4D state-space coverage.
The static baseline is sensitive to pool size: it peaks around pool-3 and degrades at both extremes, with pool-1 providing too little diversity and pool-61 diluting training across too many contexts.
Dynamic schedulers are largely unaffected by pool size, remaining competitive from pool-1 upwards.
The training curves in Figure~\ref{fig:poolscaling_score-curves} confirm this: static pool-3 converges fastest among static conditions, while pools~31 and~61 plateau noticeably lower, and sinusoidal pool-7 tracks or exceeds the best static condition throughout.

Crucially, 4D state-space coverage is flat at $\approx\!6.5\%$ across all pool sizes and all schedulers, and Table~\ref{tab:behavioral-breadth} shows that this coverage is concentrated almost entirely in the first 50\,k training steps: stages~2 and~3 contribute less than $1\%$, and virtually no cells are newly discovered after stage~1 ($\leq\!0.03\%$).
Pool size therefore modulates the \emph{distribution} of the training signal across context values, not the breadth of states visited.

\begin{table}[h]
    \centering
    \caption{Fraction of the $12^4$ CartPole hypergrid visited per 50\,k-step training window (mean over 10 seeds, \%). Coverage in stages~2--3 is below $1\%$ for all conditions; virtually no cells are newly discovered after stage~1.}
    \label{tab:behavioral-breadth}
    \setlength{\tabcolsep}{5pt}
    \begin{tabular}{r ccc ccc ccc}
      \toprule
      & \multicolumn{3}{c}{\textbf{Identity}}
      & \multicolumn{3}{c}{\textbf{Sinusoidal}}
      & \multicolumn{3}{c}{\textbf{Cosine annealing}} \\
      \cmidrule(lr){2-4}\cmidrule(lr){5-7}\cmidrule(lr){8-10}
      \textbf{Pool}
        & S1 & S2 & S3
        & S1 & S2 & S3
        & S1 & S2 & S3 \\
      \midrule
       1 & 6.45 & 0.57 & 0.27 & 6.48 & 0.52 & 0.29 & 6.68 & 0.71 & 0.27 \\
       3 & 6.93 & 0.75 & 0.32 & 6.68 & 0.61 & 0.28 & 6.52 & 0.53 & 0.26 \\
       5 & 6.65 & 0.77 & 0.29 & 6.70 & 0.69 & 0.29 & 6.80 & 0.62 & 0.32 \\
       7 & 6.65 & 0.64 & 0.29 & 6.39 & 0.77 & 0.29 & 6.61 & 0.61 & 0.30 \\
      15 & 6.67 & 0.64 & 0.29 & 6.69 & 0.58 & 0.27 & 6.61 & 0.62 & 0.31 \\
      31 & 6.82 & 0.71 & 0.31 & 6.57 & 0.56 & 0.29 & 6.70 & 0.63 & 0.30 \\
      61 & 6.41 & 0.69 & 0.32 & 6.29 & 0.57 & 0.29 & 6.90 & 0.79 & 0.34 \\
      \bottomrule
      \multicolumn{10}{l}{\small S1\,=\,0--50k;\enspace S2\,=\,50--100k;\enspace S3\,=\,100--150k steps.}
    \end{tabular}
\end{table}

\FloatBarrier
\subsubsection{Parallel vs.\ Sequential Training}
\label{app:par-vs-seq}

\begin{figure}[t]
  \centering
  \includegraphics[width=0.9\linewidth]{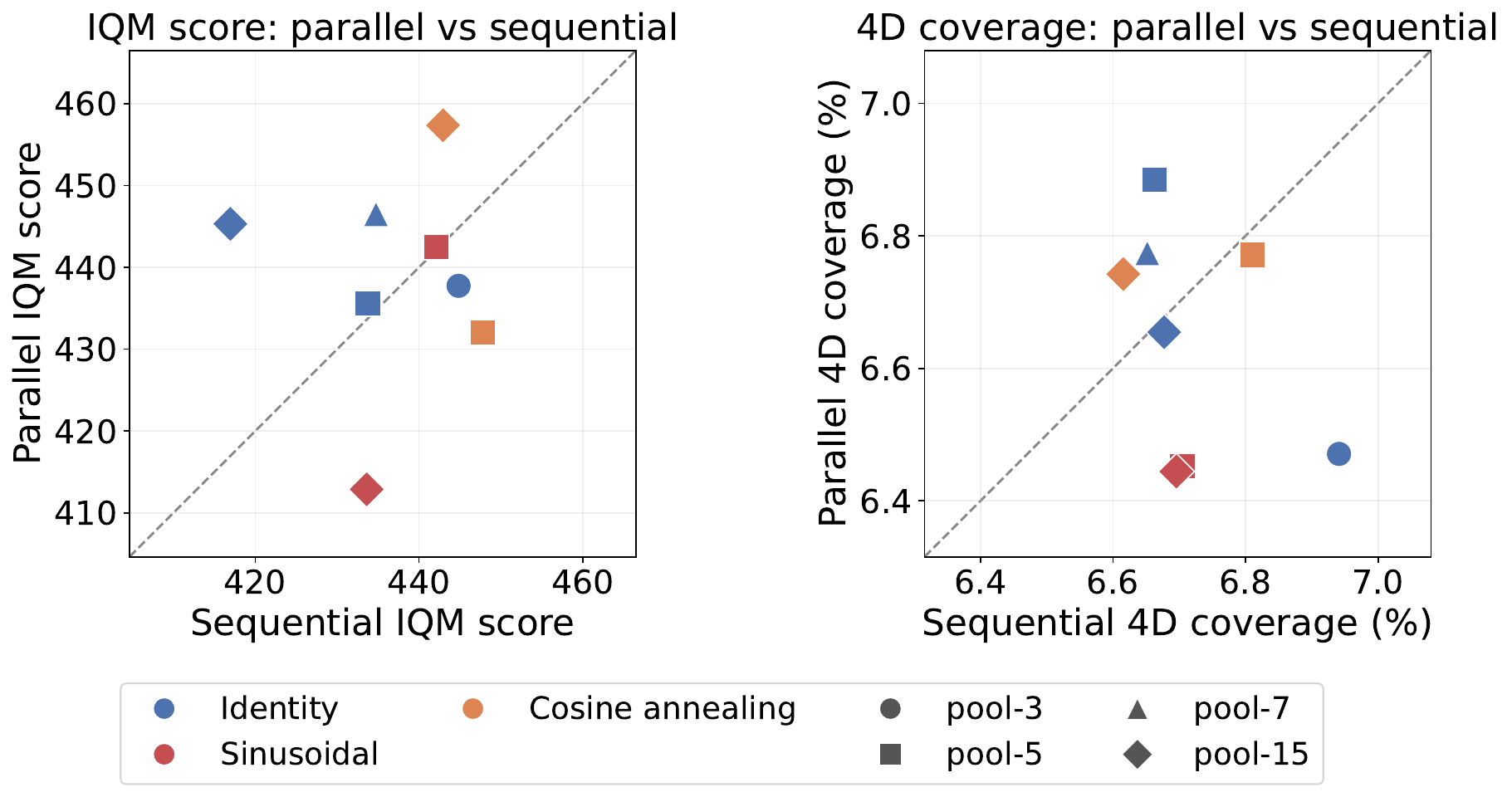}
  \caption{Sequential vs.\ parallel IQM score (A) and 4D coverage (B) for CartPole. Points on the dashed diagonal indicate no difference. Colour encodes scheduler family; marker shape encodes pool size.}
  \label{fig:par-vs-seq}
\end{figure}

We compare running 8 parallel environments against a single sequential environment for identity, sinusoidal, and cosine annealing schedulers at pool sizes 5 and 15 (and identity at pool sizes 3 and 7), using 10 seeds per condition on CartPole pole-length.
Figure~\ref{fig:par-vs-seq} shows the last-checkpoint IQM score and final 4D coverage for each condition, with the diagonal marking equal performance.

Both metrics scatter tightly around the diagonal with no consistent direction (Table~\ref{tab:par-vs-seq}): score differences range from $-21$ to $+28$ IQM with no pattern across schedulers or pool sizes, and coverage differences are below $0.5\%$ in all cases.
The summed deltas across all eight conditions are $+13.1$ IQM and $-0.56\%$ coverage, both negligible.
Parallel training neither expands state-space coverage nor reliably improves generalisation.

This result is informative beyond the numerical statement.
Parallel environments, typical for PPO, introduce context diversity \emph{across} simultaneous rollouts, enriching each policy gradient update with transitions from multiple dynamics regimes at once.
Dynamic scheduling, by contrast, introduces context variation \emph{within} a single episode, exposing the policy to a structured temporal sequence of dynamics changes during one rollout.
The fact that the former does not replicate the gains of the latter suggests that the benefit of dynamic scheduling is not simply a consequence of seeing more diverse contexts per update, but is tied to the intra-episode temporal structure of the training signal itself.

\begin{table}[t]
  \centering
  \caption{Score and coverage difference (parallel $-$ sequential) per condition. Positive = parallel preferred; negative = sequential preferred.}
  \label{tab:par-vs-seq}
  \setlength{\tabcolsep}{6pt}
  \begin{tabular}{llrr}
    \toprule
    \textbf{Scheduler} & \textbf{Pool} & $\Delta$ \textbf{IQM score} & $\Delta$ \textbf{Coverage} \\
    \midrule
    Identity         &  3 & $-7.1$  & $-0.47\%$ \\
    Identity         &  5 & $+1.8$  & $+0.22\%$ \\
    Identity         &  7 & $+11.8$ & $+0.12\%$ \\
    Identity         & 15 & $+28.3$ & $-0.02\%$ \\
    Sinusoidal       &  5 & $+0.4$  & $-0.25\%$ \\
    Sinusoidal       & 15 & $-20.7$ & $-0.25\%$ \\
    Cosine annealing &  5 & $-15.8$ & $-0.04\%$ \\
    Cosine annealing & 15 & $+14.4$ & $+0.13\%$ \\
    \midrule
    \multicolumn{2}{l}{Sum} & $+13.1$ & $-0.56\%$ \\
    \multicolumn{2}{l}{Parallel preferred} & 4/8 & 3/8 \\
    \bottomrule
  \end{tabular}
\end{table}

\FloatBarrier
\subsubsection{BipedalWalker State-Space Coverage}
\label{app:walker-coverage}

\begin{figure}[t]
  \centering
  \includegraphics[width=\linewidth]{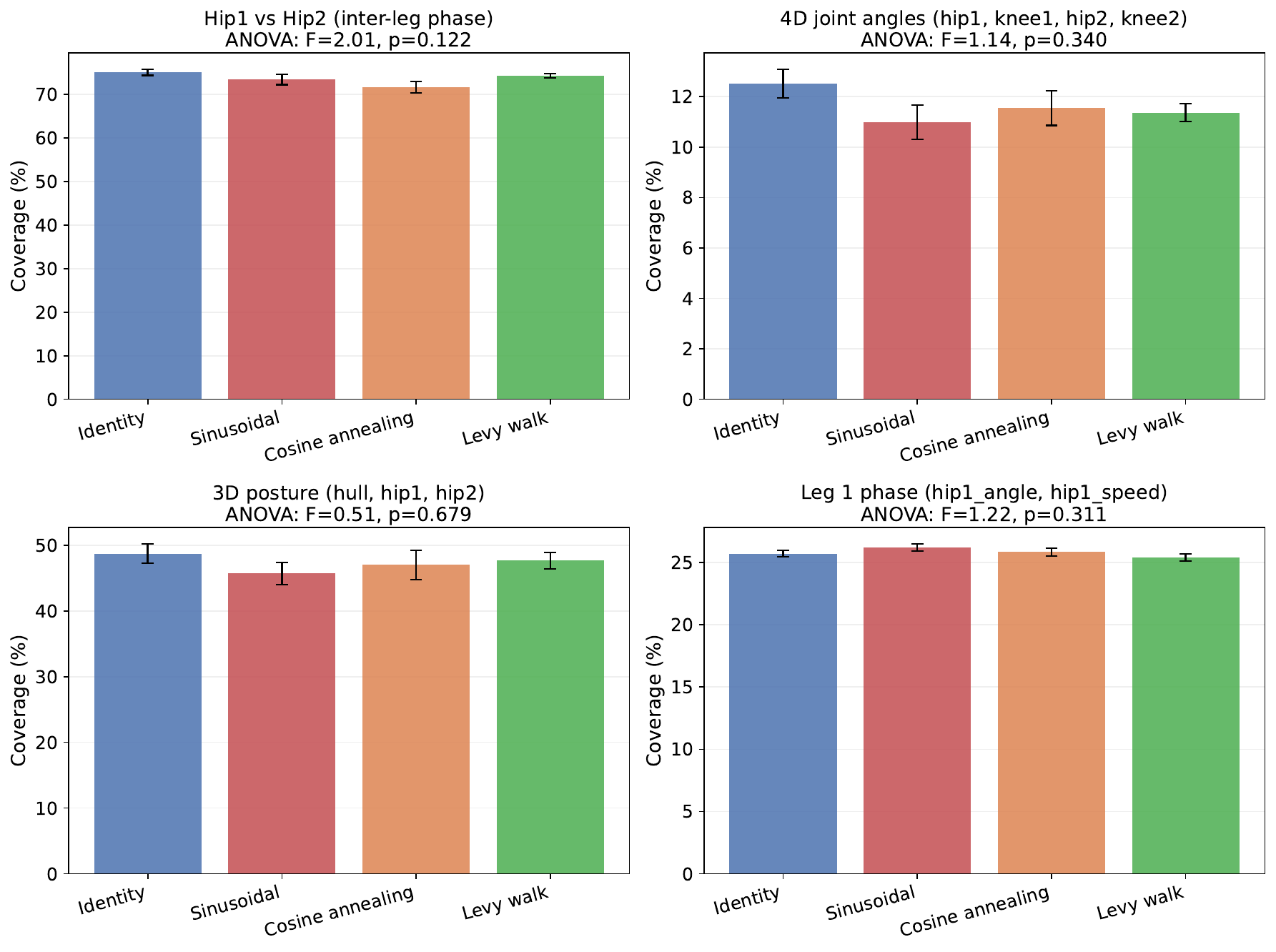}
  \caption{BipedalWalker full-run coverage per scheduler across four state-space projections (4 seeds $\times$ 4 context ranges, aggregated). \textbf{A}: inter-leg phase ($30^2$, $p=0.12$). \textbf{B}: 4D joint angles ($12^4$, $p=0.34$). \textbf{C}: 3D posture ($15^3$, $p=0.68$). \textbf{D}: leg-1 phase ($30^2$, $p=0.31$).}
  \label{fig:walker-coverage-bars}
\end{figure}

We extend the coverage analysis to BipedalWalker (COM\_X context, 4 seeds per pool size, pool sizes 1--15, sequential only) using four complementary projections of the 24-dimensional observation space.
For each projection we compute the fraction of discretised cells visited over 1\,M training steps and compare across scheduler families (identity, sinusoidal, cosine annealing, Lévy walk).
No projection yields a significant difference across schedulers (Figure~\ref{fig:walker-coverage-bars}), consistent with the CartPole finding.

Table~\ref{tab:walker-coverage-score} reports full-run coverage across all four projections alongside mean training return (averaged over pool sizes; note this is rollout return, not the IQM evaluation metric used in the main text).
None of the per-projection ANOVAs are significant.
Across projections, the static identity baseline consistently ranks highest on coverage and lowest on return, while sinusoidal ranks lowest on coverage and highest on return.
Given the limited number of seeds, this trend is suggestive rather than conclusive; the broader state-space spread of the static baseline may reflect slower convergence rather than active exploration.

\begin{figure}[t]
  \centering
  \includegraphics[width=\linewidth]{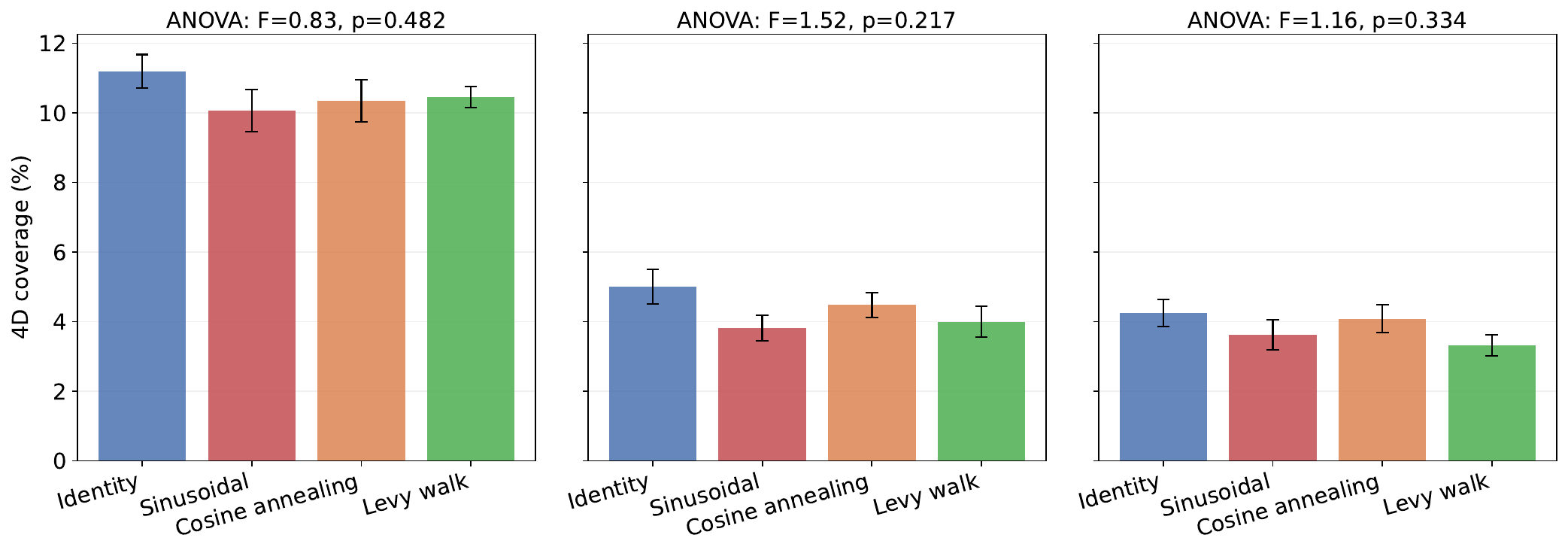}
  \caption{BipedalWalker 4D joint-space coverage per scheduler across three equal training stages ($\approx\!333$\,k steps each). Shared $y$-axis shows the progressive contraction of the visited state space. ANOVA $p > 0.05$ for all stages.}
  \label{fig:walker-coverage-stages}
\end{figure}

Figure~\ref{fig:walker-coverage-stages} breaks 4D joint coverage into three equal training stages (0--333k, 333--667k, 667k--1\,M steps) with a shared $y$-axis.
All schedulers drop from $\approx\!10$--$13\%$ in stage~1 to $\approx\!3$--$5\%$ by stage~3, confirming that BipedalWalker policies also converge to a narrow behavioral manifold well before the end of training, regardless of scheduler.

\begin{table}[t]
  \centering
  \caption{Full-run coverage (\%) per scheduler across four projections and mean rollout return, averaged over pool sizes 1, 3, 7, 15 (4 seeds each). Bold marks the highest value per column.}
  \label{tab:walker-coverage-score}
  \setlength{\tabcolsep}{5pt}
  \begin{tabular}{lrrrrc}
    \toprule
    & \multicolumn{4}{c}{\textbf{Coverage (\%)}} & \\
    \cmidrule(lr){2-5}
    \textbf{Scheduler}
      & \textbf{A: inter-leg}
      & \textbf{B: 4D joints}
      & \textbf{C: posture}
      & \textbf{D: leg phase}
      & \textbf{Return} \\
    \midrule
    Identity         & \textbf{75.1} & \textbf{12.5} & \textbf{48.7} & 25.7          & 142 \\
    Cosine annealing & 71.7          & 11.6          & 47.0          & 25.8          & 150 \\
    Lévy walk        & 74.3          & 11.4          & 47.7          & 25.4          & 163 \\
    Sinusoidal       & 73.5          & 11.0          & 45.7          & \textbf{26.2} & \textbf{177} \\
    \midrule
    ANOVA $p$        & 0.12          & 0.34          & 0.68          & 0.31          & — \\
    \bottomrule
    \multicolumn{6}{l}{\small A: hip$_1$ vs hip$_2$ ($30^2$);\enspace B: hip$_1$, knee$_1$, hip$_2$, knee$_2$ ($12^4$);\enspace C: hull, hip$_1$, hip$_2$ ($15^3$);\enspace D: leg-1 phase ($30^2$).}
  \end{tabular}
\end{table}

\FloatBarrier

\subsubsection{Context Normalization Ablation}
\label{app:normalization}

The figure below support the normalization design choice discussed in Section~\ref{subsec:wrapper}.
Results are from a preliminary subset of schedule families; the effect is consistent across conditions.

\begin{figure}[ht!]
  \centering
  \begin{subfigure}[b]{0.48\linewidth}
    \centering
    \includegraphics[width=\linewidth]{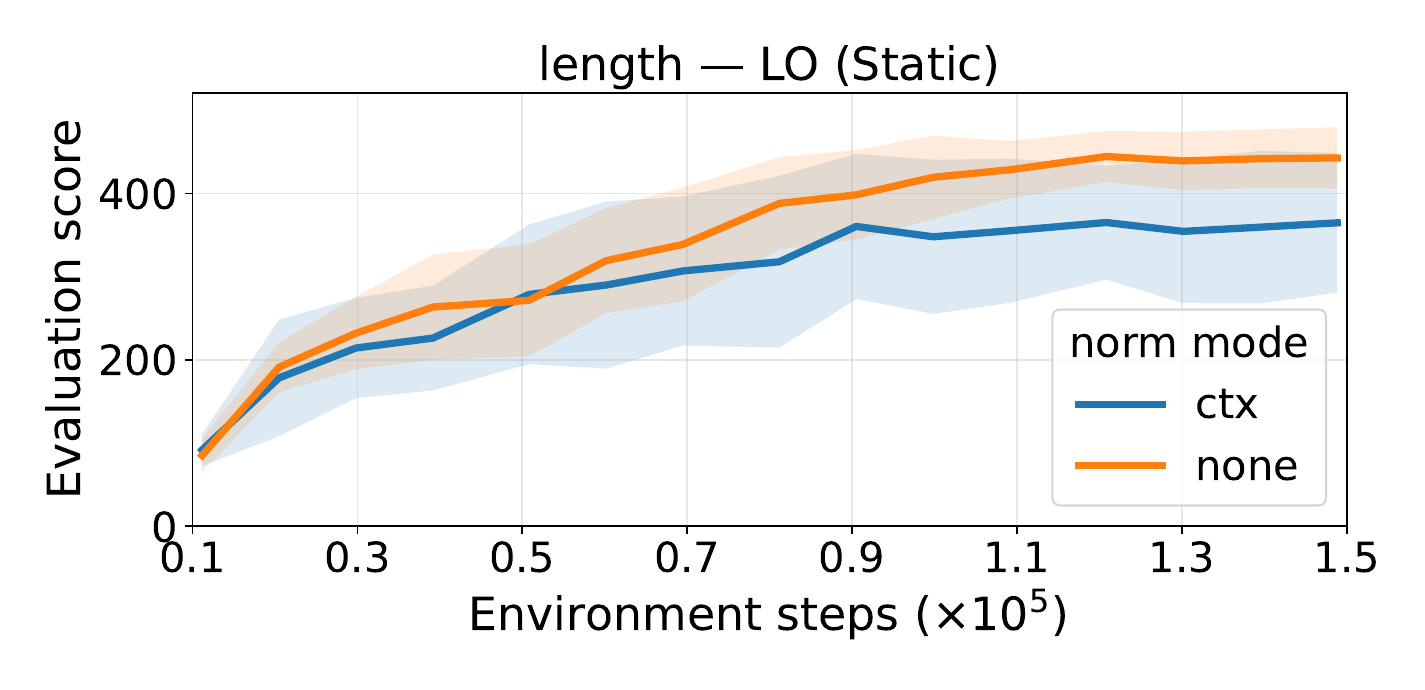}
    \caption{Static training --- low pole length.}
  \end{subfigure}
  \hfill
  \begin{subfigure}[b]{0.48\linewidth}
    \centering
    \includegraphics[width=\linewidth]{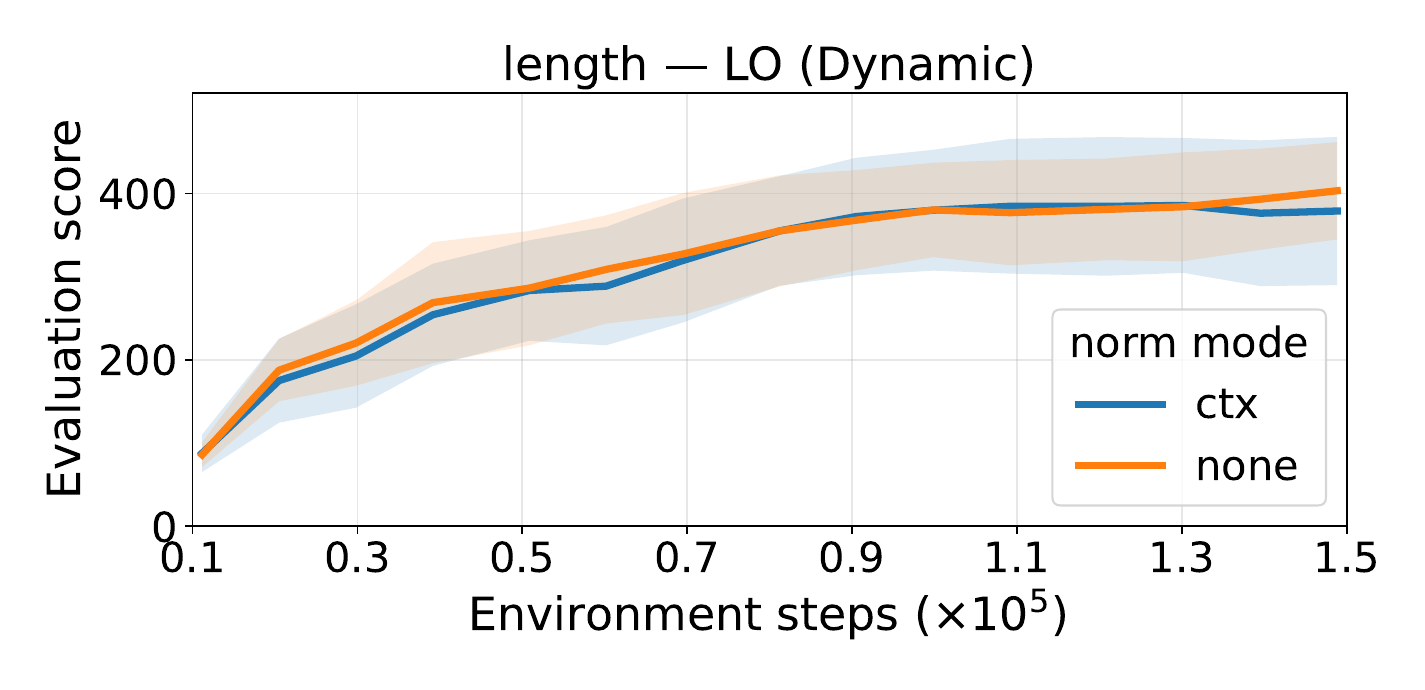}
    \caption{Dynamic training --- low pole length.}
  \end{subfigure}
  \vspace{0.75em}
  \begin{subfigure}[b]{0.48\linewidth}
    \centering
    \includegraphics[width=\linewidth]{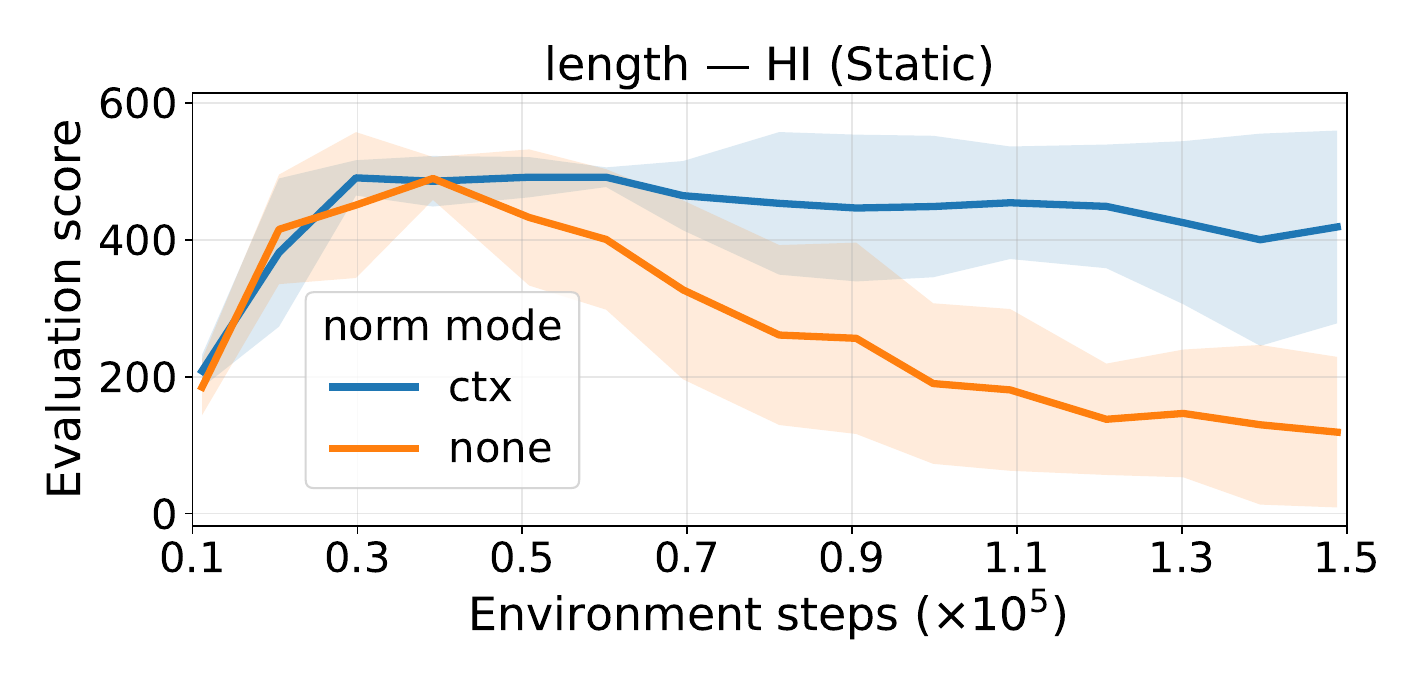}
    \caption{Static training --- high pole length.}
  \end{subfigure}
  \hfill
  \begin{subfigure}[b]{0.48\linewidth}
    \centering
    \includegraphics[width=\linewidth]{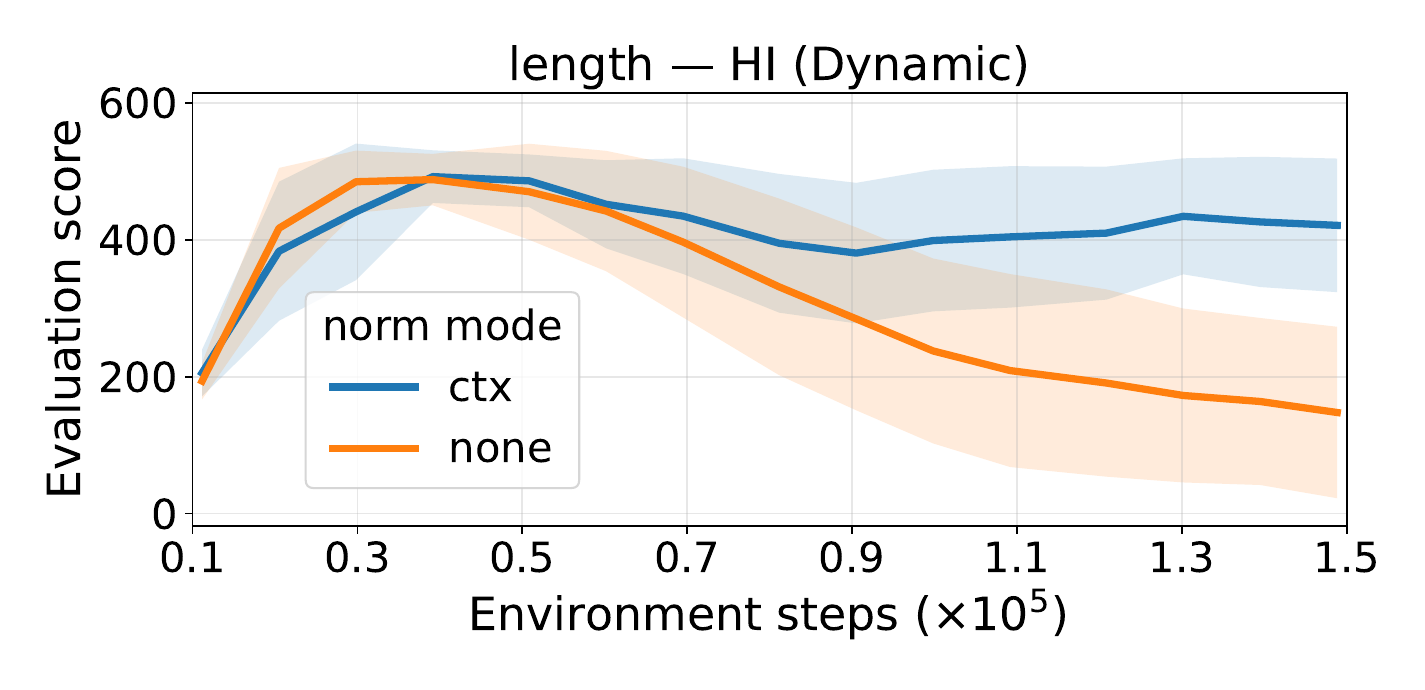}
    \caption{Dynamic training --- high pole length.}
  \end{subfigure}
  \caption{
    Effect of context normalization on static and dynamic PPO training (CartPole, pole length context).
    The dynamic curves aggregate performance across a preliminary set of non-stationary schedules.
    Lines show mean evaluation score across seeds; shaded regions indicate standard deviation.
    }
  \label{fig:app-context-norm}
\end{figure}

\FloatBarrier
\subsection{Additional Bar Plots per Eval Context}
\label{app:add-bar-plots}

\paragraph{CartPole.}
The last-checkpoint bar plot appears in the main paper (Figure~\ref{fig:cartpole-bars}); here we include the best-anytime companion.
All conditions perform strongly as the ID region saturates at the maximum reward of 500.
The meaningful spread is in OOD-low, where static observed is the weakest (IQM 313 at last checkpoint) and dynamic blind the strongest (IQM 456), with static blind and dynamic observed in between.
Dynamic observed leads at the last checkpoint (combined IQM 464) while dynamic blind leads at best-anytime (488); the two are close throughout, suggesting that even a blind scheduler captures most of the benefit of temporal context variation on this environment.

\begin{figure}[t]
    \centering
    \includegraphics[width=\linewidth]{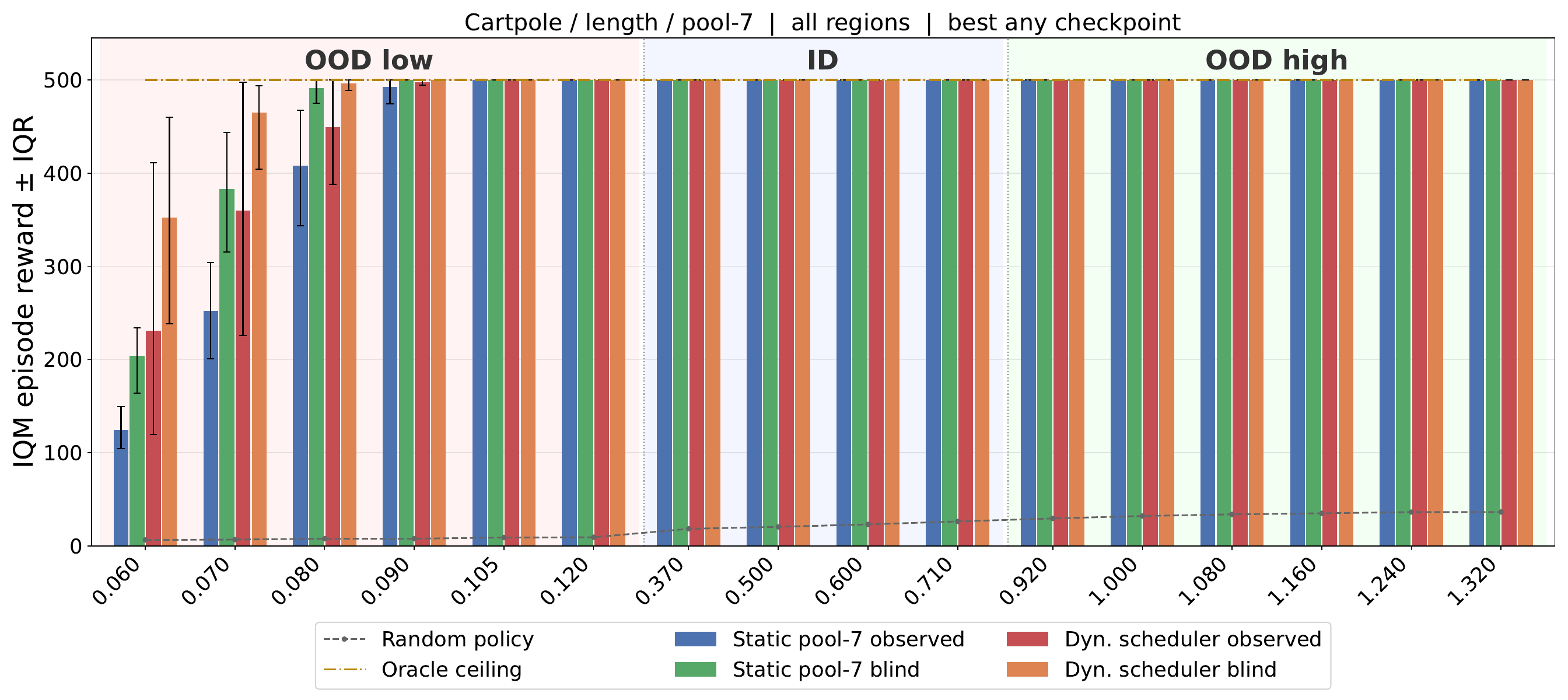}
    \caption{CartPole combined IQM (best anytime, pool-7). Per-scheduler values in Table~\ref{tab:ext_cartpole}.}
    \label{fig:cartpole-bar-combined-best}
\end{figure}

\begin{figure}[t]
    \centering
    \includegraphics[width=\linewidth]{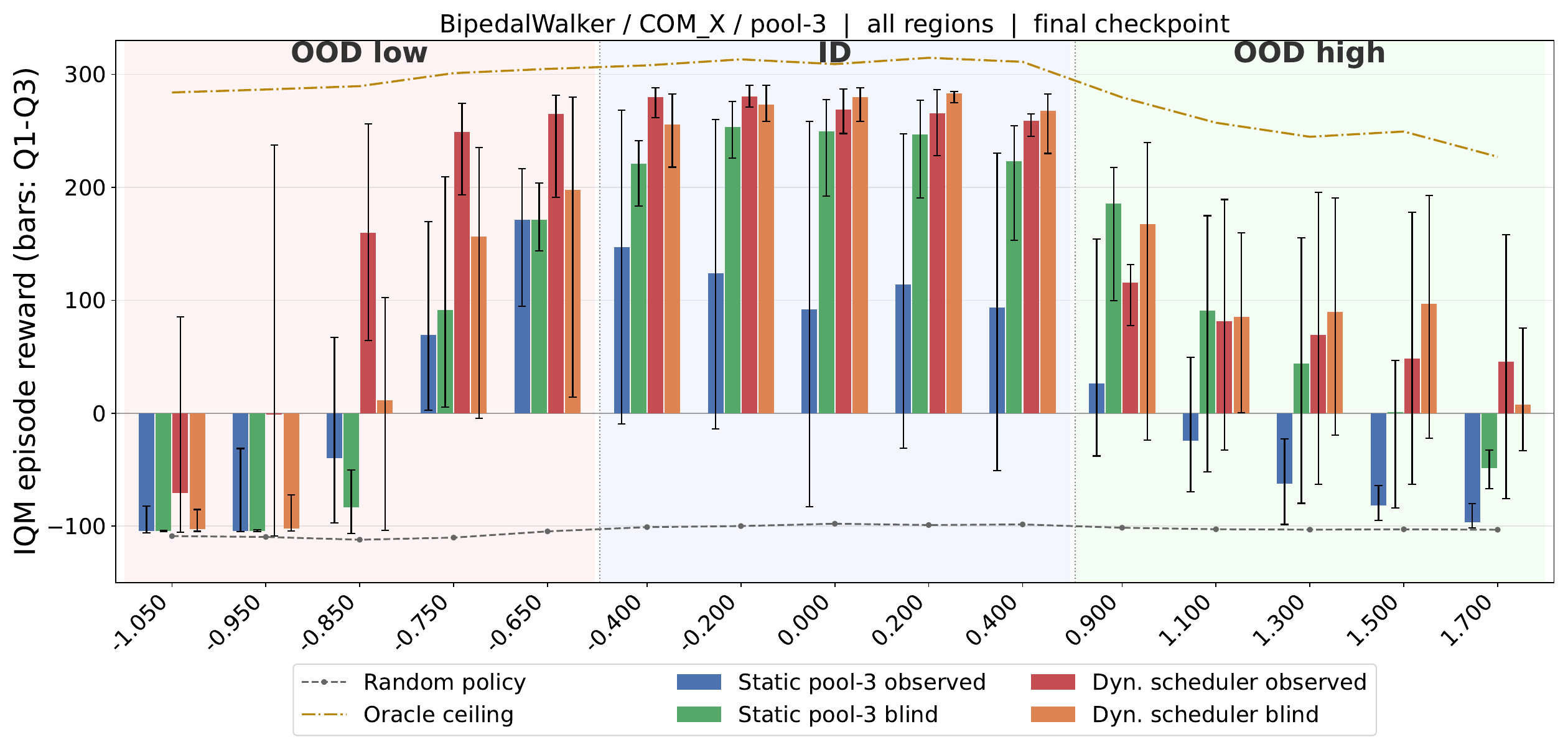}
    \caption{BipedalWalker combined IQM (last checkpoint, pool-3). Per-scheduler values in Table~\ref{tab:ext_walker}.}
    \label{fig:walker-bar-combined}
\end{figure}

\paragraph{BipedalWalker.}
Figure~\ref{fig:walker-bar-combined} shows per-context IQM at the last checkpoint (pool-3).
The static observed baseline achieves strong in-distribution performance (IQM 114) but collapses on OOD contexts (IQM $-48$ high, $-10$ low), pulling its combined IQM to only 31: with three fixed training contexts the policy overfits to those specific dynamics and fails to generalise.
Dynamic schedulers substantially outperform both static conditions; even dynamic blind comfortably beats static blind, confirming that intra-episode temporal structure adds value beyond na\"ive domain randomisation.

\begin{figure}[t]
    \centering
    \includegraphics[width=\linewidth]{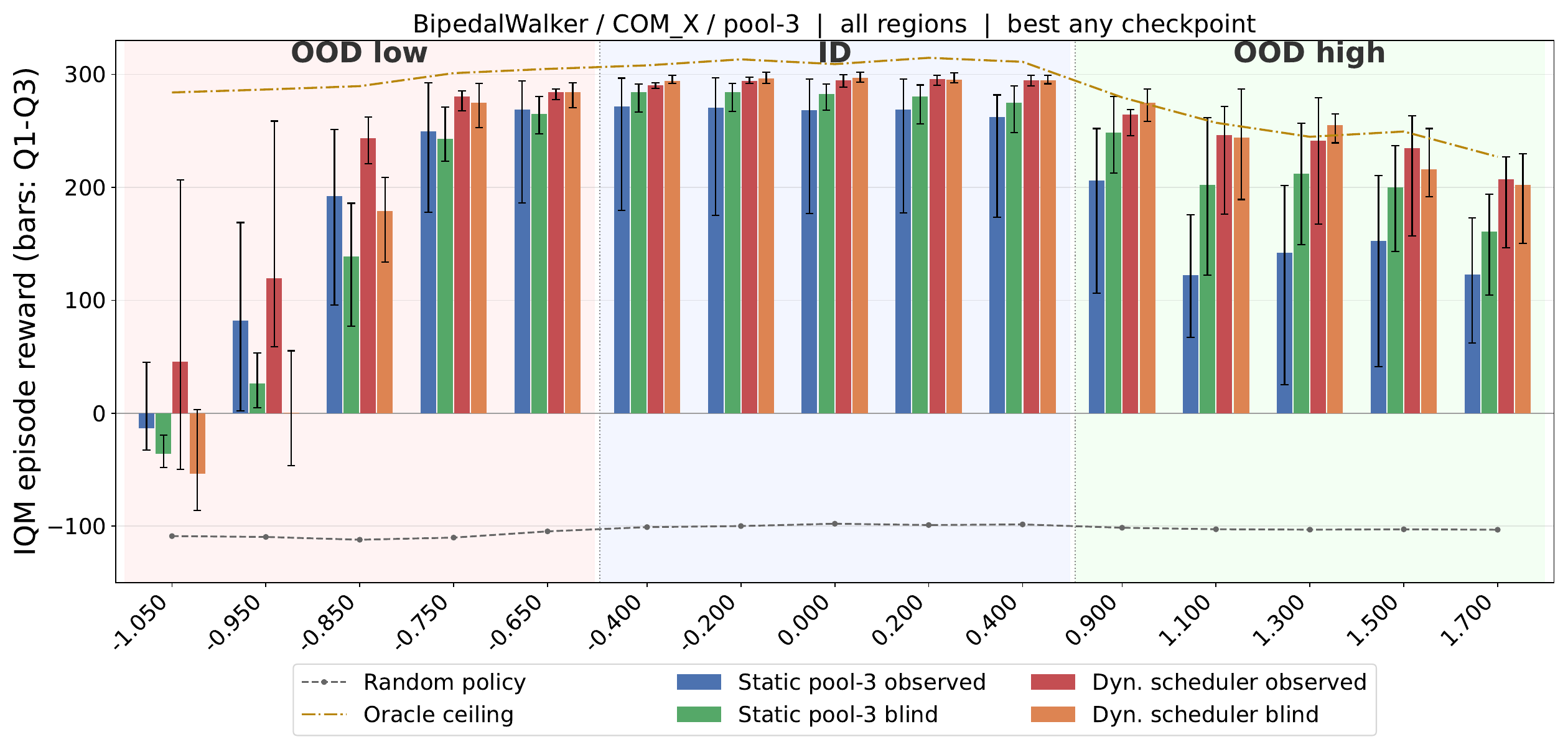}
    \caption{BipedalWalker combined IQM (best anytime, pool-3). The gap relative to the last-checkpoint figure above is most pronounced for the static observed baseline, which collapses late in training.}
    \label{fig:walker-bar-combined-best}
\end{figure}

\paragraph{CarRacing.}
Context-observed conditions exhibit significant late-training policy degradation on this environment; we therefore show both the last-checkpoint and best-anytime figures side by side for each axis.
At the last checkpoint, dynamic blind is the strongest condition on both axes.
At best-anytime, static blind leads on COM\_X (IQM 842.9 vs.\ 803.7 for dynamic blind) while dynamic blind leads on COM\_Y (760.7 vs.\ 636.4 for static blind); all observed conditions struggle to stably exploit the live context signal through the CNN pipeline.
Per-scheduler IQM values are in Tables~\ref{tab:ext_carracing_comx} and~\ref{tab:ext_carracing_comy}.

\begin{figure}[t]
    \centering
    \includegraphics[width=\linewidth]{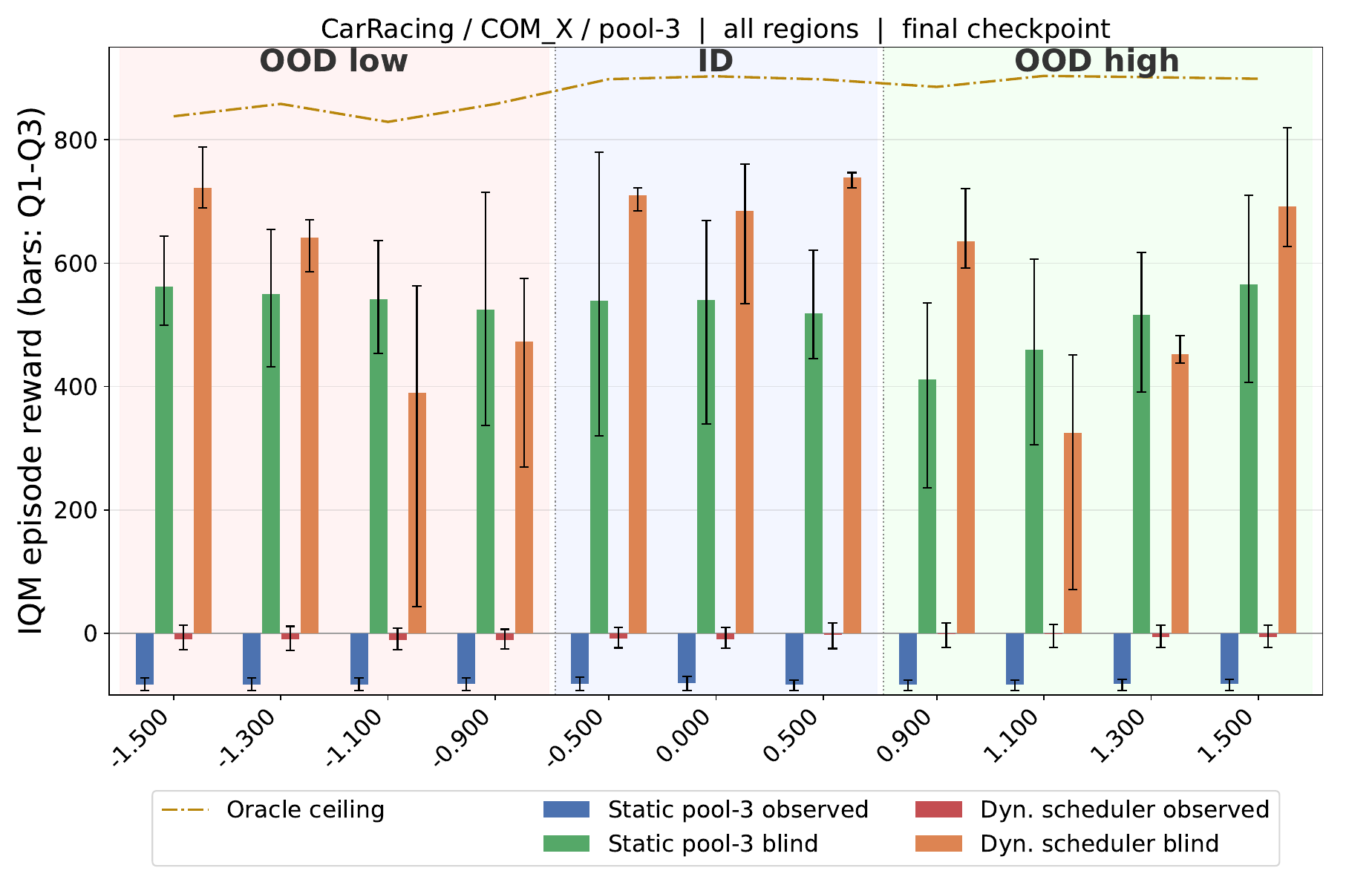}
    \caption{CarRacing COM\_X last-checkpoint IQM (pool-3).}
    \label{fig:vehiclerace-bar-combined-com_x_last}
\end{figure}

\begin{figure}[t]
    \centering
    \includegraphics[width=\linewidth]{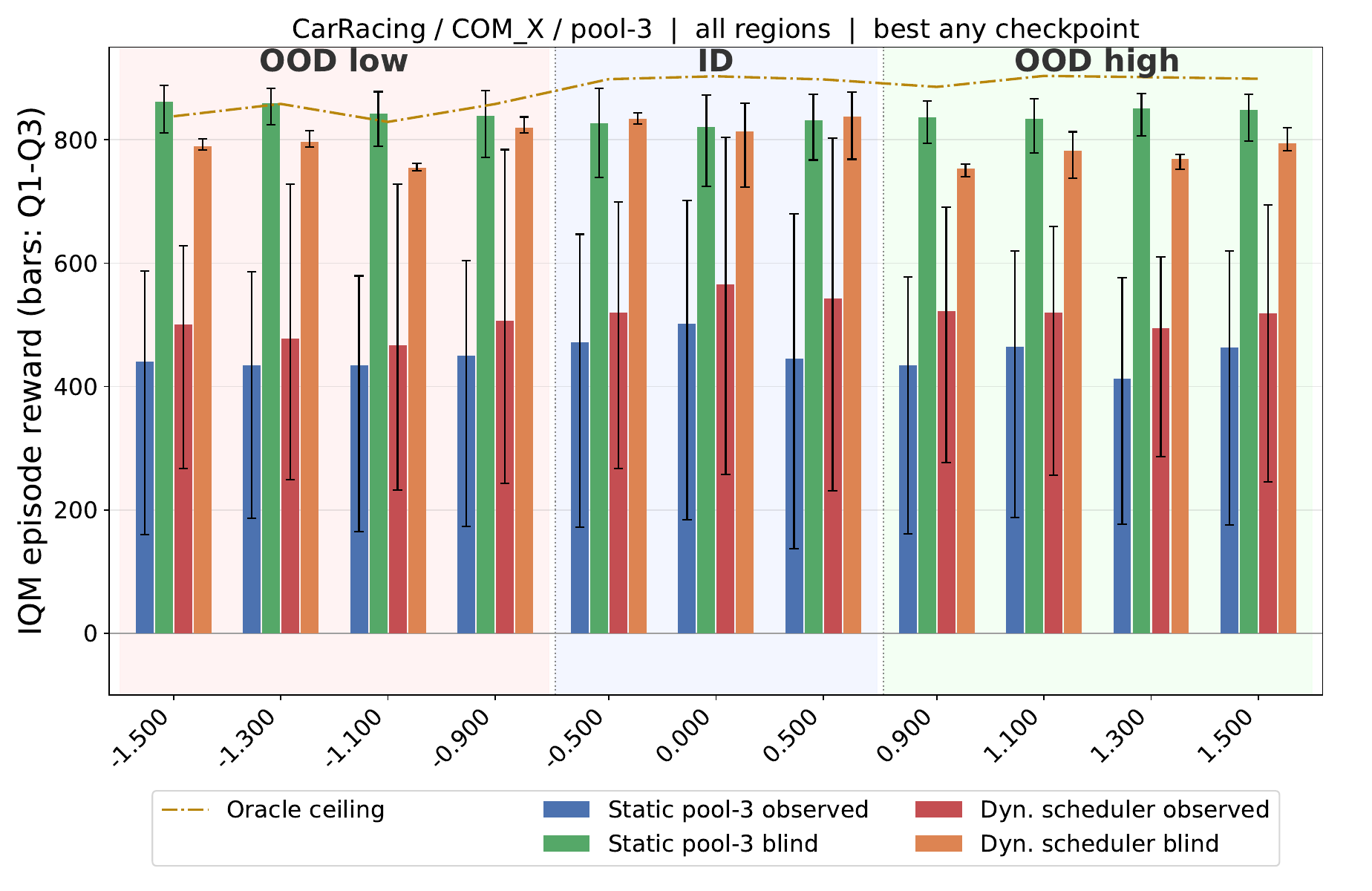}
    \caption{CarRacing COM\_X best-anytime IQM (pool-3).}
    \label{fig:vehiclerace-bar-combined-com_x_any-best}
\end{figure}

\begin{figure}[t]
    \centering
    \includegraphics[width=0.9\linewidth]{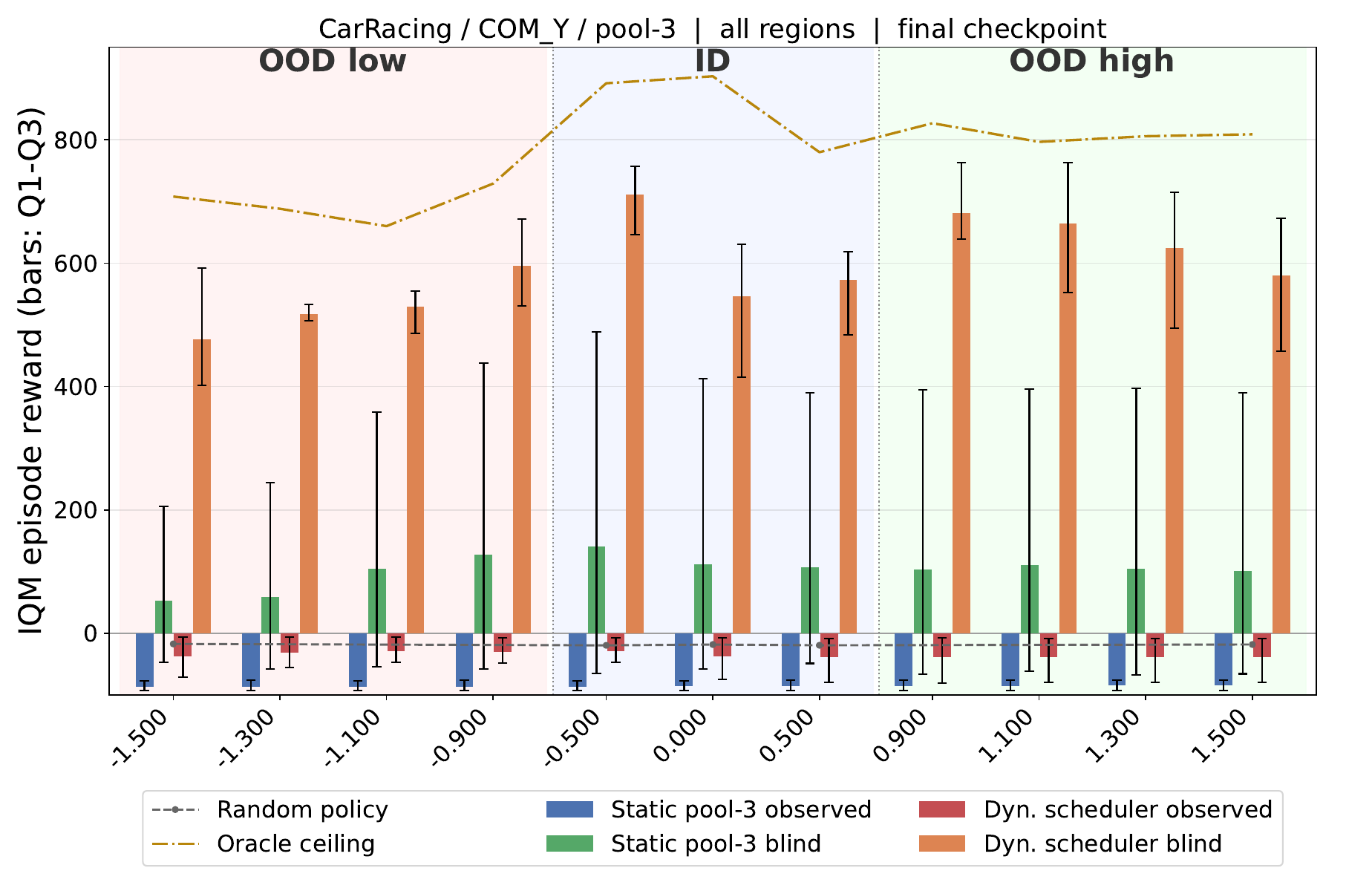}
    \caption{CarRacing COM\_Y last-checkpoint IQM (pool-3).}
    \label{fig:vehiclerace-bar-combined-com_y_last}
\end{figure}

\begin{figure}[t]
    \centering
    \includegraphics[width=0.9\linewidth]{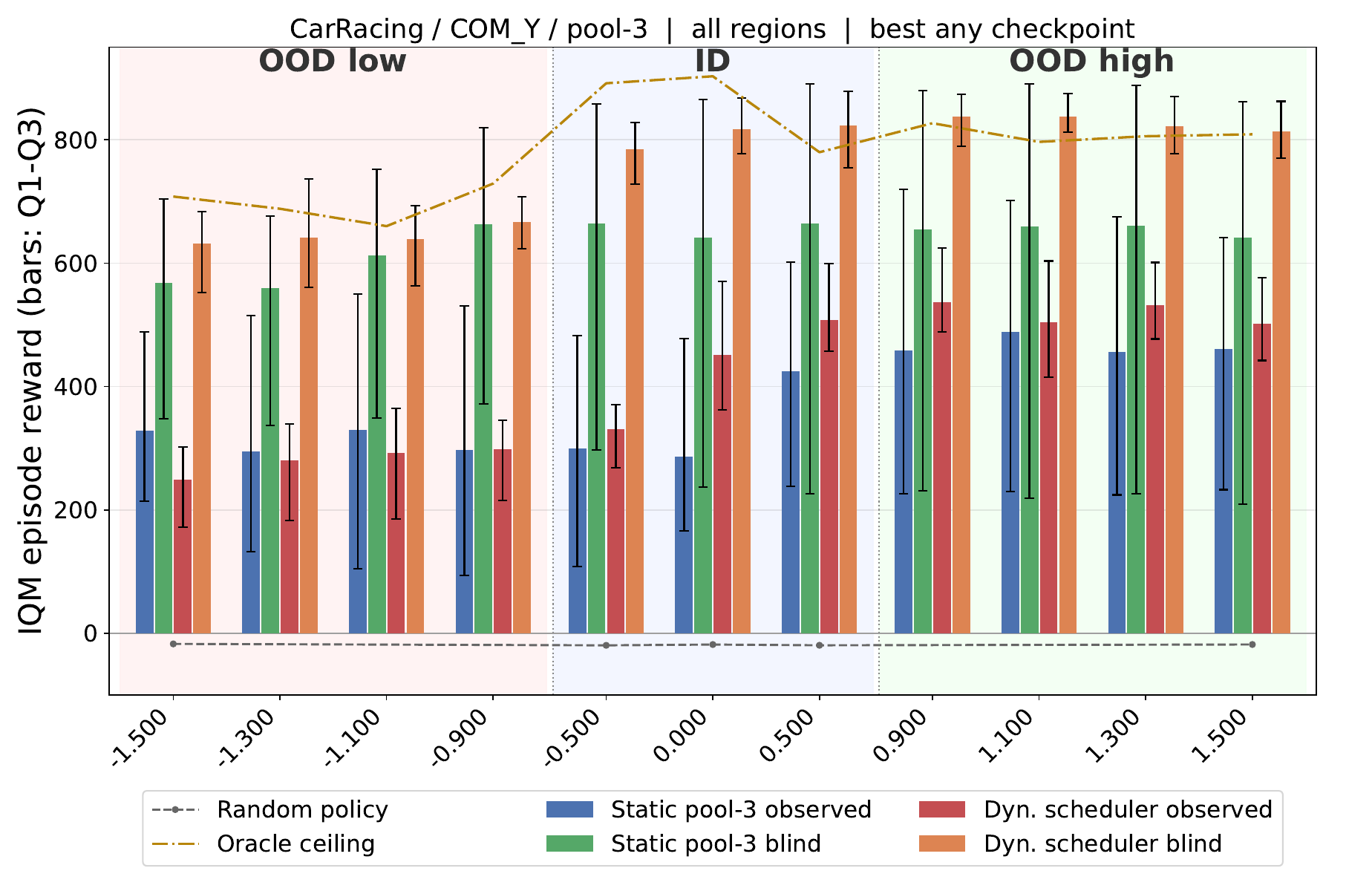}
    \caption{CarRacing COM\_Y best-anytime IQM (pool-3).}
    \label{fig:vehiclerace-bar-combined-com_y_any-best}
\end{figure}

\FloatBarrier
\section{Multi-stage Schedulers Additional Results}
\label{app:multistage-additional}

\subsection{Best-Anytime Results}
\label{app:optuna-best-anytime}

Table~\ref{tab:optuna_retrain_best} complements Table~\ref{tab:optuna_retrain} with best-anytime IQM scores — the highest combined evaluation score achieved at any stage boundary during training.
Walker values remain at the matched 1.5\,M-step budget.

\begin{table}[h]
  \centering
  \caption{Multi-stage Optuna retrain results at the \textbf{best-anytime checkpoint}.
    Walker values at the matched 1.5\,M-step budget.
    Values: IQM\,[Q1,\,Q3]. Top-3 retrained trials shown per environment.}
  \label{tab:optuna_retrain_best}
  \setlength{\tabcolsep}{5pt}
  \begin{tabular}{lcc}
    \toprule
    \textbf{Condition} & \textbf{CartPole} & \textbf{Walker (1.5\,M)} \\
    \midrule
    Static observed       & $449.7\;[440.3,\;462.8]$ & $143.6\;[114.5,\;167.9]$ \\
    Static blind          & $463.4\;[453.4,\;474.0]$ & $136.3\;[101.8,\;160.4]$ \\
    \midrule
    Best sched.\ observed & $\textbf{479.6}\;[469.3,\;491.0]$ & $\textbf{176.9}\;[135.6,\;197.0]$ \\
    Best sched.\ blind    & $479.3\;[465.5,\;495.9]$ & $140.3\;[108.7,\;164.9]$ \\
    \midrule
    Optuna \#1 (T168 / T118) & $470.1\;[450.9,\;492.1]$ & $171.4\;[148.6,\;196.2]$ \\
    Optuna \#2 (T218 / T117) & $469.3\;[447.4,\;482.9]$ & $165.2\;[149.2,\;177.3]$ \\
    Optuna \#3 (T145 / T066) & $477.6\;[458.2,\;492.6]$ & $128.8\;[111.7,\;150.1]$ \\
    \bottomrule
  \end{tabular}
\end{table}

\FloatBarrier
\subsection{Optuna Results}
\label{app:cartpole-optuna}

\begin{figure}[ht]
    \centering
    \includegraphics[width=0.8\linewidth]{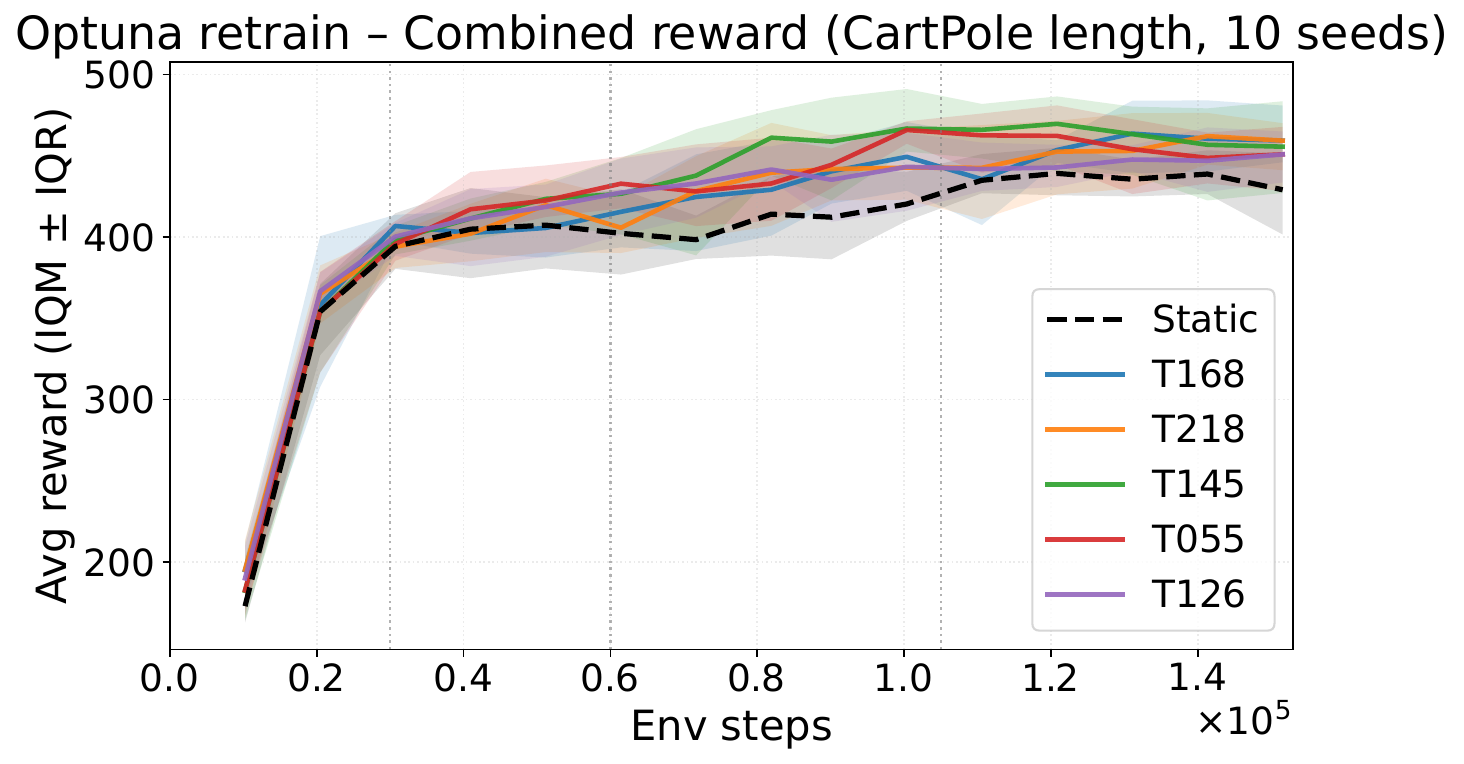}
    \caption{Learning curves (IQM $\pm$ IQR) for the top-5 retrained CartPole trials vs.\ static baseline.}
    \label{fig:cartpole-optuna-lineplot}
\end{figure}

\begin{figure}[ht]
    \centering
    \includegraphics[width=0.9\linewidth]{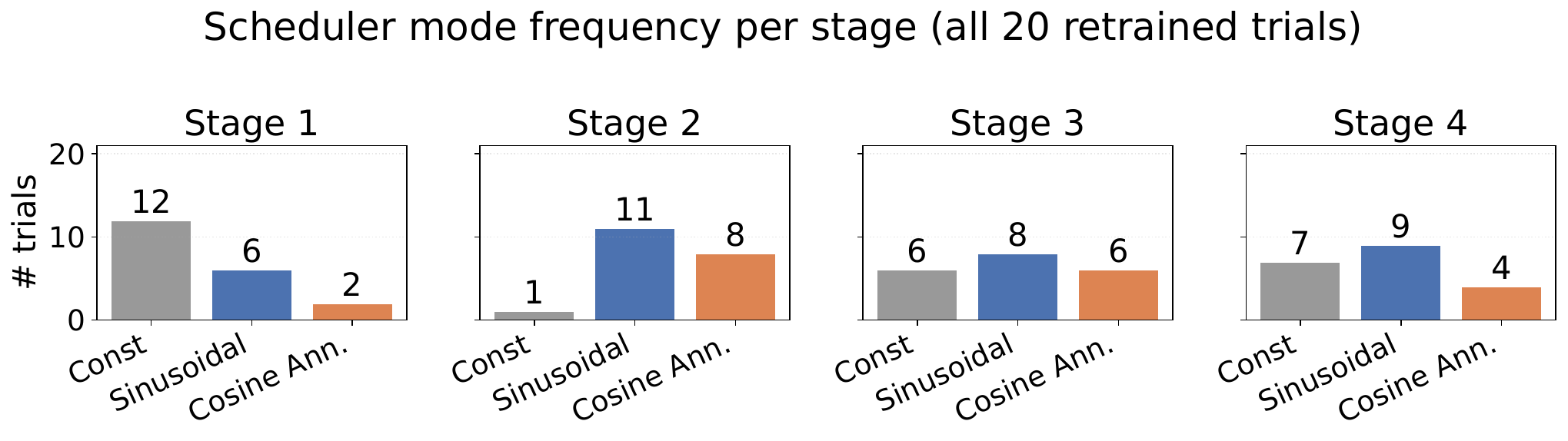}
    \caption{Frequency of each scheduler mode (constant, sinusoidal, cosine annealing) at each stage position across all 20 retrained CartPole trials. Stage~2 is active in nearly all configurations; Stage~1 is most often constant.}
    \label{fig:cartpole-optuna-mode-freq}
\end{figure}

\begin{figure}[ht]
    \centering
    \includegraphics[width=0.9\linewidth]{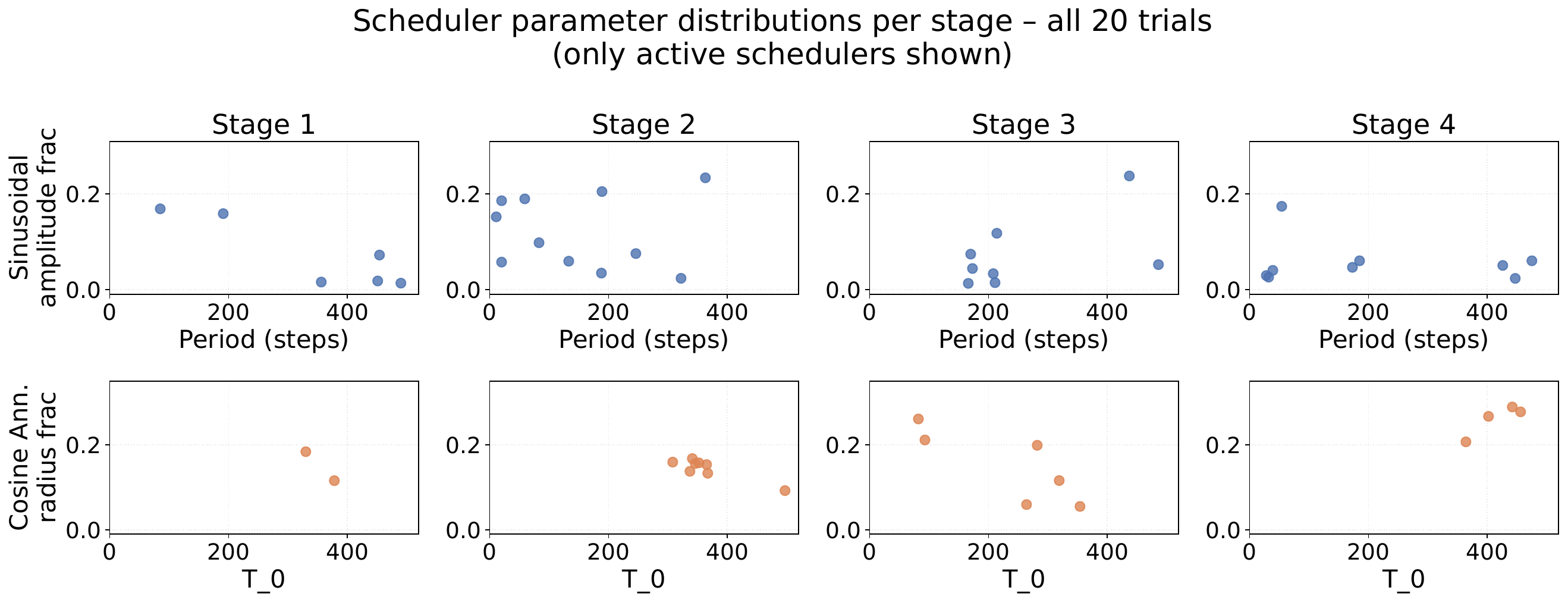}
    \caption{Hyperparameter distributions for active stages across all 20 retrained CartPole trials. Top row: sinusoidal amplitude fraction vs.\ period; bottom row: cosine annealing neighbourhood radius vs.\ $T_0$. Only trials where the respective mode is active are shown.}
    \label{fig:cartpole-optuna-param-dist}
\end{figure}

\begin{figure}[ht]
    \centering
    \includegraphics[width=0.9\linewidth]{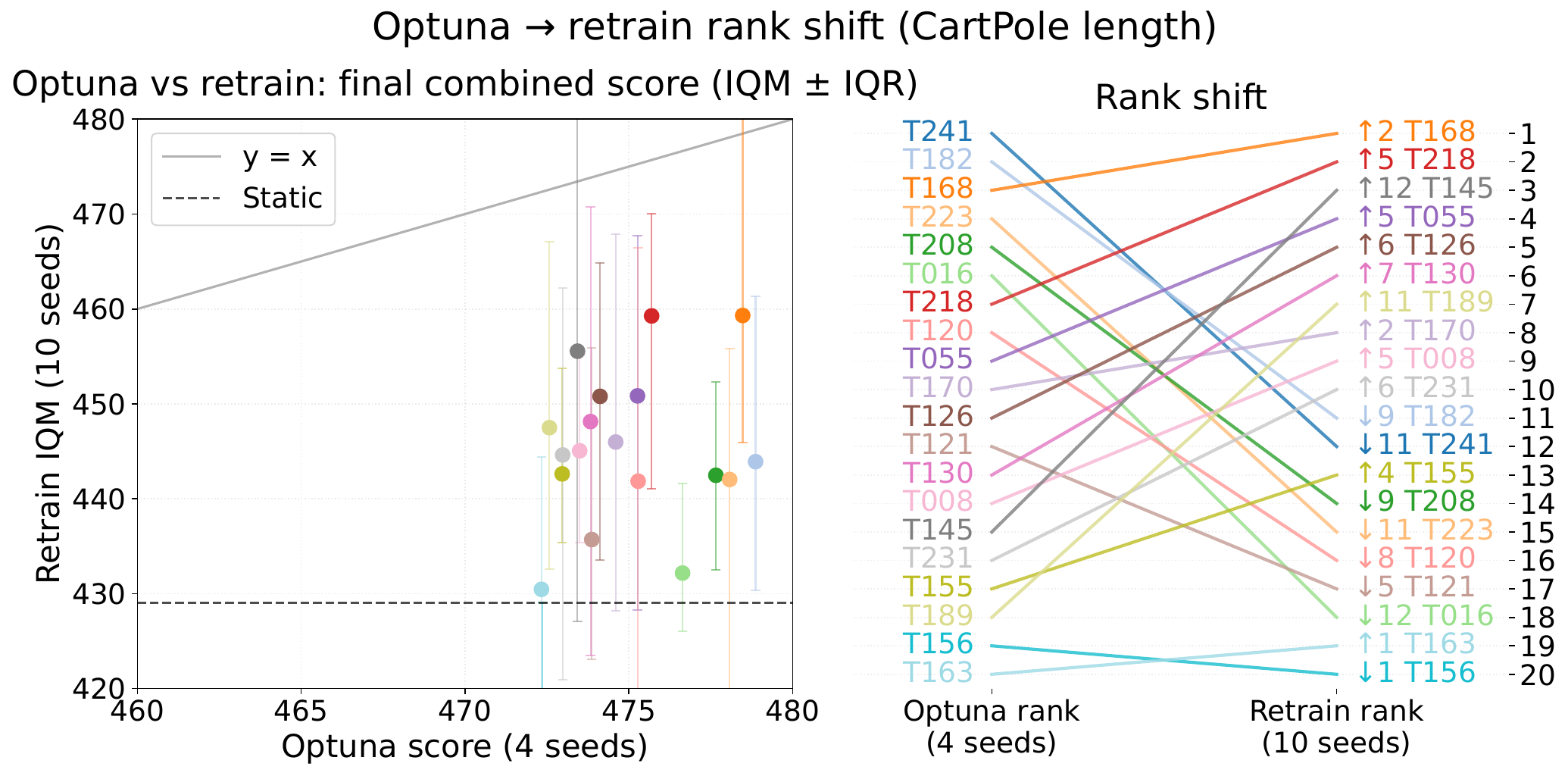}
    \caption{Rank shift between the 4-seed Optuna search ranking and the 10-seed retrain ranking for CartPole. Left: scatter of Optuna score vs.\ retrain IQM\,$\pm$\,IQR; the dashed line marks the static pool-7 baseline. Right: bump chart showing the rank change for each trial after retraining with more seeds.}
    \label{fig:cartpole-optuna-rank_shift}
\end{figure}

The figures below detail the CartPole Optuna retrain: training curves (Figure~\ref{fig:cartpole-optuna-lineplot}), scheduler mode frequency per stage (Figure~\ref{fig:cartpole-optuna-mode-freq}), hyperparameter distributions for active stages (Figure~\ref{fig:cartpole-optuna-param-dist}), and rank shift after retraining with more seeds (Figure~\ref{fig:cartpole-optuna-rank_shift}).

\subsection{Walker}
\label{app:walker-optuna}

\begin{figure}[ht]
    \centering
    \begin{subfigure}[c]{0.48\linewidth}
        \centering
        \includegraphics[width=\linewidth]{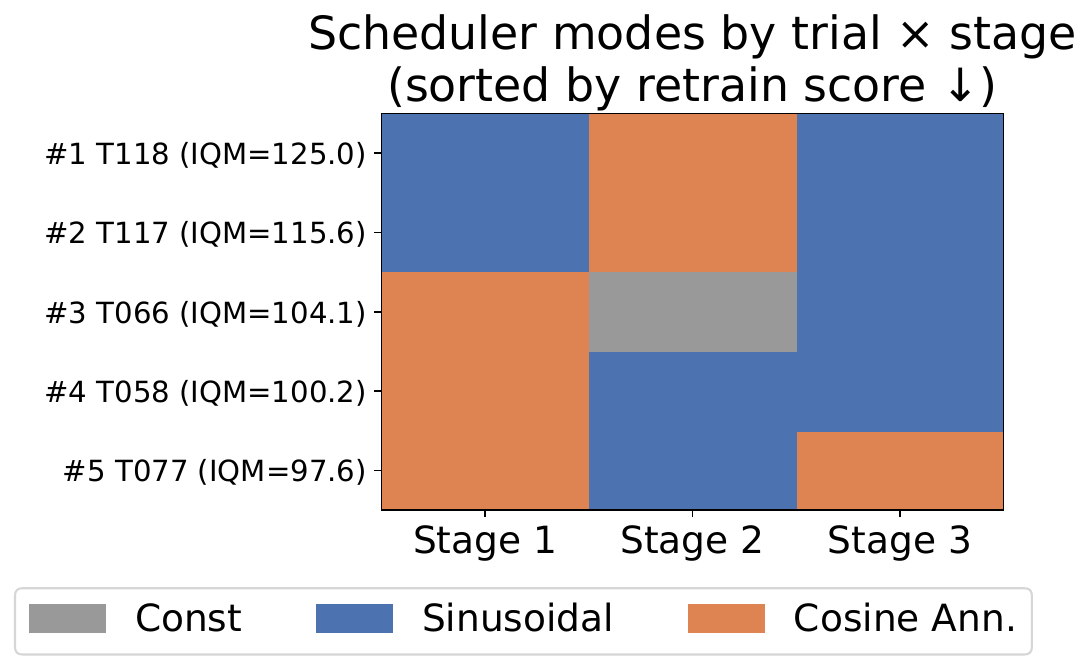}
        \caption{Scheduler mode per stage for the top-5 retrained trials, sorted by retrain IQM.}
        \label{fig:walker-optuna-heatmap}
    \end{subfigure}
    \hfill
    \begin{subfigure}[c]{0.48\linewidth}
        \centering
        \includegraphics[width=\linewidth]{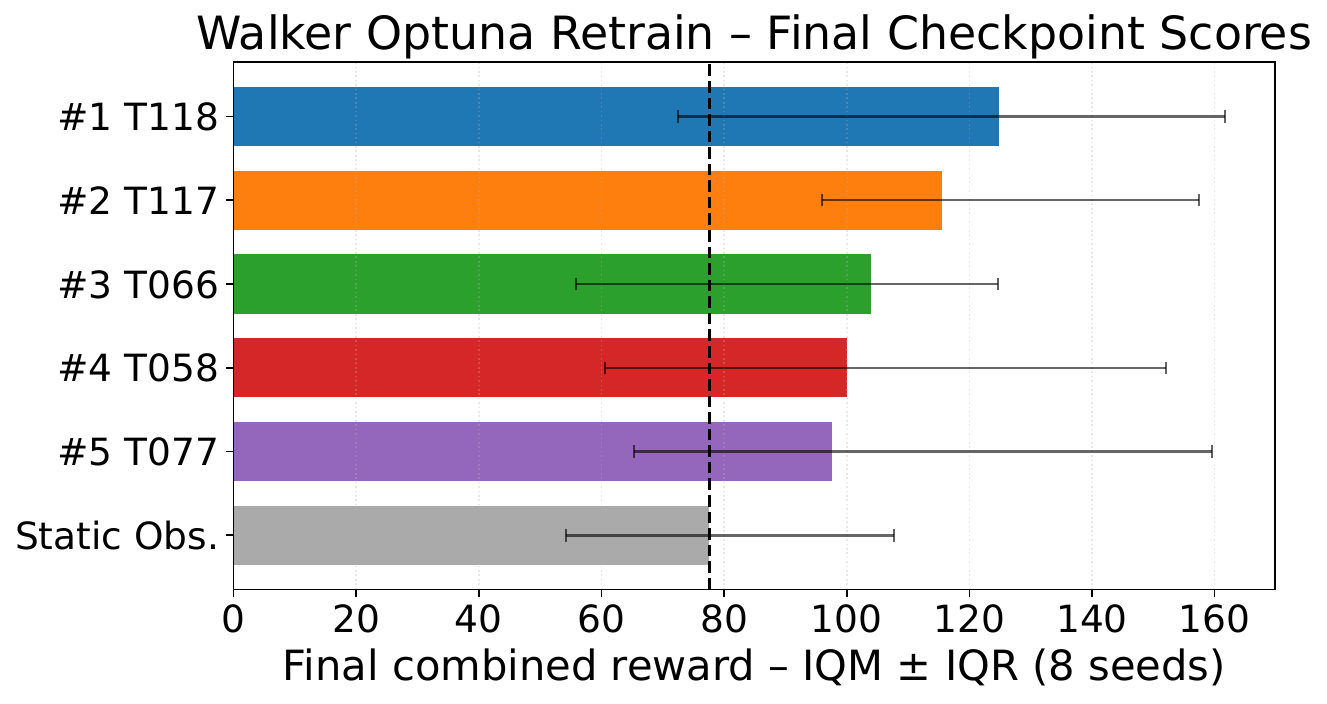}
        \caption{Final-checkpoint IQM $\pm$ IQR for top-5 retrained trials vs.\ static baseline (1.5\,M steps).}
        \label{fig:walker-optuna-scores}
    \end{subfigure}
    \caption{BipedalWalker Optuna retrain: scheduler patterns and final scores.}
\end{figure}

\begin{figure}[ht]
    \centering
    \includegraphics[width=0.8\linewidth]{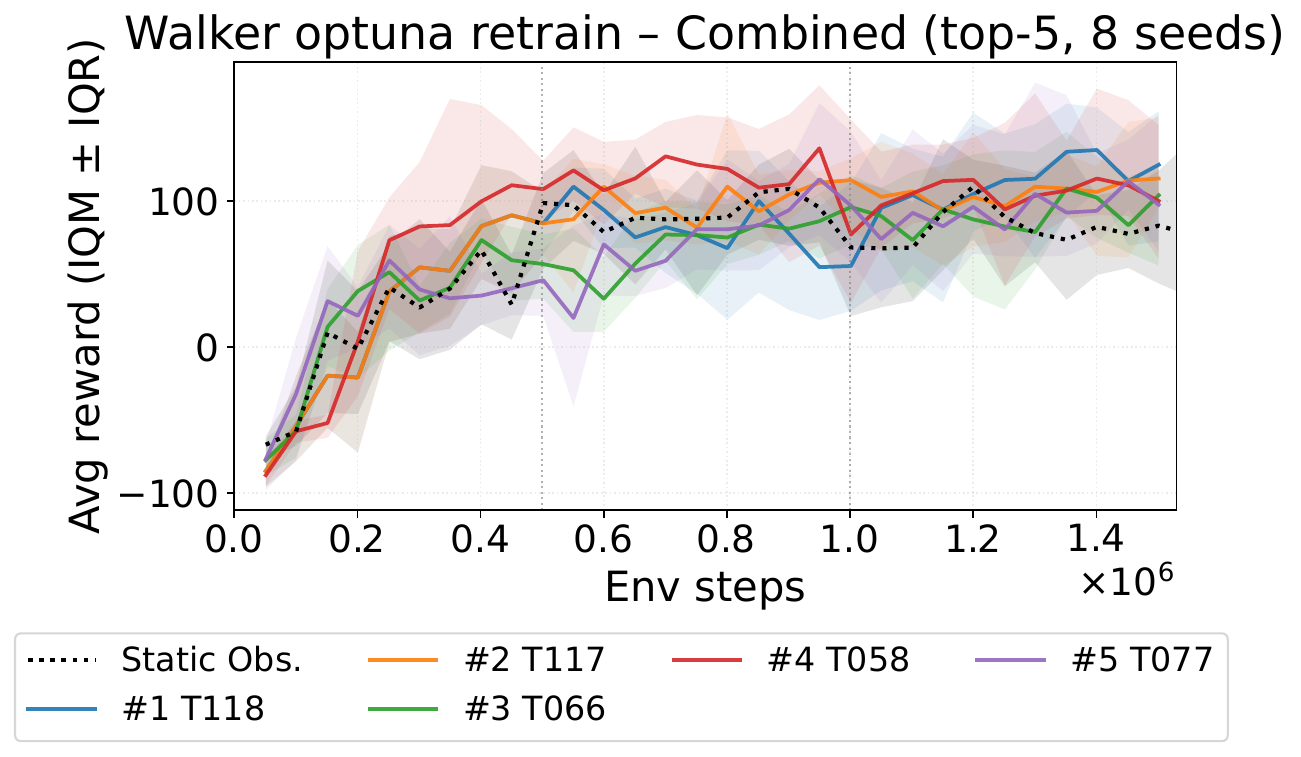}
    \caption{Learning curves (IQM $\pm$ IQR) for the top-5 retrained Walker trials vs.\ static baseline.}
    \label{fig:walker-optuna-lineplot}
\end{figure}

\begin{figure}[ht]
    \centering
    \includegraphics[width=0.9\linewidth]{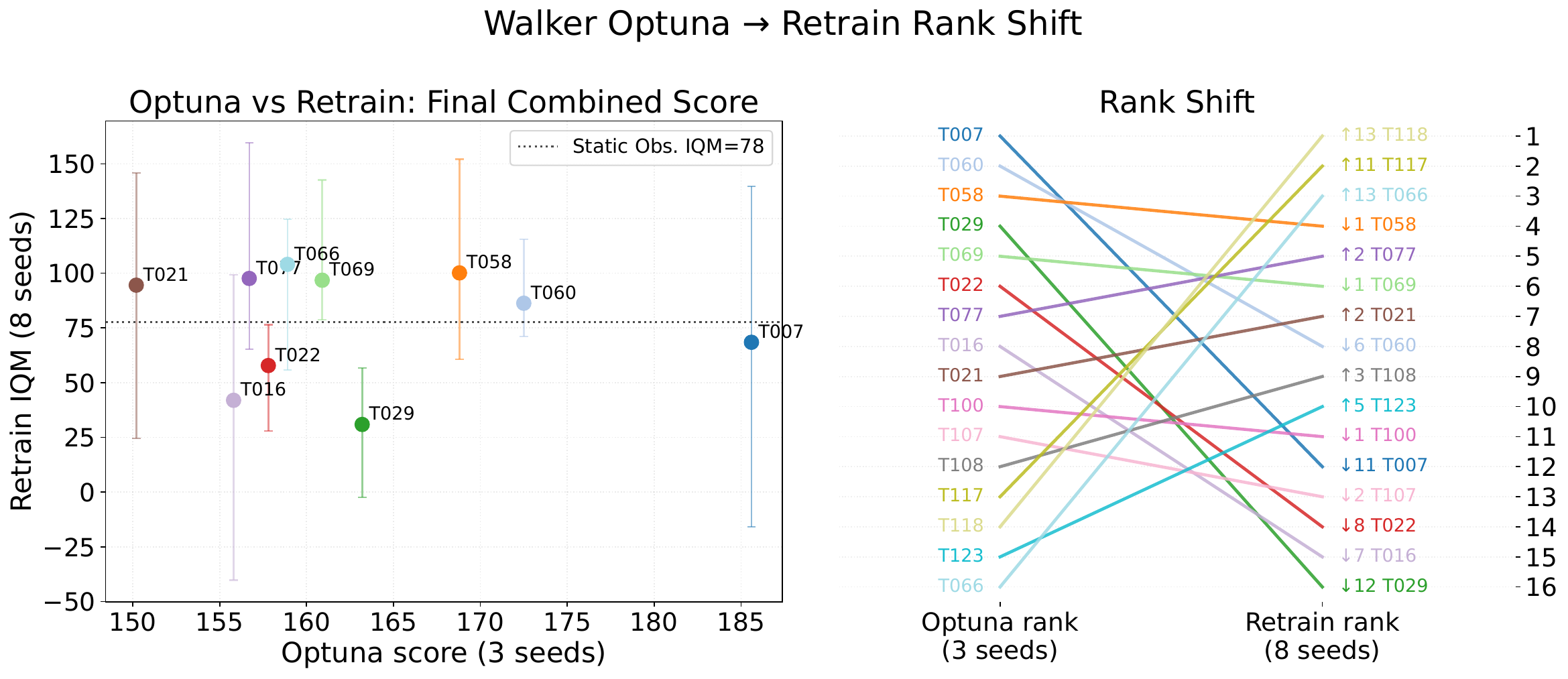}
    \caption{Rank shift between the 3-seed Optuna ranking and the 8-seed retrain ranking for BipedalWalker.}
    \label{fig:walker-optuna-rank_shift}
\end{figure}

\begin{figure}[ht]
    \centering
    \includegraphics[width=0.9\linewidth]{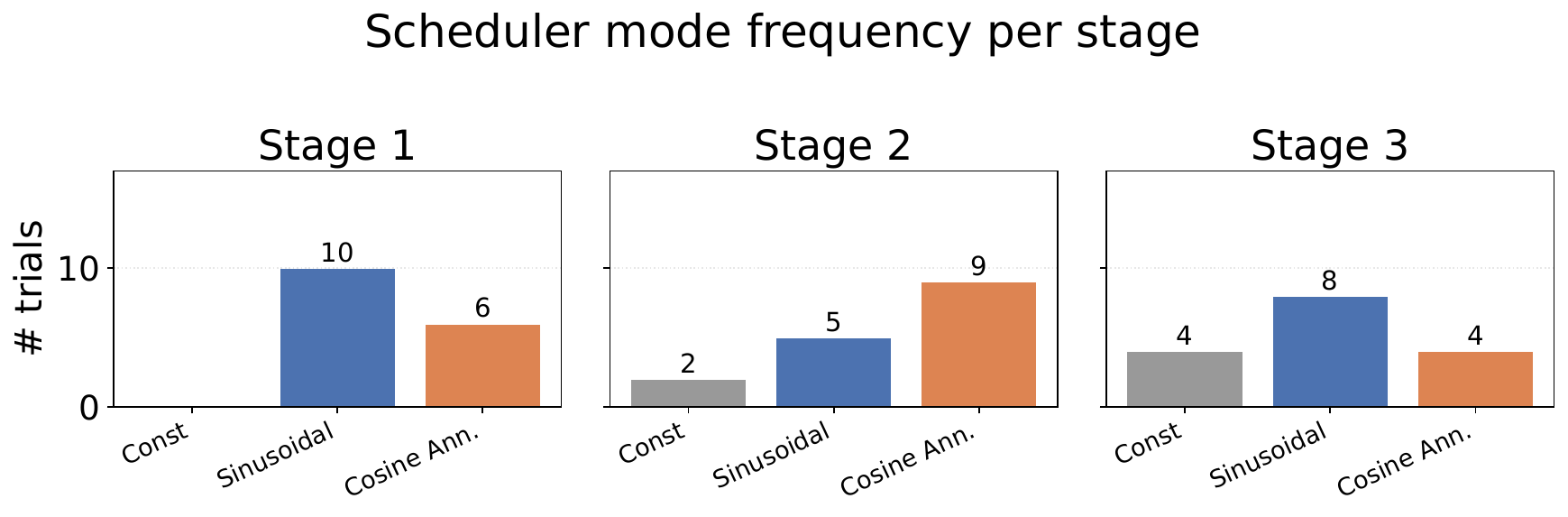}
    \caption{Scheduler mode frequency per stage across all retrained Walker trials.}
    \label{fig:walker-optuna-mode-freq}
\end{figure}

\begin{figure}[ht]
    \centering
    \includegraphics[width=0.9\linewidth]{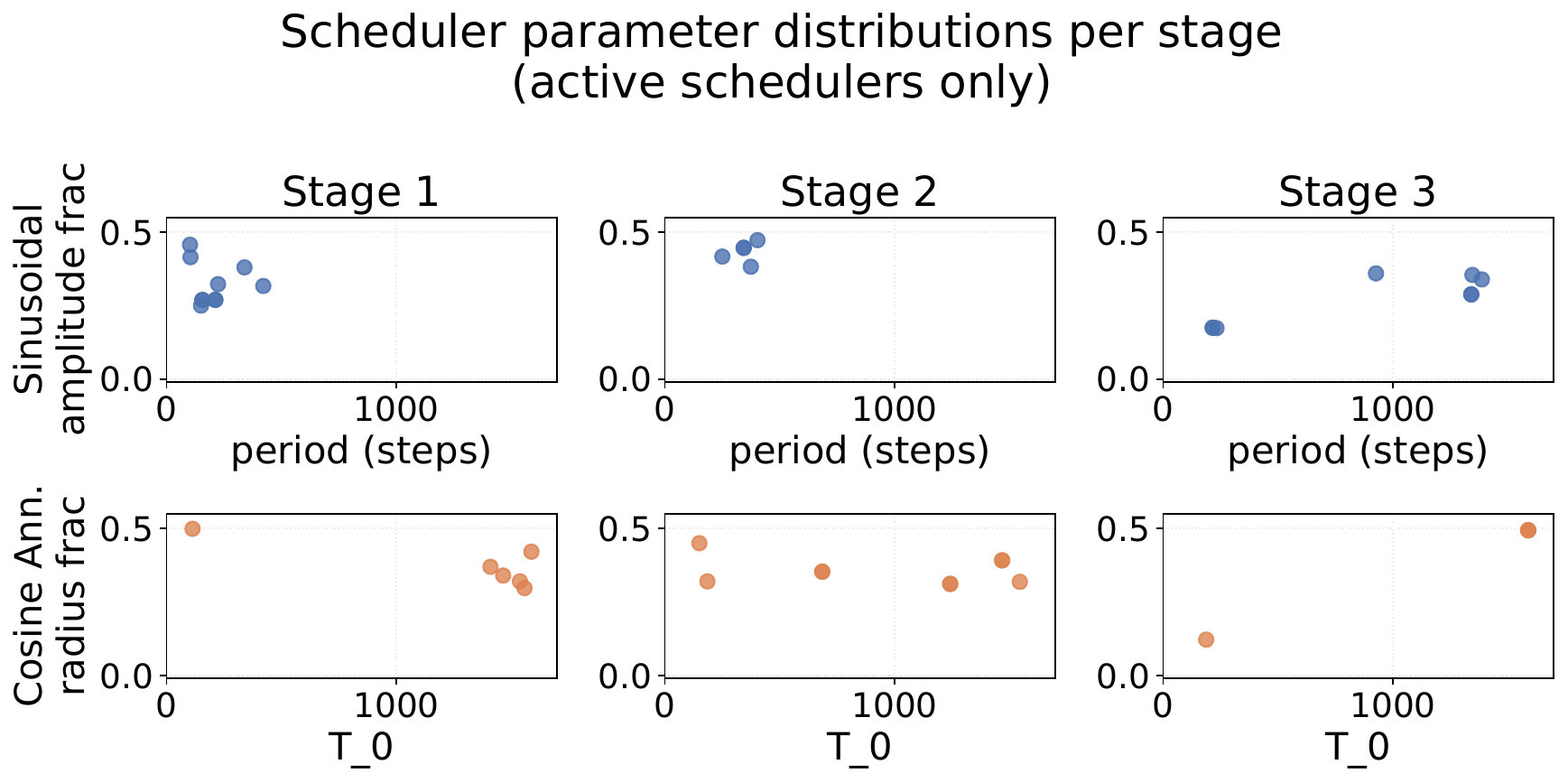}
    \caption{Scheduler parameter distributions per stage for all retrained Walker trials (active schedulers only).}
    \label{fig:walker-optuna-param-dist}
\end{figure}

The figures below mirror the CartPole analysis for BipedalWalker: scheduler patterns and final scores (Figure~\ref{fig:walker-optuna-heatmap}), training curves (Figure~\ref{fig:walker-optuna-lineplot}), rank shift (Figure~\ref{fig:walker-optuna-rank_shift}), mode frequency (Figure~\ref{fig:walker-optuna-mode-freq}), and hyperparameter distributions (Figure~\ref{fig:walker-optuna-param-dist}).

\FloatBarrier
\section{Scheduler Performance}
\label{app:per-family-results}

Tables~\ref{tab:ext_cartpole}--\ref{tab:ext_carracing_comy} report the per-scheduler-family IQM\,[Q1,\,Q3] of \texttt{avg\_combined} for both context-observability modes across all three environments. For each family and mode, we show the single best-performing configuration found during the scheduler search, selected by IQM of the last-checkpoint \texttt{avg\_combined} score. Both last-checkpoint and best-anytime IQM are reported. Bold marks the highest IQM in each column; in  Table~\ref{tab:ext_carracing_comx} the best-anytime blind column is an exception where the static baseline exceeds all dynamic schedulers, with the best dynamic result marked in italic bold for reference.

\begin{table}[h]
  \centering
  \caption{CartPole / length / pool-7. Per-scheduler-type IQM\,[Q1,\,Q3]
    of \texttt{avg\_combined} over 10 seeds.}
  \label{tab:ext_cartpole}
  \setlength{\tabcolsep}{4pt}
  \begin{tabular}{llrr}
  \toprule  
  \textbf{Scheduler} & \textbf{Mode}
    & \textbf{Last IQM [Q1, Q3]}
    & \textbf{Best IQM [Q1, Q3]} \\
  \midrule
  \multirow{2}{*}{\textit{Static Baseline}}
    & Observed & $429.0\;[401.5,\;451.4]$ & $449.7\;[440.3,\;462.8]$ \\
    & Blind    & $410.0\;[383.1,\;436.5]$ & $463.4\;[453.4,\;474.0]$ \\
  \midrule
  \multirow{2}{*}{Cont.\ Incrementer}
    & Observed & $449.8\;[411.5,\;473.2]$ & $477.6\;[449.6,\;494.7]$ \\
    & Blind    & $409.9\;[381.7,\;437.5]$ & $473.4\;[455.0,\;487.8]$ \\
  \addlinespace[3pt]
  \multirow{2}{*}{Cosine Annealing}
    & Observed & $458.2\;[444.6,\;468.8]$ & $477.9\;[469.4,\;484.6]$ \\
    & Blind    & $411.0\;[384.5,\;452.6]$ & $455.0\;[444.6,\;468.6]$ \\
  \addlinespace[3pt]
  \multirow{2}{*}{Lévy Walk}
    & Observed & $\mathbf{468.2}\;[454.7,\;484.3]$ & $471.8\;[457.1,\;493.2]$ \\
    & Blind    & $411.0\;[383.2,\;452.8]$ & $453.1\;[435.0,\;473.5]$ \\
  \addlinespace[3pt]
  \multirow{2}{*}{Ornstein--Uhlenbeck}
    & Observed & $453.8\;[443.3,\;467.0]$ & $467.1\;[460.3,\;474.6]$ \\
    & Blind    & $425.9\;[398.2,\;452.5]$ & $464.2\;[446.7,\;480.3]$ \\
  \addlinespace[3pt]
  \multirow{2}{*}{Phased OU}
    & Observed & $454.0\;[448.3,\;460.4]$ & $467.6\;[453.1,\;481.2]$ \\
    & Blind    & $418.6\;[383.9,\;447.3]$ & $458.1\;[434.7,\;475.4]$ \\
  \addlinespace[3pt]
  \multirow{2}{*}{Piecewise Const.}
    & Observed & $456.1\;[439.3,\;465.4]$ & $467.5\;[461.1,\;477.5]$ \\
    & Blind    & $\mathbf{462.9}\;[430.9,\;494.1]$ & $\mathbf{479.3}\;[465.5,\;495.9]$ \\
  \addlinespace[3pt]
  \multirow{2}{*}{Random Walk}
    & Observed & $462.4\;[452.0,\;472.9]$ & $475.3\;[462.1,\;484.2]$ \\
    & Blind    & $405.7\;[390.4,\;422.0]$ & $451.9\;[442.5,\;465.6]$ \\
  \addlinespace[3pt]
  \multirow{2}{*}{Sinusoidal}
    & Observed & $463.2\;[450.9,\;483.4]$ & $475.9\;[463.9,\;490.1]$ \\
    & Blind    & $409.7\;[396.8,\;425.3]$ & $463.0\;[457.9,\;470.3]$ \\
  \addlinespace[3pt]
  \multirow{2}{*}{Sinusoidal Jump}
    & Observed & $450.7\;[444.7,\;460.2]$ & $466.1\;[457.9,\;475.5]$ \\
    & Blind    & $436.8\;[417.6,\;468.3]$ & $453.7\;[435.4,\;472.5]$ \\
  \addlinespace[3pt]
  \multirow{2}{*}{Sudden Jump}
    & Observed & $457.9\;[442.8,\;479.9]$ & $\mathbf{479.6}\;[469.3,\;491.0]$ \\
    & Blind    & $427.0\;[402.5,\;449.2]$ & $469.6\;[452.5,\;480.5]$ \\
  \bottomrule
  \end{tabular}
\end{table}

\begin{table}[t]
  \centering
  \caption{BipedalWalker / COM\_X / pool-3. Per-scheduler-type IQM\,[Q1,\,Q3]
    of \texttt{avg\_combined} over 8 seeds.}
  \label{tab:ext_walker}
  \setlength{\tabcolsep}{4pt}
  \begin{tabular}{llrr}
  \toprule
  \textbf{Scheduler} & \textbf{Mode}
    & \textbf{Last IQM [Q1, Q3]}
    & \textbf{Best IQM [Q1, Q3]} \\
  \midrule
  \multirow{2}{*}{\textit{Static Baseline}}
    & Observed & $31.0\;[-34.9,\;63.8]$  & $134.2\;[89.5,\;167.9]$ \\
    & Blind    & $91.9\;[60.6,\;113.0]$  & $146.9\;[118.3,\;181.4]$ \\
  \midrule
  \multirow{2}{*}{Cont.\ Incrementer}
    & Observed & $113.6\;[97.6,\;130.2]$  & $154.5\;[140.1,\;179.6]$ \\
    & Blind    & $121.5\;[62.1,\;178.3]$  & $157.5\;[121.1,\;196.8]$ \\
  \addlinespace[3pt]
  \multirow{2}{*}{Cosine Annealing}
    & Observed & $106.5\;[75.9,\;142.5]$  & $153.7\;[134.7,\;180.4]$ \\
    & Blind    & $50.0\;[-36.0,\;121.8]$  & $100.6\;[10.3,\;172.1]$ \\
  \addlinespace[3pt]
  \multirow{2}{*}{Lévy Walk}
    & Observed & $77.6\;[54.9,\;105.8]$   & $137.5\;[121.6,\;154.6]$ \\
    & Blind    & $\mathbf{131.1}\;[94.9,\;167.6]$  & $\mathbf{175.5}\;[151.9,\;200.5]$ \\
  \addlinespace[3pt]
  \multirow{2}{*}{Ornstein--Uhlenbeck}
    & Observed & $83.9\;[10.9,\;157.8]$   & $131.6\;[49.1,\;220.9]$ \\
    & Blind    & $102.3\;[46.8,\;154.7]$  & $148.8\;[120.5,\;182.7]$ \\
  \addlinespace[3pt]
  \multirow{2}{*}{Phased OU}
    & Observed & $83.4\;[70.5,\;96.1]$    & $129.1\;[100.7,\;164.9]$ \\
    & Blind    & $104.4\;[90.8,\;124.3]$  & $150.0\;[117.5,\;179.9]$ \\
  \addlinespace[3pt]
  \multirow{2}{*}{Random Walk}
    & Observed & $111.6\;[81.6,\;142.2]$  & $171.3\;[158.6,\;184.1]$ \\
    & Blind    & $76.4\;[25.2,\;134.2]$   & $158.7\;[96.9,\;198.1]$ \\
  \addlinespace[3pt]
  \multirow{2}{*}{Sinusoidal}
    & Observed & $\mathbf{151.4}\;[122.2,\;185.3]$ & $\mathbf{193.0}\;[156.7,\;233.4]$ \\
    & Blind    & $111.7\;[58.7,\;153.3]$  & $137.9\;[73.2,\;183.8]$ \\
  \addlinespace[3pt]
  \multirow{2}{*}{Sinusoidal Jump}
    & Observed & $93.7\;[82.0,\;104.4]$   & $166.9\;[143.2,\;180.0]$ \\
    & Blind    & $94.0\;[79.8,\;105.4]$   & $160.1\;[144.6,\;178.5]$ \\
  \addlinespace[3pt]
  \multirow{2}{*}{Sudden Jump}
    & Observed & $97.6\;[34.9,\;147.9]$   & $143.3\;[101.9,\;160.6]$ \\
    & Blind    & $121.8\;[103.8,\;138.8]$ & $150.3\;[132.9,\;189.4]$ \\
  \bottomrule
  \end{tabular}
\end{table}

\begin{table}[t]
  \centering
  \caption{CarRacing / longitudinal / pool-3. Per-scheduler-type IQM\,[Q1,\,Q3]
    of \texttt{avg\_combined} over 5 seeds.}
  \label{tab:ext_carracing_comx}
  \setlength{\tabcolsep}{4pt}
  \begin{tabular}{llrr}
  \toprule  
  \textbf{Scheduler} & \textbf{Mode}
    & \textbf{Last IQM [Q1, Q3]}
    & \textbf{Best IQM [Q1, Q3]} \\
  \midrule
  \multirow{2}{*}{\textit{Static Baseline}}
    & Observed & $-81.0\;[-93.1,\;-69.4]$ & $446.7\;[160.9,\;606.1]$ \\
    & Blind    & $521.9\;[377.7,\;656.7]$ & $\mathbf{833.5}\;[775.6,\;873.4]$ \\
  \midrule
  \multirow{2}{*}{Cont.\ Incrementer}
    & Observed & $-28.9\;[-35.4,\;-17.4]$ & $327.5\;[221.3,\;436.2]$ \\
    & Blind    & $262.8\;[94.8,\;466.2]$  & $703.5\;[530.4,\;833.7]$ \\
  \addlinespace[3pt]
  \multirow{2}{*}{Cosine Annealing}
    & Observed & $-59.6\;[-86.5,\;-12.9]$ & $594.7\;[350.5,\;745.6]$ \\
    & Blind    & $\mathbf{602.5}\;[409.6,\;664.3]$ & $762.0\;[617.0,\;771.4]$ \\
  \addlinespace[3pt]
  \multirow{2}{*}{Lévy Walk}
    & Observed & $\mathbf{-7.0}\;[-24.6,\;12.1]$  & $503.5\;[224.7,\;716.4]$ \\
    & Blind    & $542.6\;[246.4,\;762.8]$  & $\textit{\textbf{811.8}}\;[723.0,\;854.3]$ \\
  \addlinespace[3pt]
  \multirow{2}{*}{Ornstein--Uhlenbeck}
    & Observed & $-30.8\;[-60.5,\;0.1]$   & $470.8\;[447.3,\;512.4]$ \\
    & Blind    & $503.7\;[453.5,\;537.9]$  & $731.7\;[697.4,\;795.6]$ \\
  \addlinespace[3pt]
  \multirow{2}{*}{Phased OU}
    & Observed & $-40.2\;[-78.6,\;-17.3]$ & $441.1\;[232.0,\;649.9]$ \\
    & Blind    & $254.5\;[-48.7,\;458.3]$  & $632.4\;[552.8,\;732.5]$ \\
  \addlinespace[3pt]
  \multirow{2}{*}{Random Walk}
    & Observed & $-77.5\;[-84.3,\;-66.1]$ & $454.1\;[379.1,\;582.3]$ \\
    & Blind    & $308.3\;[145.9,\;494.0]$  & $798.3\;[767.1,\;825.3]$ \\
  \addlinespace[3pt]
  \multirow{2}{*}{Sinusoidal}
    & Observed & $-11.0\;[-38.1,\;14.6]$  & $574.5\;[513.0,\;657.2]$ \\
    & Blind    & $589.6\;[534.7,\;656.8]$  & $788.4\;[727.0,\;809.7]$ \\
  \addlinespace[3pt]
  \multirow{2}{*}{Sinusoidal Jump}
    & Observed & $-50.2\;[-81.3,\;-4.9]$  & $373.9\;[206.6,\;479.2]$ \\
    & Blind    & $280.4\;[76.3,\;526.6]$   & $754.7\;[702.2,\;840.7]$ \\
  \addlinespace[3pt]
  \multirow{2}{*}{Sudden Jump}
    & Observed & $-57.0\;[-63.4,\;-51.1]$ & $\mathbf{692.3}\;[641.7,\;733.5]$ \\
    & Blind    & $305.9\;[235.6,\;477.2]$  & $742.9\;[634.8,\;815.9]$ \\
  \bottomrule
  \end{tabular}
\end{table}

\begin{table}[t]
  \centering
  \caption{CarRacing / lateral / pool-3. Per-scheduler-type IQM\,[Q1,\,Q3]
    of \texttt{avg\_combined} over 5 seeds.}
  \label{tab:ext_carracing_comy}
  \setlength{\tabcolsep}{4pt}
  \begin{tabular}{llrr}
  \toprule
  \textbf{Scheduler} & \textbf{Mode}
    & \textbf{Last IQM [Q1, Q3]}
    & \textbf{Best IQM [Q1, Q3]} \\
  \midrule
  \multirow{2}{*}{\textit{Static Baseline}}
    & Observed & $-85.9\;[-93.1,\;-76.2]$ & $358.5\;[183.6,\;450.8]$ \\
    & Blind    & $103.6\;[-59.0,\;378.9]$ & $617.7\;[272.5,\;803.1]$ \\
  \midrule
  \multirow{2}{*}{Cont.\ Incrementer}
    & Observed & $-70.5\;[-83.7,\;-45.1]$ & $380.3\;[110.8,\;643.4]$ \\
    & Blind    & $153.8\;[-16.6,\;391.1]$ & $564.0\;[443.6,\;656.1]$ \\
  \addlinespace[3pt]
  \multirow{2}{*}{Cosine Annealing}
    & Observed & $-40.1\;[-82.5,\;-10.1]$ & $395.8\;[273.8,\;560.3]$ \\
    & Blind    & $533.2\;[458.5,\;618.7]$ & $762.3\;[720.9,\;796.3]$ \\
  \addlinespace[3pt]
  \multirow{2}{*}{Lévy Walk}
    & Observed & $-51.3\;[-77.7,\;-17.4]$ & $488.1\;[354.0,\;570.5]$ \\
    & Blind    & $263.4\;[164.6,\;372.5]$ & $756.3\;[661.9,\;781.4]$ \\
  \addlinespace[3pt]
  \multirow{2}{*}{Ornstein--Uhlenbeck}
    & Observed & $\mathbf{-35.2}\;[-67.3,\;-7.4]$ & $397.9\;[323.9,\;471.0]$ \\
    & Blind    & $177.2\;[-67.5,\;333.2]$ & $638.0\;[374.1,\;771.8]$ \\
  \addlinespace[3pt]
  \multirow{2}{*}{Phased OU}
    & Observed & $-88.7\;[-93.1,\;-82.7]$ & $554.4\;[491.3,\;624.4]$ \\
    & Blind    & $302.5\;[242.7,\;392.1]$ & $654.3\;[536.9,\;730.3]$ \\
  \addlinespace[3pt]
  \multirow{2}{*}{Random Walk}
    & Observed & $-47.3\;[-68.8,\;-18.1]$ & $587.0\;[556.9,\;639.5]$ \\
    & Blind    & $375.6\;[336.6,\;423.1]$ & $\mathbf{778.9}\;[718.5,\;813.5]$ \\
  \addlinespace[3pt]
  \multirow{2}{*}{Sinusoidal}
    & Observed & $-54.1\;[-79.9,\;-31.2]$ & $446.3\;[397.0,\;513.5]$ \\
    & Blind    & $360.6\;[226.8,\;545.9]$ & $705.5\;[647.8,\;755.8]$ \\
  \addlinespace[3pt]
  \multirow{2}{*}{Sinusoidal Jump}
    & Observed & $-72.8\;[-82.1,\;-60.3]$ & $525.7\;[342.8,\;667.6]$ \\
    & Blind    & $463.9\;[364.5,\;548.6]$ & $729.8\;[659.1,\;817.7]$ \\
  \addlinespace[3pt]
  \multirow{2}{*}{Sudden Jump}
    & Observed & $-85.1\;[-90.9,\;-60.6]$ & $\mathbf{634.1}\;[520.7,\;659.4]$ \\
    & Blind    & $\mathbf{592.3}\;[549.1,\;625.7]$ & $735.2\;[713.3,\;754.5]$ \\
  \bottomrule
  \end{tabular}
\end{table}

\FloatBarrier
\section{Scheduler Training Progression}
\label{sec:appendix-line-plots}

Each figure shows the IQM episode return averaged over training checkpoints for all scheduler families, with one line per family using its best representative run. Left panel: dynamic \textbf{observed} schedulers.
Right panel: dynamic \textbf{blind} schedulers. The dashed line is the static pool baseline. Shaded bands indicate Q1--Q3 across seeds.

\begin{figure}[h]
    \centering 
    \includegraphics[width=\linewidth]{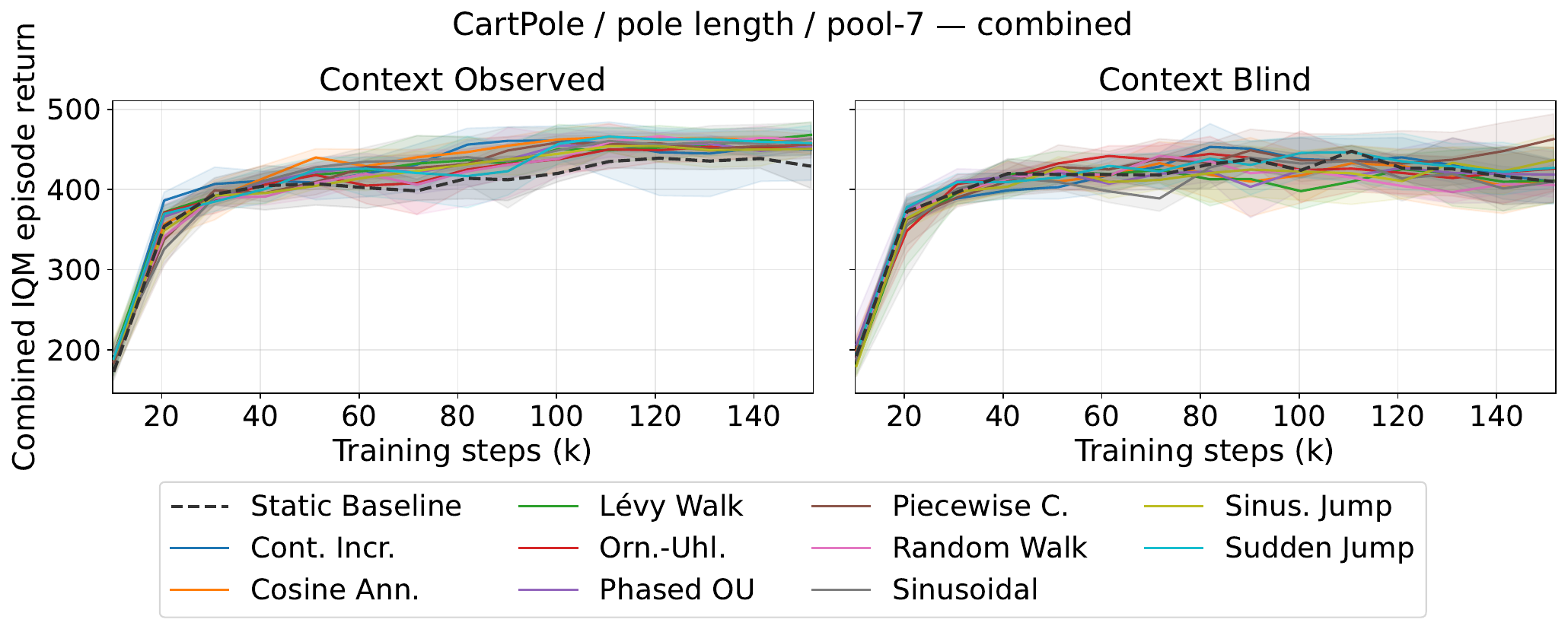}
    \caption{CartPole / pole length / pool-7 — \textbf{combined} IQM (all
      contexts, 10 seeds). Each family's best representative is selected by
      final combined IQM.}
    \label{fig:line-cartpole-combined}
\end{figure}

\begin{figure}[h]
    \centering
    \includegraphics[width=\linewidth]{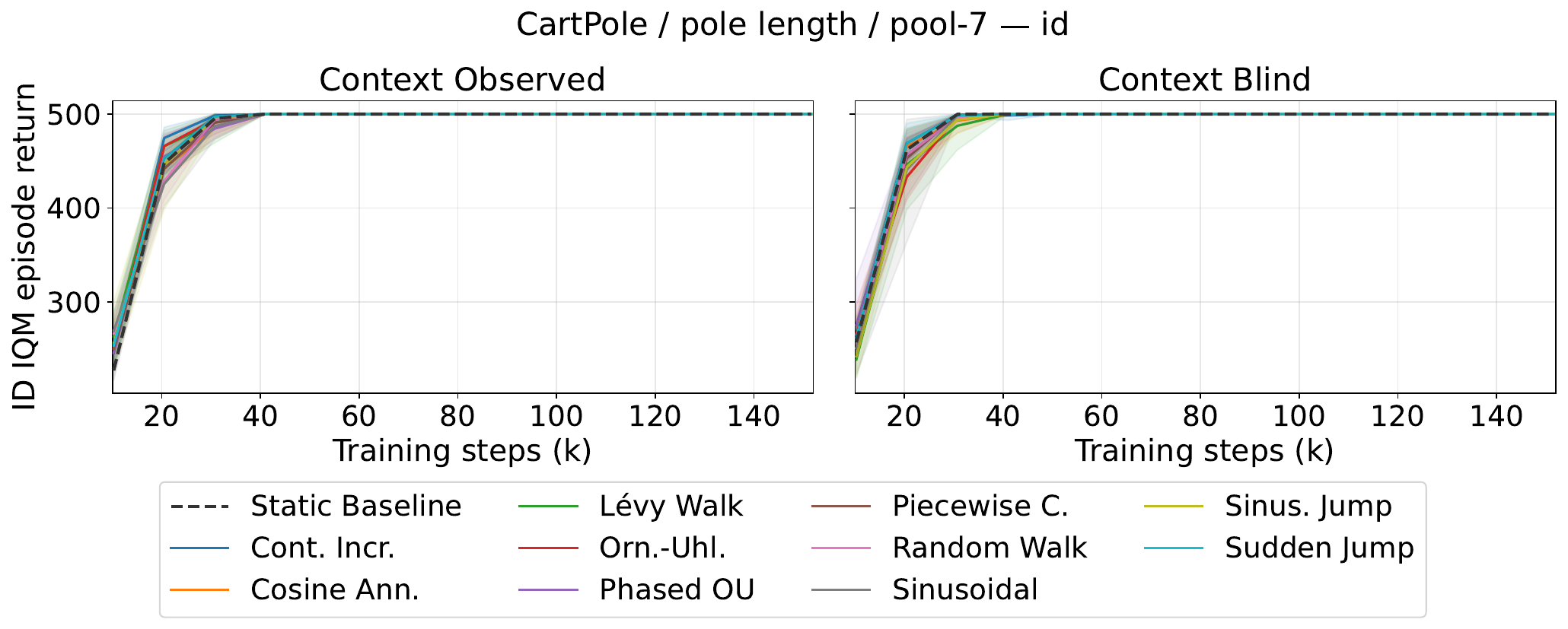}
    \caption{CartPole / pole length / pool-7 — \textbf{ID} IQM
      (in-distribution contexts only, 10 seeds).}
    \label{fig:line-cartpole-id}
\end{figure} 

\begin{figure}[h]
    \centering
    \includegraphics[width=\linewidth]{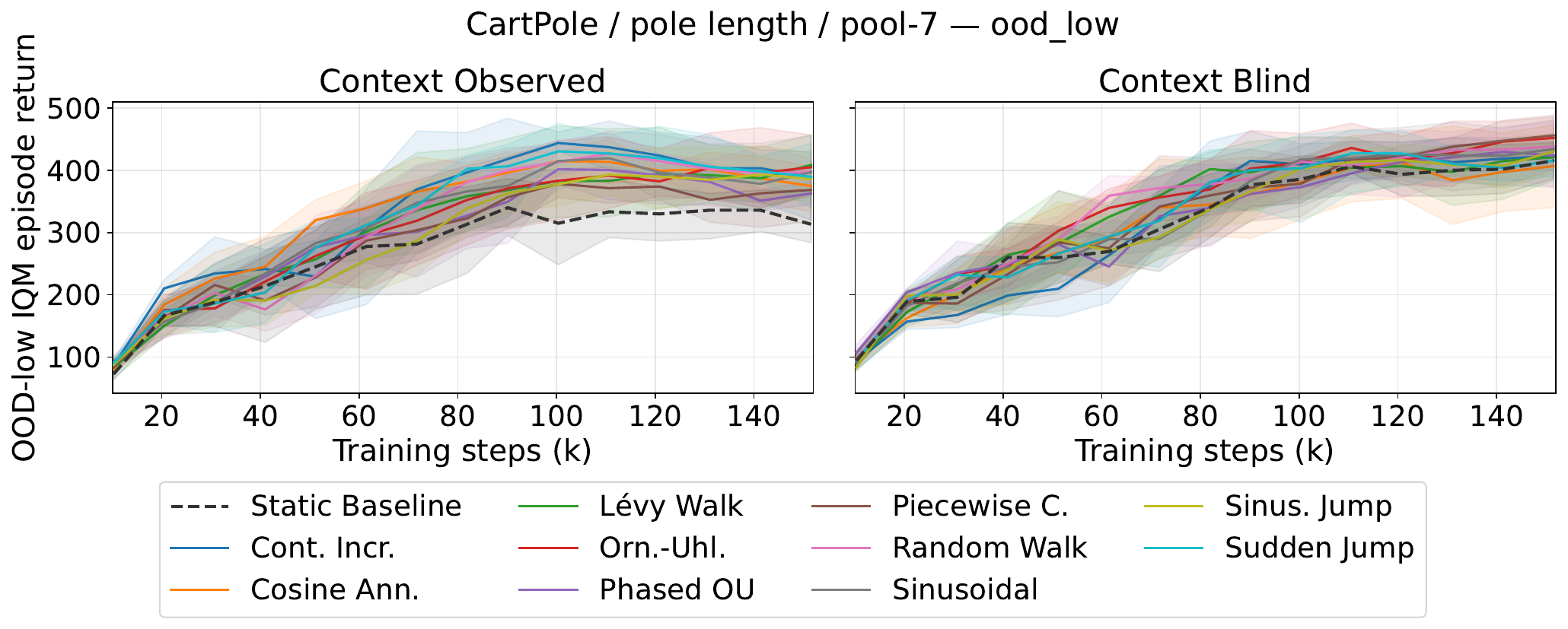}
    \caption{CartPole / pole length / pool-7 — \textbf{OOD-low} IQM
      (short-pole contexts below training range, 10 seeds).}
    \label{fig:line-cartpole-ood-low}
\end{figure} 

\begin{figure}[h]
    \centering
    \includegraphics[width=\linewidth]{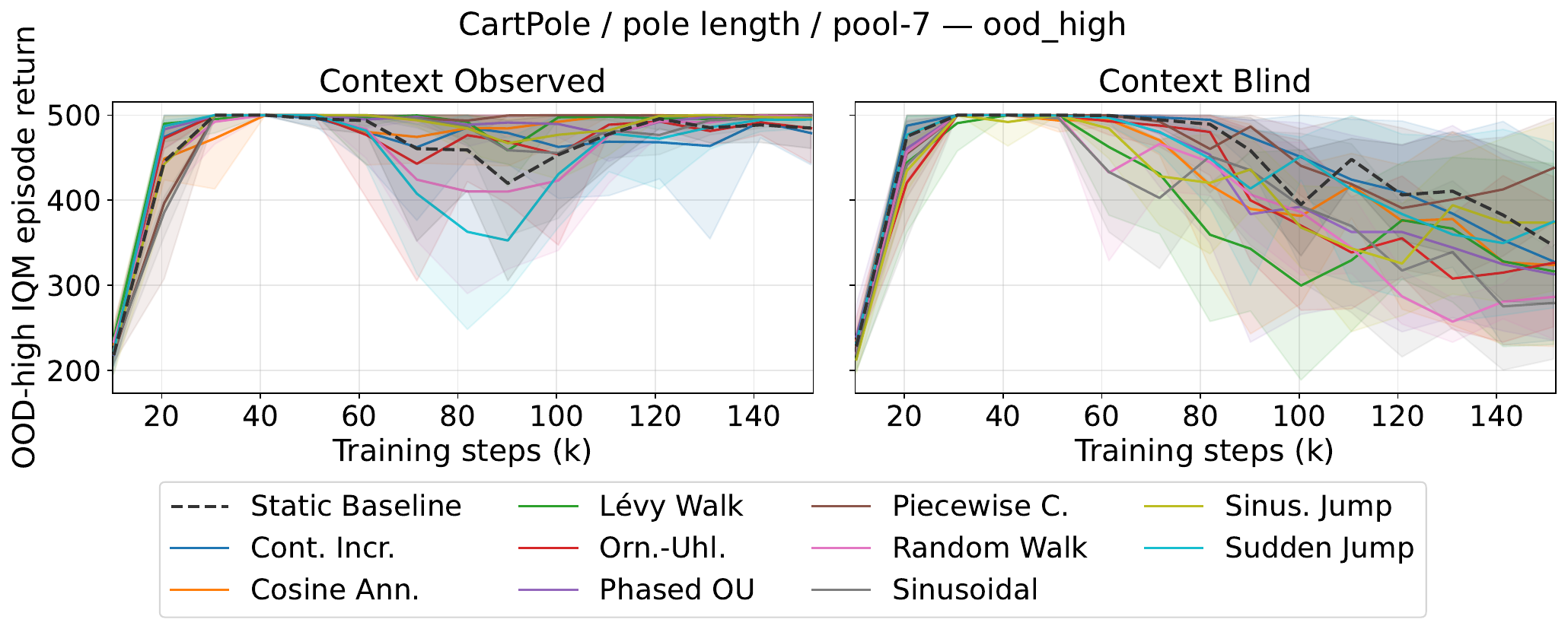}
    \caption{CartPole / pole length / pool-7 — \textbf{OOD-high} IQM
      (long-pole contexts above training range, 10 seeds).}
    \label{fig:line-cartpole-ood-high}
\end{figure}


\begin{figure}[h]
    \centering
    \includegraphics[width=\linewidth]{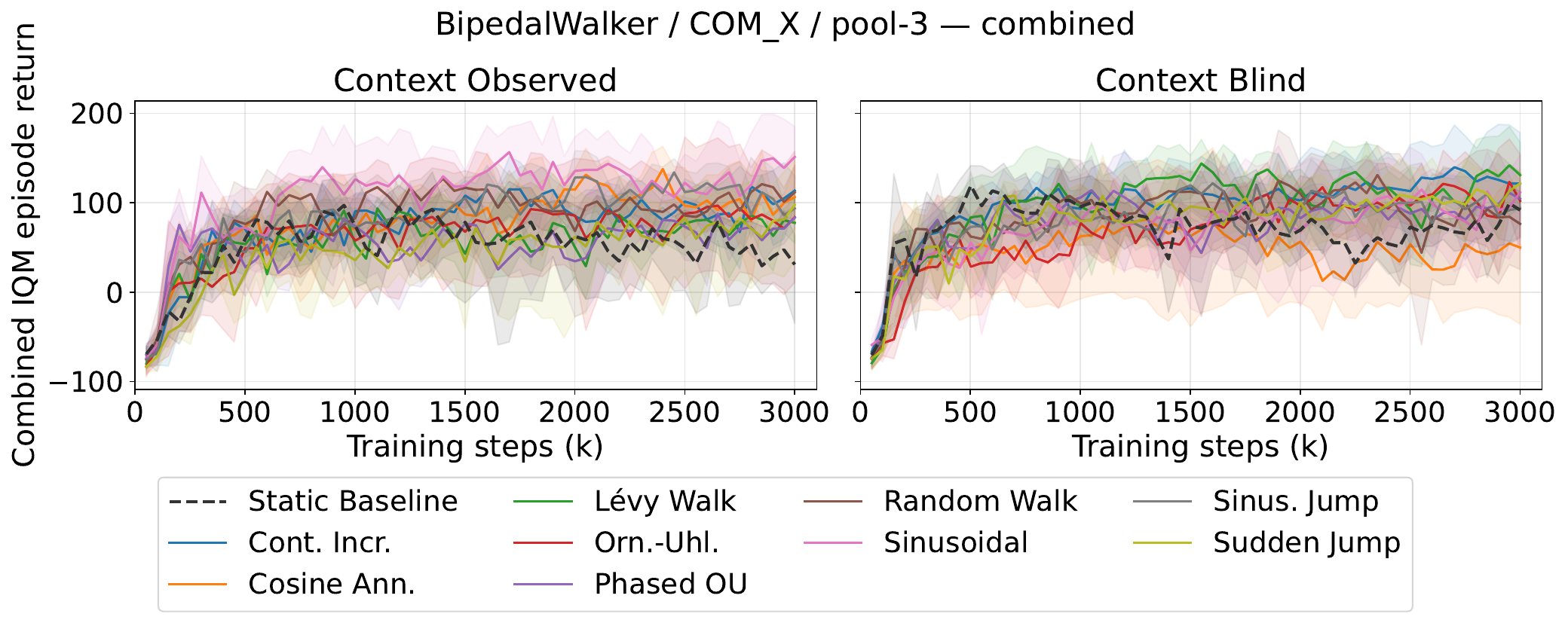}
    \caption{BipedalWalker / COM\_X / pool-3 — \textbf{combined} IQM
      (all contexts, 8 seeds).}
    \label{fig:line-walker-combined}
\end{figure} 

\begin{figure}[p]
    \centering
    \includegraphics[width=\linewidth]{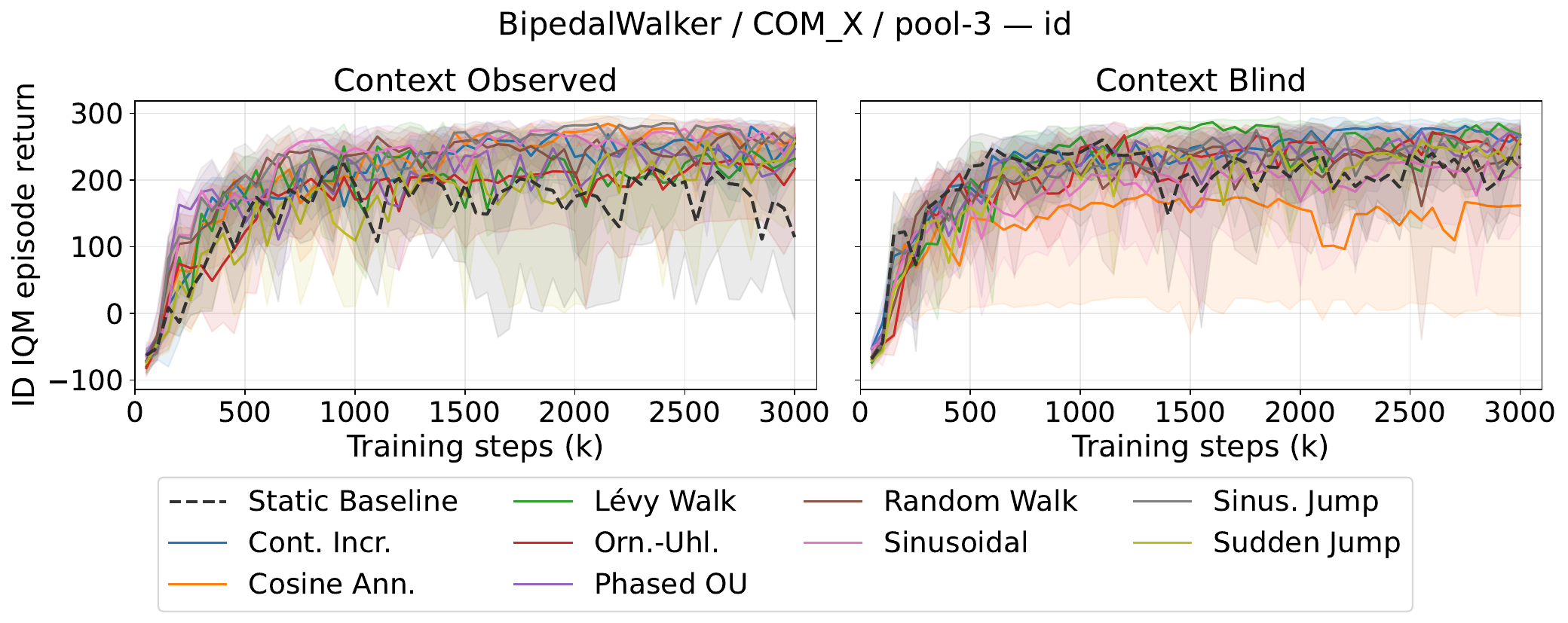}
    \caption{BipedalWalker / COM\_X / pool-3 — \textbf{ID} IQM
      (in-distribution contexts only, 8 seeds).}
    \label{fig:line-walker-id}
\end{figure}
  
\begin{figure}[p]
    \centering
    \includegraphics[width=\linewidth]{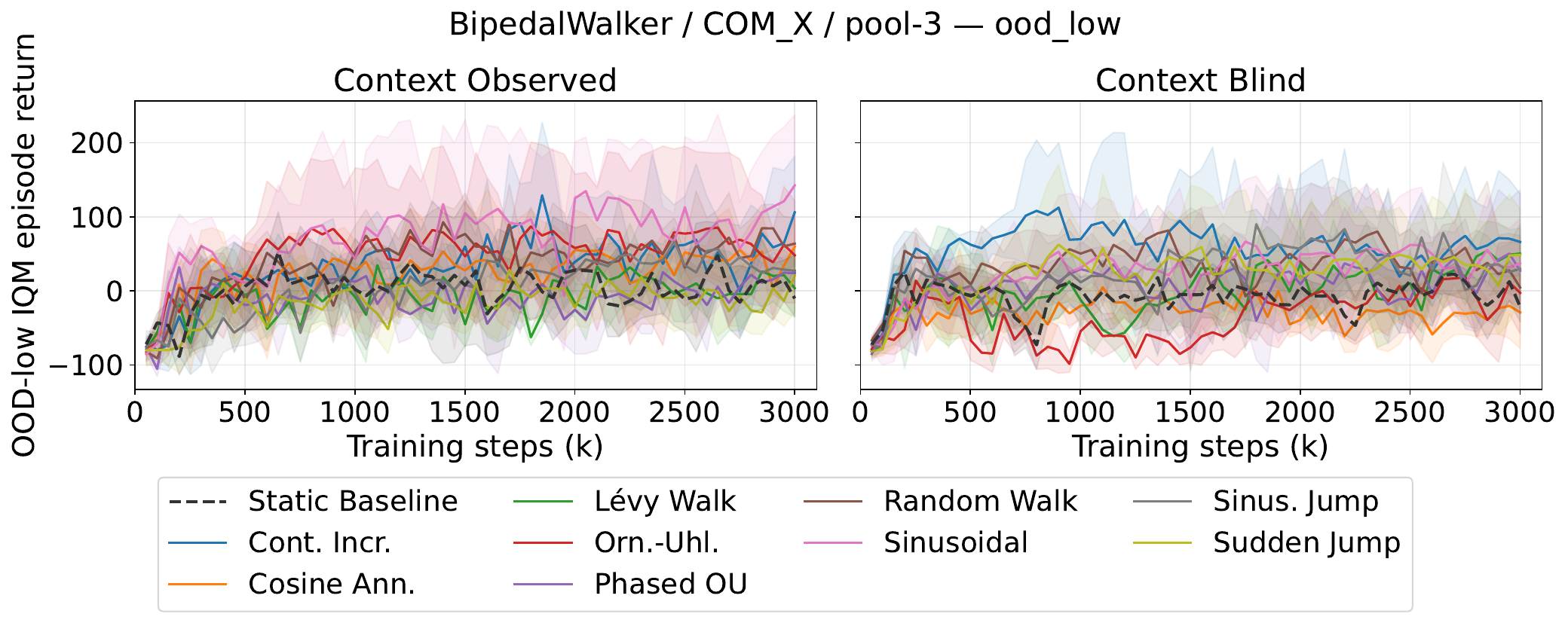}
    \caption{BipedalWalker / COM\_X / pool-3 — \textbf{OOD-low} IQM
      (centre-of-mass shifted below training range, 8 seeds).}
    \label{fig:line-walker-ood-low}
\end{figure}

\begin{figure}[p]
    \centering
    \includegraphics[width=\linewidth]{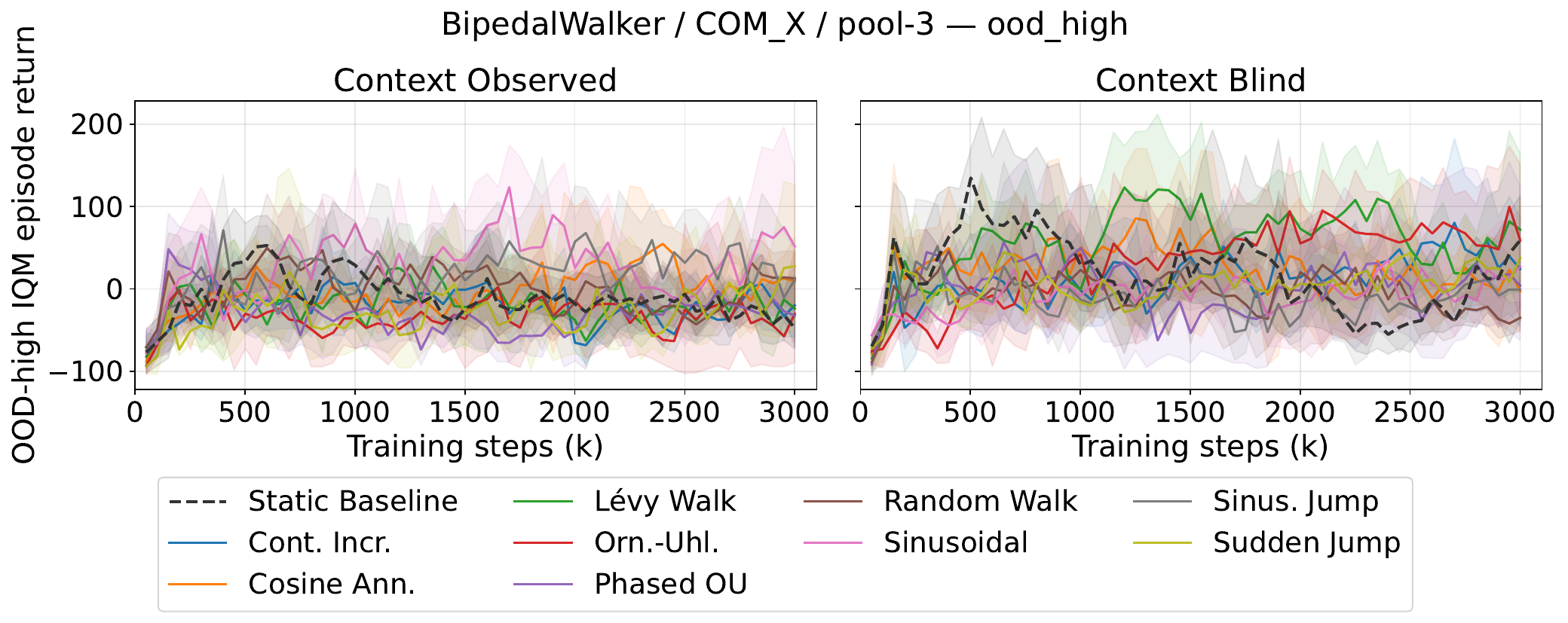}
    \caption{BipedalWalker / COM\_X / pool-3 — \textbf{OOD-high} IQM
      (centre-of-mass shifted above training range, 8 seeds).}
    \label{fig:line-walker-ood-high}
\end{figure} 


\begin{figure}[p]
    \centering
    \includegraphics[width=\linewidth]{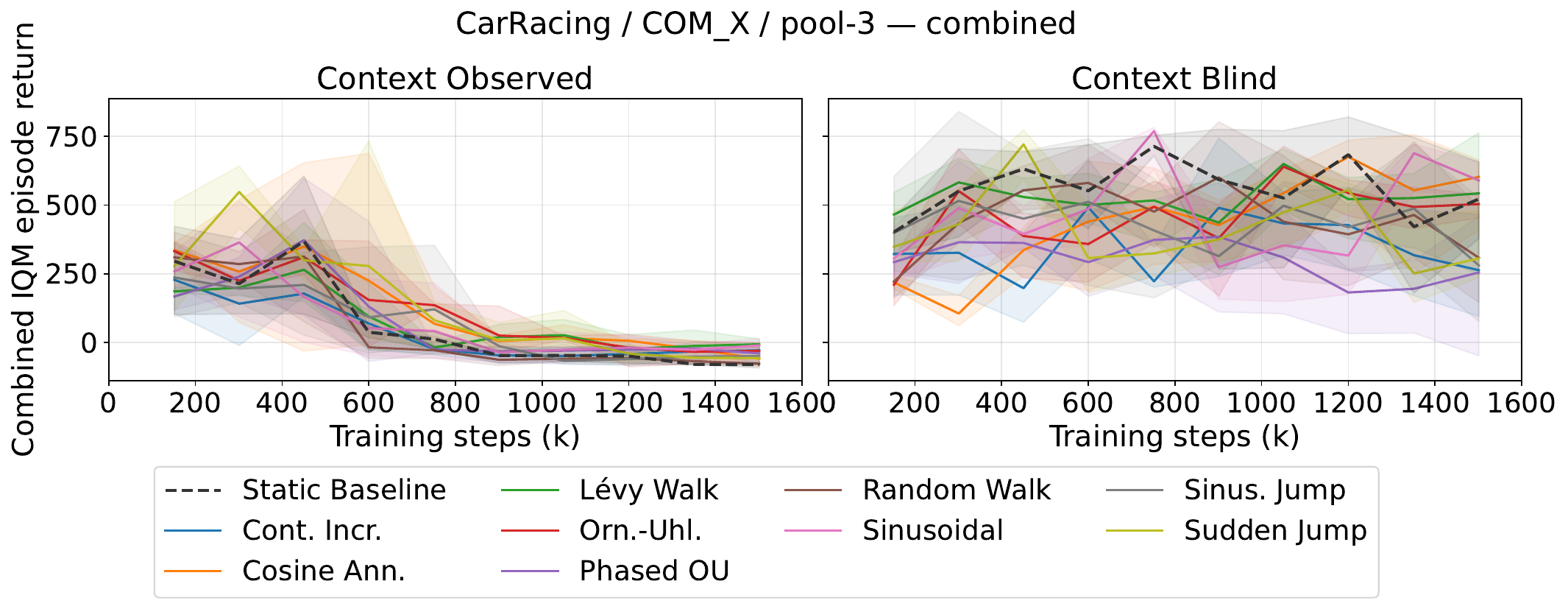}
    \caption{CarRacing / COM\_X (longitudinal CoM) / pool-3 —
      \textbf{combined} IQM (all contexts, 5 seeds).}
    \label{fig:line-carracing-comx-combined}
\end{figure}
  
\begin{figure}[p]
    \centering
    \includegraphics[width=\linewidth]{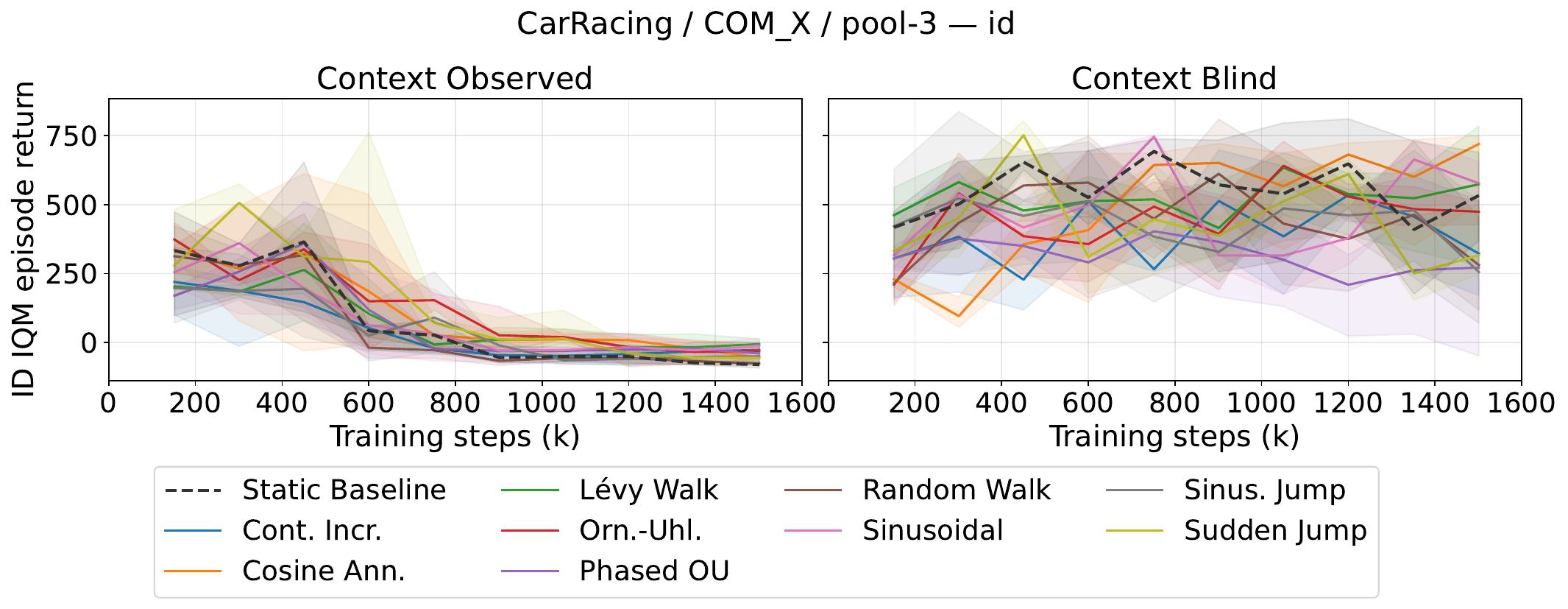}
    \caption{CarRacing / COM\_X / pool-3 — \textbf{ID} IQM
      (in-distribution contexts only, 5 seeds).}
    \label{fig:line-carracing-comx-id}
\end{figure}

\begin{figure}[p]
    \centering 
    \includegraphics[width=\linewidth]{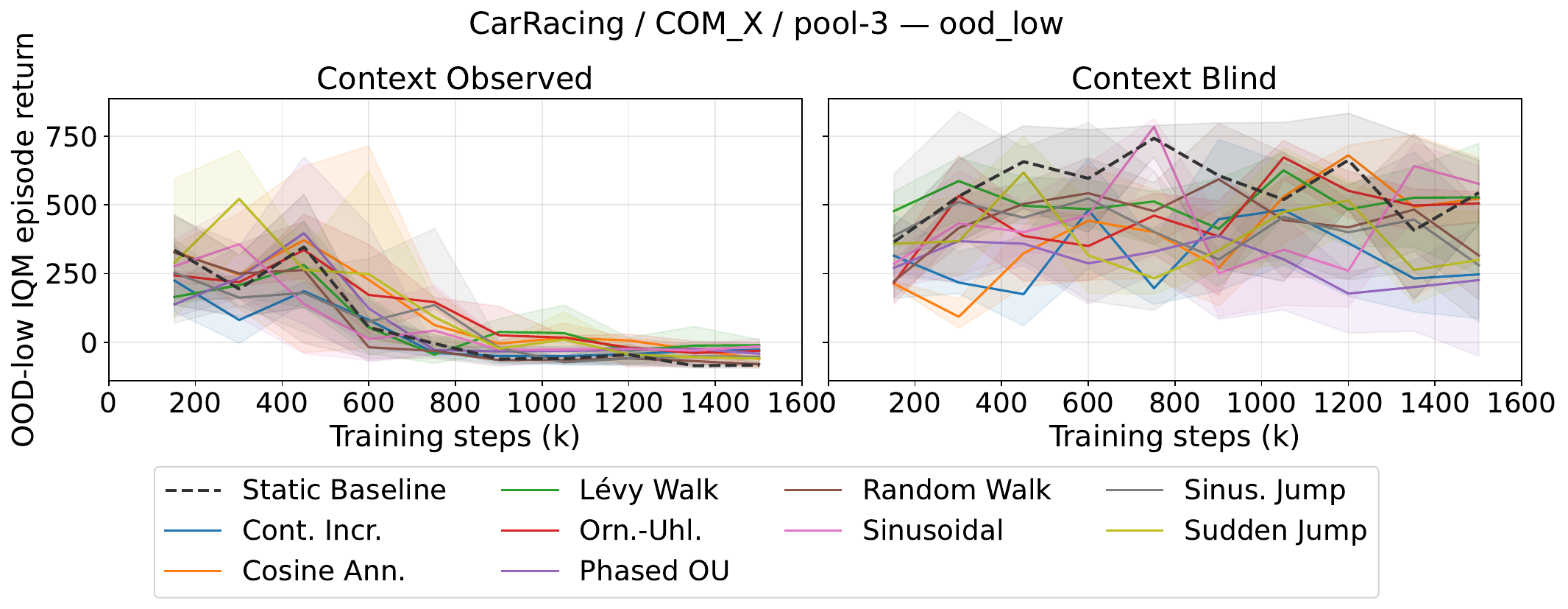}
    \caption{CarRacing / COM\_X / pool-3 — \textbf{OOD-low} IQM
      (longitudinal CoM shifted below training range, 5 seeds).}
    \label{fig:line-carracing-comx-ood-low}
\end{figure}
  
\begin{figure}[p]
    \centering
    \includegraphics[width=\linewidth]{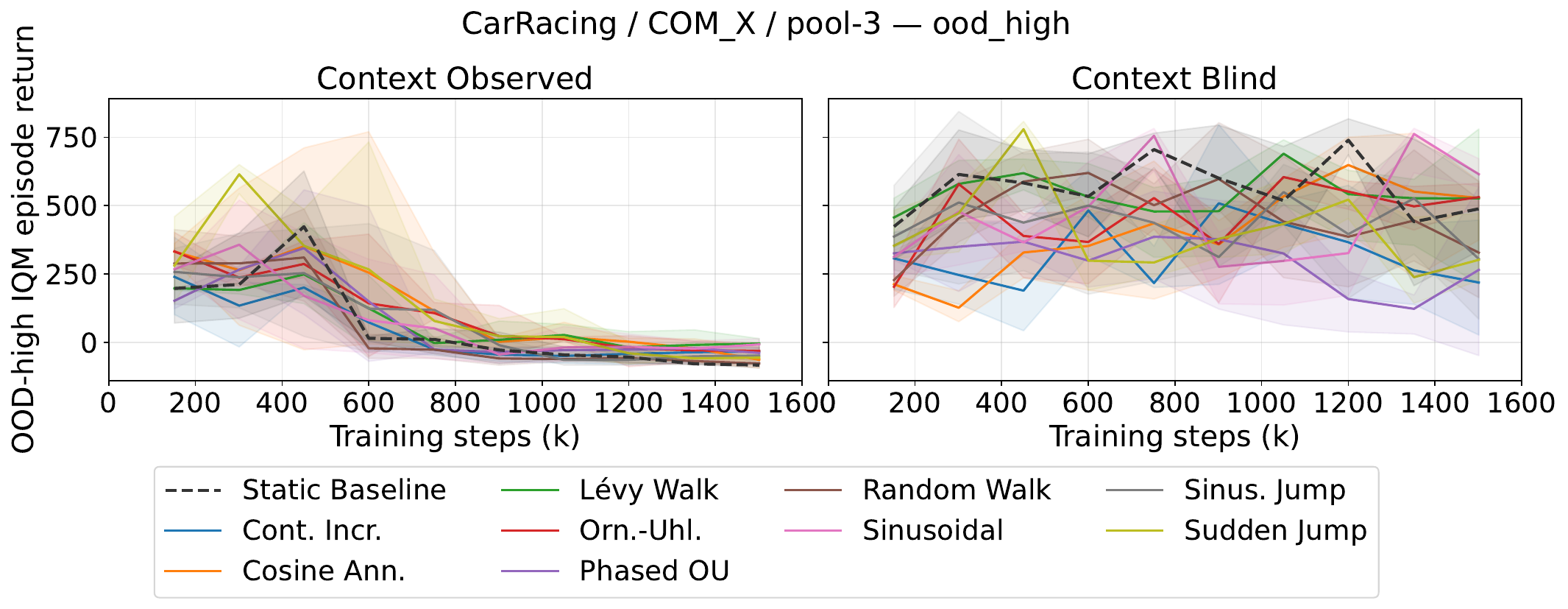}
    \caption{CarRacing / COM\_X / pool-3 — \textbf{OOD-high} IQM
      (longitudinal CoM shifted above training range, 5 seeds).}
    \label{fig:line-carracing-comx-ood-high}
\end{figure}
  

\begin{figure}[p]
    \centering
    \includegraphics[width=\linewidth]{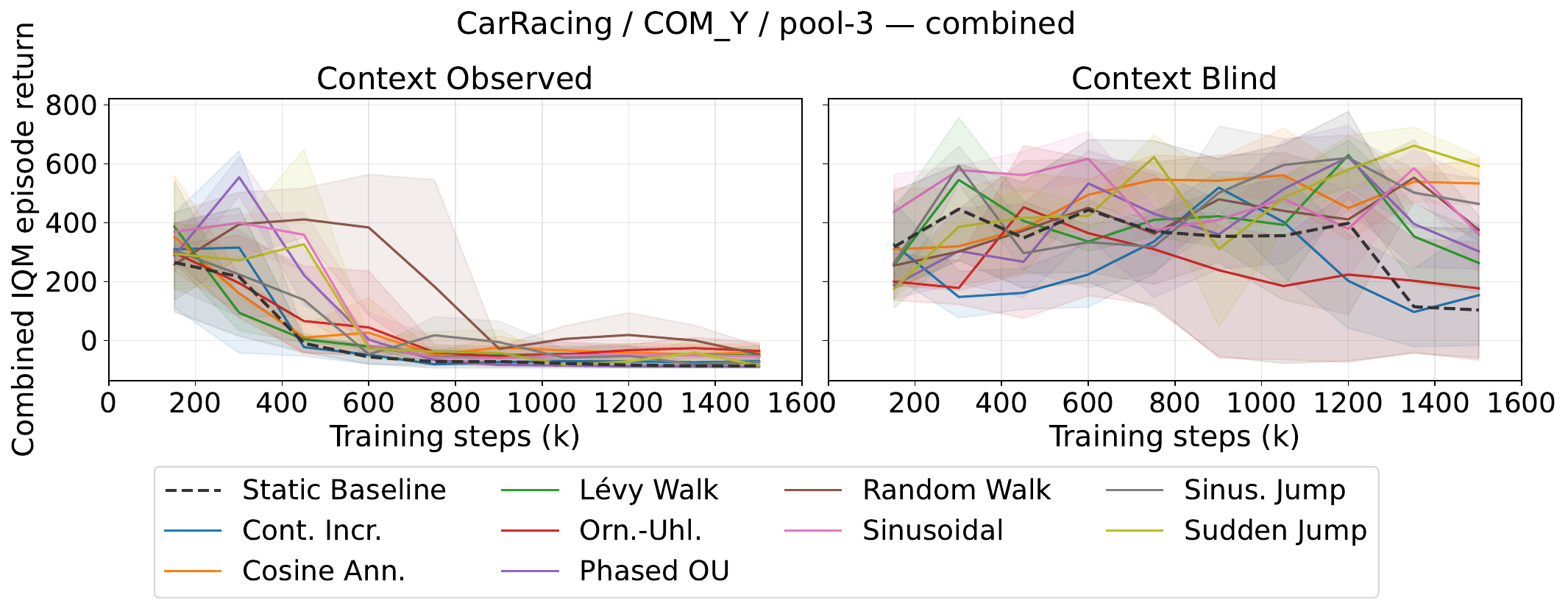}
    \caption{CarRacing / COM\_Y (lateral CoM) / pool-3 —
      \textbf{combined} IQM (all contexts, 5 seeds).}
    \label{fig:line-carracing-comy-combined}
\end{figure}

\begin{figure}[p]
    \centering
    \includegraphics[width=\linewidth]{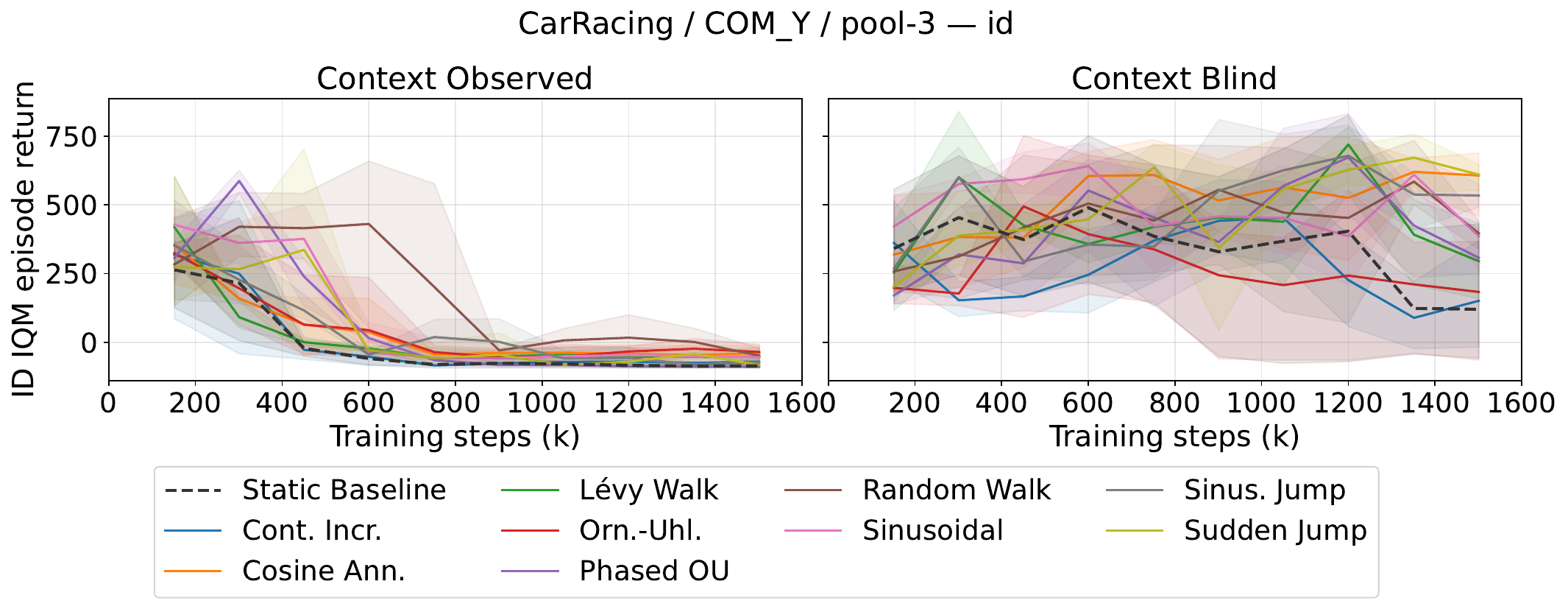}
    \caption{CarRacing / COM\_Y / pool-3 — \textbf{ID} IQM
      (in-distribution contexts only, 5 seeds).}
    \label{fig:line-carracing-comy-id}
\end{figure}

\begin{figure}[p]
    \centering
    \includegraphics[width=\linewidth]{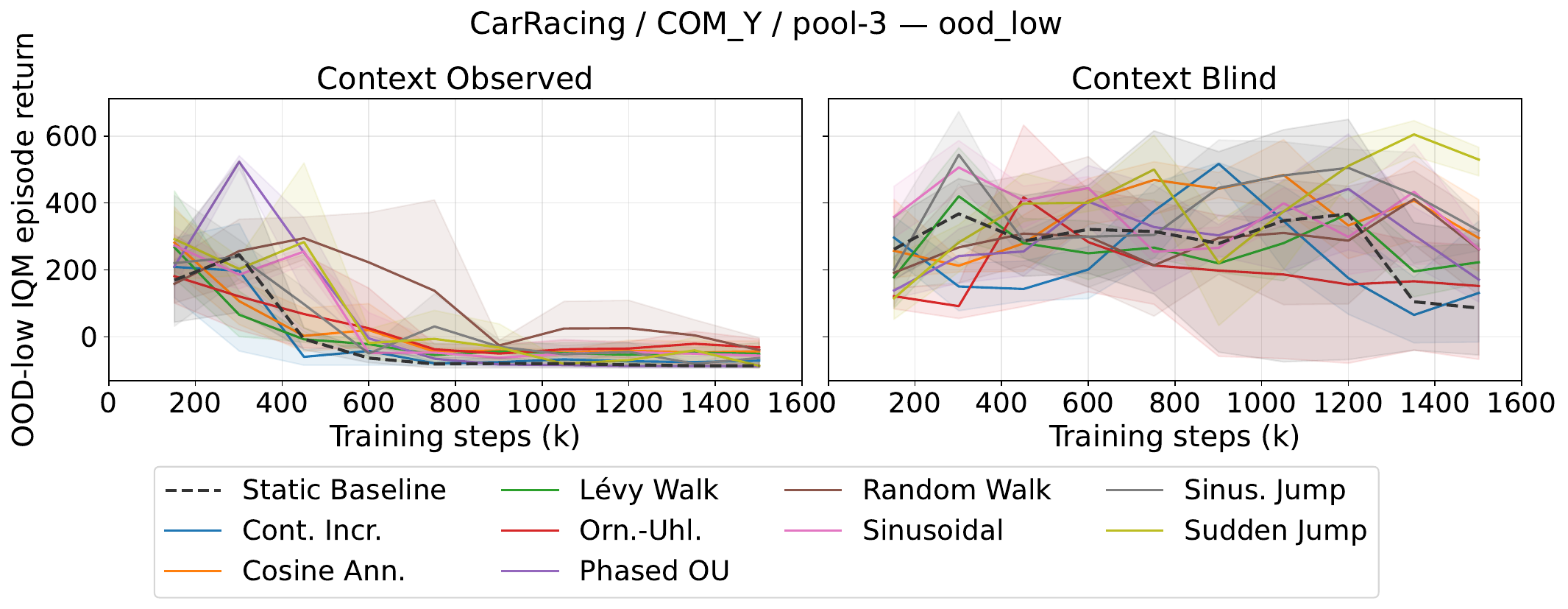}
    \caption{CarRacing / COM\_Y / pool-3 — \textbf{OOD-low} IQM
      (lateral CoM shifted below training range, 5 seeds).}
    \label{fig:line-carracing-comy-ood-low}
\end{figure}

\begin{figure}[p]
    \centering
    \includegraphics[width=\linewidth]{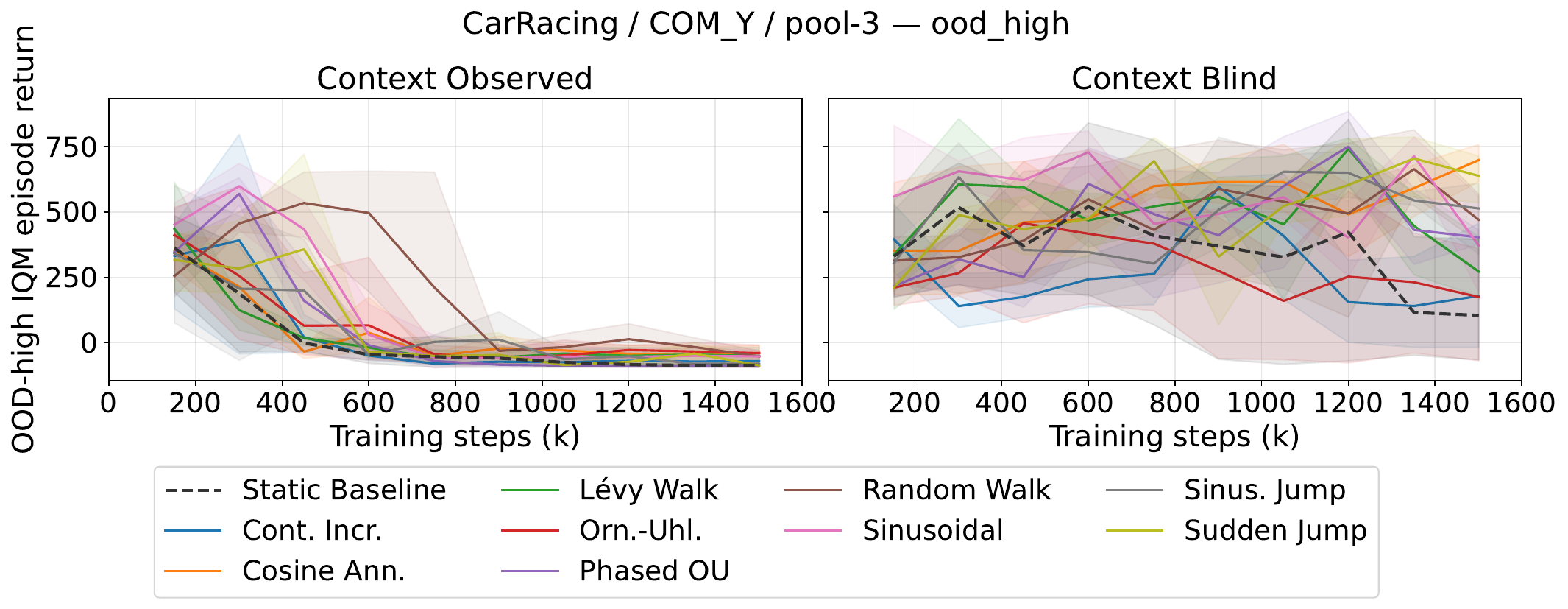}
    \caption{CarRacing / COM\_Y / pool-3 — \textbf{OOD-high} IQM
      (lateral CoM shifted above training range, 5 seeds).}
    \label{fig:line-carracing-comy-ood-high}
\end{figure}

\end{document}